\documentclass{article} % For LaTeX2e
\usepackage[final]{colm2026_conference}

\usepackage{microtype}
\usepackage{hyperref}
\usepackage{url}
\usepackage{booktabs}
\usepackage{graphicx}
\usepackage{subcaption}
\usepackage{xspace}
\usepackage{tcolorbox}
\usepackage{amsmath}
\usepackage{amssymb}
\usepackage{algorithm}
\usepackage{algpseudocode}
\usepackage{enumitem}
\usepackage{listings}
\usepackage{longtable}
\usepackage{array}
\usepackage{multirow}
\usepackage{placeins}
\usepackage{xparse}
\usepackage{xcolor}
\usepackage{amsfonts}
\usepackage{bbm}
\usepackage{wrapfig}
\usepackage{tikz}
\usetikzlibrary{positioning, shapes.geometric, calc, fit, backgrounds}

\usepackage{lineno}

\definecolor{darkblue}{RGB}{30,70,140}      % dark blue for text emphasis
\hypersetup{colorlinks=true, citecolor=darkblue, linkcolor=darkblue, urlcolor=darkblue}
\definecolor{bandblue}{RGB}{218,235,250}    % light blue band (controller/general)
\definecolor{bandgreen}{RGB}{218,245,225}   % light green band (generator)
\definecolor{bandorange}{RGB}{253,235,210}  % light orange band (evaluator)
\definecolor{bandred}{RGB}{250,220,220}     % light red band (select)
\definecolor{borderblue}{RGB}{50,100,180}   % blue border for nodes

\title{STATe-of-Thoughts: Structured Action Templates for Tree-of-Thoughts}

\author{
\textbf{Zachary E. Bamberger}$^{1, \ast}$ \quad \textbf{Till R. Saenger}$^{2, \ast}$ \quad \textbf{Gilad Morad}$^3$ \\ 
\textbf{Ofra Amir}$^1$ \quad \textbf{Brandon M. Stewart}$^2$ \quad \textbf{Amir Feder}$^4$ \\
\\[-0.5em]
$^1$Technion \quad $^2$Princeton University \quad
$^3$ Independent \quad $^4$Hebrew University \\
\\
{\small $^\ast$Equal contribution}
}

\lstdefinestyle{appendixprompt}{
  basicstyle=\ttfamily\footnotesize,
  breaklines=true,
  breakatwhitespace=false,
  frame=lines,
  backgroundcolor=\color{gray!10},
  xleftmargin=0pt,
  framexleftmargin=0pt,
  framexrightmargin=0pt,
}

\lstdefinestyle{dspypython}{
  language=Python,
  basicstyle=\ttfamily\footnotesize,
  breaklines=true,
  breakatwhitespace=false,
  frame=lines,
  backgroundcolor=\color{gray!8},
  xleftmargin=1em,
  framexleftmargin=0.5em,
  keywordstyle=\color{blue}\bfseries,
  commentstyle=\color{gray}\itshape,
  stringstyle=\color{teal},
  showstringspaces=false,
  tabsize=4,
  columns=flexible,
  keepspaces=true,
}

\newcommand{\STATeOfThoughts}{\textsc{STAT}{e}-of-Thoughts\xspace}
\newcommand{\state}{\textsc{STAT}{e}\xspace}
\newcommand{\prefixex}[1]{\textcolor{blue}{#1}}
\newcommand{\cont}[1]{\textcolor{orange}{#1}}
\newcommand{\internalr}[1]{\textcolor{teal}{#1}}
\newcommand{\repourl}{https://github.com/zbambergerNLP/state-of-thoughts}
\newcommand{\repolinkfile}[1]{%
  \href{\repourl/blob/main/\detokenize{#1}}{\nolinkurl{\detokenize{#1}}}%
}
\newcommand{\repolinkdir}[1]{%
  \href{\repourl/tree/main/\detokenize{#1}}{\nolinkurl{\detokenize{#1}}}%
}
\newcolumntype{L}[1]{>{\raggedright\arraybackslash}p{#1}}

\newcounter{prompt}
\newcommand{\promptcaption}[2]{%
  \refstepcounter{prompt}\label{#1}%
  \vspace{2pt}%
  \noindent\small\textit{Prompt~\theprompt: #2}\par\medskip
}
\newcommand{\promptcaptionnote}[3]{%
  \refstepcounter{prompt}\label{#1}%
  \vspace{2pt}%
  \noindent\small\textit{Prompt~\theprompt: #2} #3\par\medskip
}

\ifcolmfinal
\fi

\begin{document}

\ifcolmsubmission
\linenumbers
\fi

\maketitle
\begin{abstract}

Inference-Time-Compute (ITC) methods like Best-of-$n$ and Tree-of-Thoughts are meant to produce output candidates that are both high-quality and diverse, but their use of high-temperature sampling often fails to achieve meaningful output diversity.
Moreover, existing ITC methods offer limited control over \textit{how} to perform reasoning, which in turn limits their interpretability. 
We present \textbf{\STATeOfThoughts} (\state), an interpretable ITC method that \emph{searches} over high-level reasoning patterns.
\state branches over discrete and interpretable textual interventions rather than over token-level samples: a \textit{controller} selects actions encoding high-level reasoning choices; a \textit{generator} produces reasoning steps conditioned on those choices; and an \textit{evaluator} scores candidates to guide search.
% \state replaces stochastic sampling with discrete and interpretable textual interventions: a \textit{controller} selects actions encoding high-level reasoning choices; a \textit{generator} produces reasoning steps conditioned on those choices; and an \textit{evaluator} scores candidates to guide search.
This structured approach yields three main advantages. 
First, action-guided textual interventions reliably influence LLM generations and produce greater response diversity than temperature-based sampling. 
Second, in a case study on argument generation, \state's explicit action sequences capture interpretable features that are highly predictive of output quality. 
Third, estimating the association between performance and action choices allows us to identify promising yet unexplored regions of the action space and steer generation toward them.
\state is most useful when a task admits multiple solutions and when understanding \emph{why} an output succeeds matters beyond \emph{whether} the output succeeds.
Together, these results establish \state as both a practical framework for diverse and controllable text generation, and as a tool for understanding the reasoning patterns that drive performance.
\end{abstract}

\section{Introduction}

Many applications of LLMs require more than generating high-quality responses: they need systematic and interpretable control over how text is produced.
For example, in subjective tasks like persuasive writing, researchers vary the rhetorical structure and content themes of arguments to study the features that drive belief change \citep{tan_etal_2016_winning_arguments, saenger-etal-2024-autopersuade, salvi2024conversationalpersuasivenesslargelanguage, hackenburg_etal_2025_the_levers_of_political_persuasion, costello2026conspiracies}.
Similarly, in creative writing, researchers are concerned with generating diverse yet high-quality outputs that satisfy the preferences of the audience \citep{doshi2024generativeaicreativity, Lee2024AnEI, xu2025echoesinai}.
In both settings, the challenge is to produce text that varies systematically along dimensions of interest while maintaining coherence and quality.
Both settings admit multiple high-quality responses, and in both, \emph{why} a response succeeds matters beyond just \emph{whether} it does.

ITC methods address part of this challenge by allocating additional compute for LLM reasoning \citep{wei2022chain, kojima_etal_2022_zero_shot_reasoners} and for producing multiple candidate responses \citep{brown_etal_2020_gpt3, stiennon_etal_2020_learning_to_summarize, wang2022self}.
Tree-based methods \citep{beeching2024scalingtesttimecompute, hao2024llm, hao-etal-2023-reasoning}, like \citet{yao2023tree}'s Tree of Thoughts (ToT), further enhance quality by branching on intermediate thoughts and pruning less-promising reasoning trajectories.
However, these methods rely primarily on temperature-based sampling for diversity, which yields limited meaningful variation \citep{zhang2025noveltybenchevaluatinglanguagemodels, jiang2025artificial}.
Moreover, since ITC methods sample at the token-level, decisions about what to say and how to say it remain implicit in the decoding process \citep{Holtzman2020The, xie_etal_2020_top_k}. 
% As a result, they provide limited insight into which reasoning patterns drive success or failure.
As a result, they provide limited control over \emph{which} decisions are explored and limited insight into which decision patterns drive success or failure.

\begin{figure}[t!]
\centering
\includegraphics[width=0.9\textwidth]{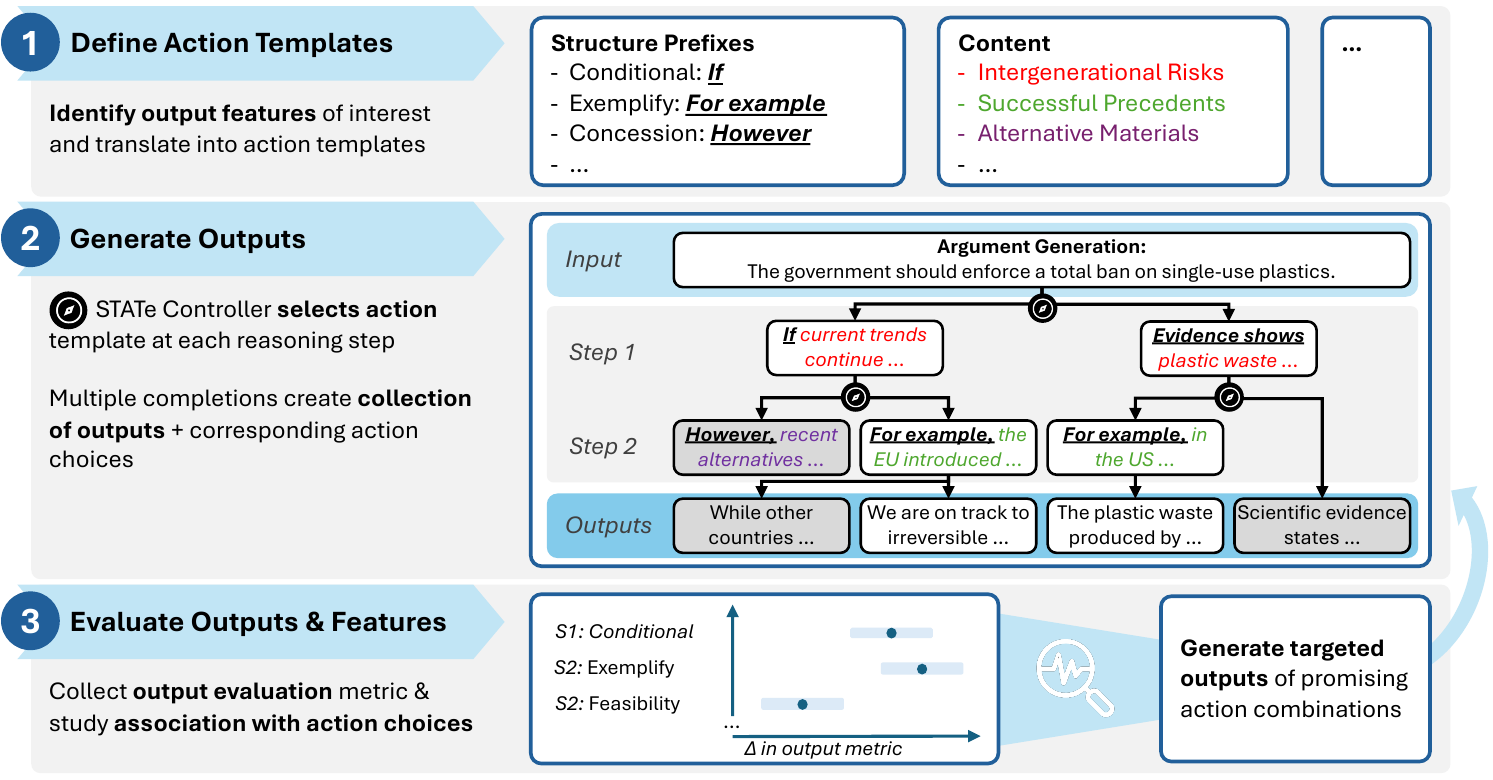}
\caption{
\textbf{\state for argument generation.} Tasked with generating persuasive arguments in favor of banning single-use plastics, \state's workflow involves the following steps: (1)~Define action templates that control output features of interest, such as structural prefixes and content themes. (2)~Generate outputs via tree search (Grey nodes indicate pruned branches; the rightmost path illustrates \textit{early stopping} after a single step). (3)~Evaluate outputs on a downstream metric, and study associations between action choices and performance.
}
\label{fig:workflow}
\end{figure}

To induce interpretable yet diverse sampling, we prepend prefixes to each LLM completion. 
Specifically, we define discrete \emph{action templates} that encode high-level reasoning choices (such as which rhetorical structure to employ, which content theme to develop, or which writing operation to perform).
We use intervention-based sampling to build \textbf{\STATeOfThoughts} (\state), an inference-time compute framework that searches over sequences of high-level reasoning actions.
\state's \emph{controller} selects which actions to explore at each reasoning step, and then its \emph{generator} produces reasoning steps conditioned on the selected actions.\footnote{
    Unlike latent interventions \citep{anthropic_golden_gate_claude_2024, durmus2024evaluating_feature_steering, anthropic_tracing_thoughts_2025, feldman2026causaleffectestimationlatent}, interventions in \state are explicit text prefixes and thus directly auditable.
}
An \emph{evaluator} scores both intermediate and final states to guide beam search \citep{beeching2024scalingtesttimecompute, hao2024llm}. We illustrate \state's three-step workflow in Figure~\ref{fig:workflow} through the lens of an argument generation task.

We compare \state to existing ITC methods in both creative writing and argument generation.
On NoveltyBench \citep{zhang2025noveltybenchevaluatinglanguagemodels} (Section \ref{sec:experiments-noveltybench}), we find that \state's branching mechanism produces outputs that are both more diverse and of higher quality than standard ITC branching.
In our case study on argument generation (Section \ref{sec:experiments-argument-generation}), we find that textual interventions reliably manifest in the generated reasoning steps and responses (Section \ref{sec:results-argument-generation-controllability}), that sequential action features are highly predictive of argument quality on held-out data (Section \ref{sec:results-argument-generation-predictability}), and that model-guided trajectory selection allows for generating high-quality outputs from promising yet unexplored regions of the action space (Section \ref{sec:results-targeted-trajectory}). 
Our work is open-sourced (\href{https://github.com/zbambergerNLP/state-of-thoughts}{github repo}) and provides the following contributions:
\begin{enumerate}[itemsep=2pt, topsep=4pt]
    \item A controllable ITC framework for action-space search.
    \item A diversity mechanism beyond high-temperature sampling.
    \item An action-based framework for analyzing the quality of reasoning patterns.
\end{enumerate}

\section{Background}

\subsection{Inference-Time Compute}
\label{sec:inference-time-compute}

The Input-Output (I/O) approach to using LLMs applies an input sequence (prompt) $x$ to model $p_\theta$ and produces output sequence $y$: $G_{I/O}(x) \rightarrow y$. 
While effective for many tasks, this approach exhibits limited robustness to common failure modes such as hallucinations \citep{simhi2024distinguishing, orgad2024llmsknowshowintrinsic, simhi-etal-2025-trust}, sycophancy \citep{sharma2024towards}, and other biases \citep{itzhak-etal-2024-instructed, orgad-belinkov-2023-blind}.
Building on the intuition that human reasoning benefits from more ``time to think'' \citep{kahneman2011thinking}, ITC methods provide LLMs with additional ``reasoning'' tokens \citep{pfau2024lets} to scale reasoning depth (Appendix~\ref{app:related-work-itc-depth}), and permit parallel reasoning attempts to scale reasoning breadth (Appendix~\ref{app:related-work-itc-breadth}). 

Chain-of-Thought (CoT) reasoning \citep{wei2022chain, kojima_etal_2022_zero_shot_reasoners} scales \emph{depth} by enabling models to generate intermediate reasoning steps before arriving at a final answer.
Formally, we define CoT as $G_{CoT}(x) \rightarrow Z, y$, where $Z$ is the chain of reasoning steps and $y$ the final answer.
While CoT improves performance on many reasoning tasks \citep{sprague2024cotcotchainofthoughthelps, deepseekai2025deepseekr1incentivizingreasoningcapability, openai2024openaio1card}, errors can propagate through the reasoning chain, and there is no principled mechanism to revisit decisions or explore alternative strategies.
Instead of scaling depth, Best-of-$n$ methods \citep{brown_etal_2020_gpt3, stiennon_etal_2020_learning_to_summarize} scale \emph{breadth} by generating $n$ independent candidate outputs and selecting the best according to some criterion.
This enhances robustness by reducing the impact of individual generation failures.
We refer to sampling more than one completion from an LLM as \emph{branching} \citep{yao2023tree}. 
In Best-of-$n$ methods, branching produces multiple complete reasoning chains along with their associated answers: $G_{CoT}(x; n, \texttt{temp}) \rightarrow \{(Z^1, y^1), \ldots, (Z^n, y^n)\}$, where each $Z^j$ represents a complete reasoning chain and $y^j$ is its corresponding final answer. 
Best-of-$n$ methods typically branch only at the initial reasoning step, without principled exploration of intermediate reasoning decisions.
Moreover, inducing diversity across branches through high-temperature sampling often yields homogeneous outputs or degrades quality \citep{minh2025turning, jiang2025artificial,zhang2025noveltybenchevaluatinglanguagemodels, zhang2025verbalized_sampling}.

\subsection{Tree of Thoughts}

Tree of Thoughts (ToT) \citep{yao2023tree} combines both ITC strategies: improving reasoning quality by scaling depth, and enhancing robustness by scaling breadth.
ToT methods (Appendix \ref{app:related-work-itc-tot}) reframe LLM generation as a search problem over a tree of partial reasoning steps.
ToT branches at each reasoning step, evaluates the quality of each branch, and prunes unpromising paths. 
Each node contains a state $s_i := [x, Z_i]$ that captures a partial solution with the input ($x$) and reasoning steps so far ($Z_{i-1}:= [z_1, \ldots, z_{i-1}]$).\footnote{
    For ease of notation, we denote the reasoning steps $z_1, \ldots, z_i$ by $Z_i$, but treat $s_i$ as a flat vector of inputs ($x$), reasoning steps ($z_1, \ldots, z_i$), and optionally final outputs ($y$), not a nested vector.
} A leaf node $s_{d+1}$ represents a complete solution $[x, Z_d, y]$ where $y$ is the final answer and $d$ is the predefined maximum reasoning depth. 
Formally, at step $i$, we sample candidate reasoning steps $\{z_i^1, \ldots, z_i^n\} \sim G_{ToT}(z_i)$, where $G_{ToT}(z_i) := p_\theta(z_i \mid x, Z_{i-1}; n, \texttt{temp})$. 
In practice, ToT often implements both intermediate and final evaluation through LLM-as-a-Judge \citep{zheng_etal_2023_llm_as_a_judge, alpaca_eval}.
Process Reward Models (PRMs) \citep{yao2023tree, lightman2024lets, wang2025valueguided} score partial trajectories to prune low-value branches and prevent exponential tree growth: $V(Z_i \mid x) \rightarrow [0, 1]$, $i \leq d$.
Conversely, Outcome Reward Models (ORMs) \citep{zheng_etal_2023_llm_as_a_judge, kim2024prometheus, kim_etal_2024_prometheus2} score completed outputs to select the best final answer: $V(y \mid x) \rightarrow [0, 1]$. 

Traditional ToT implementations face two primary limitations. 
First, sampling at high temperatures fails to promote diversity, since branches tend to cluster around similar content \citep{jiang2025artificial, zhang2025noveltybenchevaluatinglanguagemodels}. 
Second, ToT implementations perform a predetermined number of reasoning steps, which can lead to ``overthinking'' \citep{sprague2024cotcotchainofthoughthelps, liu2025mind, muennighoff2025s1, hong2025context} or insufficient reasoning.

\section{Methods}
\label{sec:methods}

\state replaces ToT's stochastic temperature sampling with discrete action templates that diversify branches in tree search. 
This allows each branch to explore fundamentally different reasoning strategies from its neighbors and enables ``early stopping'' (producing a final answer before depth $d$) if the reasoning so far is sufficient.
Moreover, \state tracks actions along a trajectory, enabling researchers to study associations between controllable, concept-level interventions and downstream outcomes \citep{goyal2019explaining, abraham_etal_2022_cebab}.

\subsection{\state Components}
\label{sec:methods-tree-of-thoughts-components}

At each layer $i$, \state starts with a list of states, each of the form $s_i = [x, Z_i]$. 
The controller selects $n$ interventions for each state in the frontier.
The generator then produces completions that extend each of these interventions.
Finally, the evaluator scores the resulting trajectories, and retains the top-$k$ states for the next layer. 
We present the full process in Algorithm \ref{alg:tot_controller_beam} and discuss its computational complexity in Appendix \ref{app:beam_search_call_complexity}.

\paragraph{Controller ($C$):} We treat each action as a \emph{tool call}. 
Selecting an action corresponds to choosing a tool name from a fixed set of action templates (Appendix \ref{app:action-spaces}) and providing values for the tool's arguments.
Given a parent state $s_{i-1} = [x, Z_{i-1}]$ representing the input and reasoning so far, the controller must choose up to $n$ actions from the action space $\mathcal{A}$ to explore in parallel branches (for guidance on how to build an action space, see Appendix \ref{app:action-space-guidance}).
Formally, we define the controller output as:
$\{a_{i}^1, \ldots, a_i^n\} = \text{C}(s_{i-1}, \mathcal{A}, n)$.
Implicitly, the controller implements a scoring function $Q(s_{i-1}, a_i)$ that estimates the value of taking action $a_i$ from state $s_{i-1}$ such that $\{a_i^1, \ldots, a_i^n\} = \arg\max_{A \subset \mathcal{A}, |A|=n} \sum_{a_i \in A} Q(s_{i-1}, a_i)$.
If the controller determines that reasoning is sufficient, it selects a dedicated \texttt{FINISH} action, signifying that the generator should produce a final answer.
This mechanism helps prevent ``overthinking'' where additional steps become degenerate after the model has effectively converged \citep{liu2025mind, ringel-etal-2025-learning, sui2025stop, hong2025context, muennighoff2025s1}. 
We present additional implementation details in Appendix \ref{app:state-modules-controller}.

\begin{figure}[h!]
\begin{tcolorbox}[
  colback=gray!6, colframe=gray!45, arc=3pt,
  left=5pt, right=5pt, top=3pt, bottom=3pt,
]
\begin{lstlisting}[
  basicstyle=\ttfamily\small,
  breaklines=true,
  frame=none,
  backgroundcolor={},
  xleftmargin=0pt,
  moredelim={**[is][\color{teal}]{@T}{T@}},
  moredelim={**[is][\color{blue}]{@B}{B@}},
  moredelim={**[is][\color{orange}]{@O}{O@}},
]
<thinking>
<step>
...
</step>
...
<step>
## internal_reasoning
@TI should examine case studies from ...T@
## claim
@BFor example,@B @OCalifornia's ban on single-use plastics demonstrates...@O
\end{lstlisting}
\end{tcolorbox}
\caption{Task: generate an argument for banning single-use plastics. The controller selects \texttt{\{"subtopic": "success\_of\_existing\_bans", "structure": "exemplification"\}} from the action space in Appendix~\ref{app:action-spaces-argument-generation}.
\textcolor{teal}{Internal reasoning} guides the model's next completion, while the \textcolor{blue}{prefix} forces the \textcolor{orange}{model's completion} to open with ``For example''.}
\label{fig:step-template}
\end{figure}

\paragraph{Generator ($G$):} For each action $a_i^j \in \{a^1_i, \ldots, a^n_i\}$, we ``execute'' the corresponding tool to obtain text guidance $a_i^j()$.
We append $a_i^j()$ to the state's existing reasoning,\footnote{
  prefilling \citep{muennighoff2025s1, bricken2025automating} injects text into the assistant message.
} and force it to generate text consistent with the chosen action (Figure \ref{fig:step-template}).
Formally, given the parent state $s_{i-1}=[x, Z_{i-1}]$, we sample a continuation using the generator: 
\begin{equation}
z_i^j \sim G \bigl(z \mid x, \texttt{prefill}({Z_{i-1}, a_i^j()}); \texttt{temp}\bigr)\,[:{\texttt{</step>}}]
\end{equation} 
for each action $a_i^j$.
We combine each generated thought $z_i^j$ with the current state to create a child state $s_i^j = [s_{i-1}, z_i^j]$.
At maximum depth $d$, or when the controller selects the \texttt{FINISH} action, \state reaches the \emph{synthesis} step, which produces final outputs:
\begin{equation}
y^j \sim G(y \mid x, \texttt{prefill}(Z_{i-1}, \texttt{FINISH}()); \texttt{temp})[:\texttt{</answer>}].
\end{equation}
We provide additional details on the Generator in Appendix \ref{app:state-modules-generator}.

\paragraph{Evaluator ($V_{PRM}$ \& $V_{ORM}$)} After generating child states from each parent, we evaluate their quality using either score-based LLM-as-a-Judge models \citep{zheng_etal_2023_llm_as_a_judge, kim2024prometheus, kim_etal_2024_prometheus2, liu2026humanai}, or verifiable rewards \citep{lambert2025tulu3pushingfrontiers, gao2024designingeffectiverlreward, kimiteam2025kimik15scalingreinforcement}. 
Following the Tree-of-Thoughts framework, we evaluate intermediate states $s_i = [x, Z_{i}]$ using a PRM, $V_{PRM}(s_i) := V(Z_i|x) \rightarrow [0, 1]$, and complete solution states $s_i = [x, Z_{i-1}, y]$ using an ORM, $V_{ORM}(s_i) := V(y|x) \rightarrow [0, 1]$.
Our LLM-based evaluators use custom rubrics that explicitly assess backward compatibility (coherence with prior reasoning steps) and forward compatibility (projected final answer quality) for intermediate reasoning steps, and task-specific criteria such as instruction adherence, coherence, and stylistic appropriateness for final outputs. 
See additional details in Appendix~\ref{app:state-modules-evaluator}.

\begin{algorithm}[h!]
    \caption{STATe-of-Thoughts$(x, G, C, V_\text{PRM}, V_\text{ORM}, \mathcal{A}, n, k, d, \texttt{temp})$}
    \label{alg:tot_controller_beam}
    \begin{algorithmic}[1]
    \Require Input $x$, generator $G$, controller $C$, process evaluator $V_\text{PRM}$, outcome evaluator $V_\text{ORM}$, action space $\mathcal{A}$, branching factor $n$, beam width $k$, depth $d$, temperature \texttt{temp}
    \State Initialize $L_0 \gets \{x\}$ \Comment{Initial layer with just the input}
    \State Initialize $F \gets \emptyset$ \Comment{Collection of final states with answers}
    \For{$i = 1$ to $d+1$}
        \State $L'_i \gets \emptyset$ \Comment{Candidate states for layer $i$}
        \For{each state $s_{i-1} \in L_{i-1}$}
            \State $\mathcal{A}_i \gets \{\texttt{FINISH}\}$ if $i=d+1$, else $C(s_{i-1}, \mathcal{A}, n)$ \Comment{Select actions or finish}
            \For{each action $a_i^j \in \mathcal{A}_i$}
                \If{$a_i^j$ is \texttt{FINISH}}
                    \State $y^j \sim G(s_{i-1}, \texttt{prefill}(Z_{i-1}, a_i^j()); \texttt{temp})[:\texttt{</answer>}]$
                    \Comment{Generate response}
                    \State $s_i \gets [s_{i-1}, y^j]$ \Comment{Create final state}
                    \State $F \gets F \cup \{s_i\}$ \Comment{Add to collection of final states}
                \Else
                    \State $z_i^j \sim G(s_{i-1}, \texttt{prefill}(Z_{i-1}, a_i^j()); \texttt{temp})[:\texttt{</step>}]$ \Comment{Generate thought}
                    \State $s_i \gets [s_{i-1}, z_i^j]$ \Comment{Create new intermediate state}
                    \State $L'_i \gets L'_i \cup \{s_i\}$ \Comment{Add to next layer's candidates}
                \EndIf
            \EndFor
        \EndFor
        \If{$L'_i = \emptyset$} \textbf{break} \Comment{All branches terminated via early stopping} \EndIf
        \State Score all candidates: $v_{s_i} \gets V_\text{PRM}(s_i)$ for all $s_i \in L'_i$
        \State $L_i \gets \arg\max_{L \subset L'_i, |L|=\min(k, |L'_i|)} \sum_{s_i \in L} v_{s_i}$ \Comment{Select top-$k$ states for layer $i$}
    \EndFor
    \State Score all final states: $v_s \gets V_\text{ORM}(s)$ for all $s \in F$
    \State \textbf{return} $\arg\max_{s \in F} v_s$ \Comment{Return highest-scoring final state}
    \end{algorithmic}
    \end{algorithm}

\subsection{Attributing Outcomes to Controller Actions}
\label{sec:association-analysis}

A key advantage of \STATeOfThoughts is its ability to attribute differences in outcomes to specific controller actions, since each branch in the reasoning tree carries a logged action sequence.
However, estimating causal effects is complicated by sequential confounding: actions are selected conditional on prior actions in the same sequence.
We therefore focus on associational analysis, aiming to identify action patterns that consistently correlate with better or worse outcomes.
Let $\tau = (a_1, a_2, \ldots, a_n)$ denote a complete action sequence.
We explore whether the \textit{sequential structure} of actions matters beyond their mere presence.
% A central question is whether the \textit{sequential structure} of actions matters beyond their mere presence.

A \textbf{presence-based model} represents actions through binary indicators, $\mathbf{1}_a(\tau) \in \{0,1\}^{|\mathcal{A}|-1}$, to determine whether the action type $a$ appears anywhere in $\tau$, and fits $Y_i = \alpha + \mathbf{1}_a(\tau_i) \boldsymbol{\beta} + \epsilon_i$.
Conversely, a \textbf{sequential model} extends this with (i) \textit{position features} $\mathbf{1}_{a,k}(\tau)$, indicating whether action $a$ occurs at step $k$, and (ii) \textit{transition features} $\mathbf{1}_{a \to a'}^{k \to k+1}(\tau) = \mathbf{1}_{a,k}(\tau) \cdot \mathbf{1}_{a',k+1}(\tau)$, capturing consecutive action bigrams.
When the action space is multi-dimensional, cross-dimensional interactions at each step can be included as additional features.

\section{Experiments}
\label{sec:experiments}
We evaluate \state in two settings that probe its capacity for diversity, controllability, and interpretability.
First, we compare \state to existing ITC methods on NoveltyBench \citep{zhang2025noveltybenchevaluatinglanguagemodels} to test whether structured interventions improve semantic \emph{diversity} (Section~\ref{sec:experiments-noveltybench}).
Next, in Section \ref{sec:experiments-argument-generation}, we use \state for a case study on argument generation. 
In Section \ref{sec:results-argument-generation-controllability} we measure the \emph{controllability} of our interventions by the frequency with which they materialize in generated reasoning steps (claims) and final responses (arguments).
We then test whether \state's action sequences improve \emph{predictability} of argument quality scores by LLM-judges (Section \ref{sec:results-argument-generation-predictability}), and demonstrate that these scores correlate with human annotations (Section \ref{sec:results-human-validation}).
Finally, we show that learned associations can guide \textit{discovery} of promising regions of the action space (Section \ref{sec:results-targeted-trajectory}).

\subsection{Improving Diversity and Quality in Creative Writing}
\label{sec:experiments-noveltybench}

\textbf{Setup:} We evaluate the diversity of \state's branching mechanism on NoveltyBench \citep{zhang2025noveltybenchevaluatinglanguagemodels}, using its curated 100-prompt set for creative writing.\footnote{
    We validate our setup on 10 prompts and test on the remaining 90 (see Appendix \ref{app:noveltybench}).
}
Each generation method (I/O, CoT, ToT, and \state) produces 8 responses per prompt. 
For \state, the action space combines two dimensions (Appendix~\ref{app:action-spaces-noveltybench}): \emph{personality traits} (following the Big Five model; \citealp{goldberg1990alternative}) and \emph{target audience} (demographic age to appeal to). 
We report NoveltyBench's \textit{diversity} metric, the number of functional equivalence classes induced by a fine-tuned DeBERTa \citep{he2021deberta} embedding space across the response set.
Since diversity often comes at the cost of \textit{quality}, we also report NoveltyBench's quality score, based on LLM evaluations \citep{liu2026humanai}.
For ToT and \state, we isolate and measure the diversity of the branching mechanism by restricting search to shallow trees ($d$=1).\footnote{
    Deeper heuristic search optimizes for evaluator-aligned scores rather than frontier diversity. 
    Moreover, deeper trajectories often share parent states, introducing overlapping reasoning.
} We set $n$=$k$=8 and repeat each configuration across 10 random seeds and 3 temperature regimes (low, medium, high). 
All experiments follow a strict development/evaluation separation to prevent post-hoc selection.
We provide additional details and ablation studies in Appendix~\ref{app:noveltybench}.

\textbf{Results:} \state improves both the semantic diversity of responses and their perceived quality across all three temperature regimes (Table~\ref{tab:noveltybench-qwen3-30b-diversity-quality}).
Relative to the best non-\state baseline (CoT with action space), \state improves diversity by $42\%$ at $T{=}0.5$ ($4.24$ vs.\ $2.98$), $37\%$ at $T{=}0.7$ ($4.57$ vs.\ $3.33$), and $31\%$ at $T{=}1.0$ ($4.94$ vs.\ $3.76$).
While diversity often comes at the cost of quality, \state's intervention-based branching mechanism also outperforms the strongest baseline in quality: $30\%$ gains at $T{=}0.5$ ($3.36$ vs.\ $2.59$), $21\%$ at $T{=}0.7$ ($3.52$ vs.\ $2.90$), and $16\%$ ($3.73$ vs.\ $3.23$) at $T{=}1.0$.
With \state, Qwen-3-30B-A3B-Instruct comes closest to reaching human performance on NoveltyBench (accessible \href{https://github.com/novelty-bench/novelty-bench/blob/main/data/humans.jsonl}{here}) in diversity (5.58) and quality (4.37).
Neither ToT nor the inclusion of actions in the prompt, in isolation, matches \state's performance, suggesting that prefix-based interventions provide a meaningful boost.
In Appendix~\ref{app:noveltybench-generalizability} we demonstrate that \state's performance on NoveltyBench generalizes over 7 models from 4 families: Qwen3 \citep{yang2025qwen3technicalreport}, Gemma-3 \citep{gemmateam2025gemma3technicalreport}, Nemotron-3 \citep{nvidia2025nvidianemotron3efficient}, and Ministral-3 \citep{liu2026ministral3}.

\begin{table}[h!]
\centering
\footnotesize
\begin{tabular}{lcccccc}
\toprule
 & \multicolumn{2}{c}{T=0.5} & \multicolumn{2}{c}{T=0.7} & \multicolumn{2}{c}{T=1.0} \\
\cmidrule(lr){2-3} \cmidrule(lr){4-5} \cmidrule(lr){6-7}
\textbf{Method} & \textbf{Diversity} & \textbf{Quality} & \textbf{Diversity} & \textbf{Quality} & \textbf{Diversity} & \textbf{Quality} \\
\midrule
I/O & 1.68\,$\pm$\,0.05 & 1.67\,$\pm$\,0.05 & 1.98\,$\pm$\,0.03 & 1.9\,$\pm$\,0.04 & 2.41\,$\pm$\,0.05 & 2.25\,$\pm$\,0.05 \\
CoT & 2.31\,$\pm$\,0.06 & 2.13\,$\pm$\,0.06 & 2.59\,$\pm$\,0.09 & 2.31\,$\pm$\,0.08 & 3.0\,$\pm$\,0.1 & 2.66\,$\pm$\,0.11 \\
I/O w/ Actions & 1.94\,$\pm$\,0.05 & 1.69\,$\pm$\,0.04 & 2.26\,$\pm$\,0.1 & 1.91\,$\pm$\,0.1 & 2.84\,$\pm$\,0.09 & 2.37\,$\pm$\,0.09 \\
CoT w/ Actions & \underline{2.98\,$\pm$\,0.09} & \underline{2.59\,$\pm$\,0.08} & \underline{3.33\,$\pm$\,0.12} & \underline{2.9\,$\pm$\,0.1} & \underline{3.76\,$\pm$\,0.1} & \underline{3.23\,$\pm$\,0.1} \\
ToT & 1.97\,$\pm$\,0.05 & 1.72\,$\pm$\,0.06 & 2.27\,$\pm$\,0.05 & 1.99\,$\pm$\,0.06 & 2.78\,$\pm$\,0.11 & 2.4\,$\pm$\,0.08 \\
ToT w/ Actions & 2.38\,$\pm$\,0.06 & 1.99\,$\pm$\,0.06 & 2.76\,$\pm$\,0.08 & 2.32\,$\pm$\,0.06 & 3.29\,$\pm$\,0.11 & 2.7\,$\pm$\,0.12 \\
STATe of Thoughts & \textbf{4.24\,$\pm$\,0.11} & \textbf{3.36\,$\pm$\,0.09} & \textbf{4.57\,$\pm$\,0.13} & \textbf{3.52\,$\pm$\,0.08} & \textbf{4.94\,$\pm$\,0.1} & \textbf{3.73\,$\pm$\,0.09} \\
\bottomrule
\end{tabular}
\caption{NoveltyBench diversity and quality for Qwen3-30B across ITC methods and temperatures (mean$\pm$ std over 10 seeds). Best performance in \textbf{bold}, runner-up \underline{underlined}.}
\label{tab:noveltybench-qwen3-30b-diversity-quality}
\end{table}

\subsection{Analyzing What Makes an Argument Effective}
\label{sec:experiments-argument-generation}

We conduct a case study on argument generation, in which an LLM must produce an argument in favor of a provided topic.
For our action space, we instantiate two dimensions suggested by \citet{wachsmuth-etal-2017-computational}: content (subtopics to discuss) and structure (discourse relations; \citealp{prasad-etal-2008-penn, webber2019penn}), detailed in Appendix~\ref{app:action-spaces-argument-generation}.

\subsubsection{Granular Control of Argumentative Reasoning} 
\label{sec:results-argument-generation-controllability}

\textbf{Setup:} We generate 1,000 arguments with \state on a fixed topic and use an LLM as a Judge (GPT-5-mini; \citealp{openai_gpt5_systemcard_2025}) to verify whether interventions materialize in individual reasoning steps (claims) and in final responses (arguments).
At the \emph{step-level} we verify for each step whether (i) it exhibits its prescribed discourse structure and (ii) it discusses its prescribed subtopic.
At the \emph{response-level} we verify whether the argument reflects each step's prescribed structure and subtopic, and whether the prescribed ordering across steps is preserved (see prompt templates in Appendix~\ref{app:controllability-evaluation}).

\textbf{Results:} We find that controller interventions reliably manifest in the LLM's generated text (Table~\ref{tab:controllability-strict}).
Structure adherence at the step level is near-perfect (99.7\%), confirming that the prefix mechanism reliably controls the discourse structure of the reasoning step.
Subtopic adherence at the step level is strong but lower (87.8\%), reflecting that content guidance operates through text-based guidance rather than explicit prefilling.
Response-level structure (96.2\%) and subtopic (93.5\%) pass rates confirm that prescribed properties propagate through response synthesis.
Moreover, the order of subtopics and structural decisions is mostly preserved (87.9\%).
In Appendix~\ref{app:controllability-evaluation} we discuss the impact of the Generator's synthesis prompt (Appendix \ref{app:state-modules-generator}) on the faithfulness of interventions.

\begin{table}[htbp]
\centering
\small
\begin{tabular}{lrcc}
\toprule
Check Category & N & Pass Rate (\%) & 95\% CI \\
\midrule
Step structure & 3,000 & 99.7 & [99.5, 99.9] \\
Step subtopic & 3,000 & 87.8 & [86.6, 89.0] \\
Final structure & 3,000 & 96.2 & [95.5, 96.9] \\
Final subtopic & 3,000 & 93.5 & [92.6, 94.4] \\
Order preservation & 1,000 & 87.9 & [85.9, 89.9] \\
\midrule
\textbf{Overall (all 13)} & \textbf{13,000} & \textbf{93.8} & \textbf{[93.3, 94.4]} \\
\bottomrule
\end{tabular}
\caption{Controllability evaluation for strict synthesis (plastic pollution). Pass rates assessed by LLM judge across 13 boolean checks per argument. 95\% bootstrap CIs computed by resampling arguments ($B = 10{,}000$).}
\label{tab:controllability-strict}
\end{table}

\subsubsection{Predicting the Quality of Arguments through Action Sequences}
\label{sec:results-argument-generation-predictability}

\textbf{Setup:} We evaluate the quality of arguments across 5 topics with 3 LLM judges \citep{openai_gpt5_systemcard_2025,deepmind_gemini_3_1_flash_lite_2026,anthropic_claude_haiku_4_5_2025} (Table \ref{tab:topic-overview}).
We quantify the quality of arguments through pairwise comparisons that we aggregate into ranks based on Bradley--Terry scores \citep{bradley1952rank}.
Using \state with Qwen3-30B-A3B-Instruct, we generate 5,000 arguments from 20 trees (with $d$ = 3, $n$ = 100, $k$ = 250), each initialized with a different random seed.
We then fit attribution models (Section \ref{sec:association-analysis}) that map controller-action trajectories to final argument quality.
Our simplest model (M0) only captures argument length.\footnote{ 
    All attribution models include argument length (number of characters) as a baseline feature since LLM-as-a-Judge is biased towards long responses \citep{dubois2024length, saenger-etal-2024-autopersuade}.
}
The presence-only models (M1a: structure only, M1b: content only, M1c: both) add binary indicators for which actions appeared in the trajectory.
The sequential model (M2) additionally encodes step position, within-step content--structure interactions, and cross-step transitions.
See Appendix~\ref{app:argument-generation-additional-results} for additional details.

\begin{figure}[ht]
  \centering
  \includegraphics[width=\textwidth]{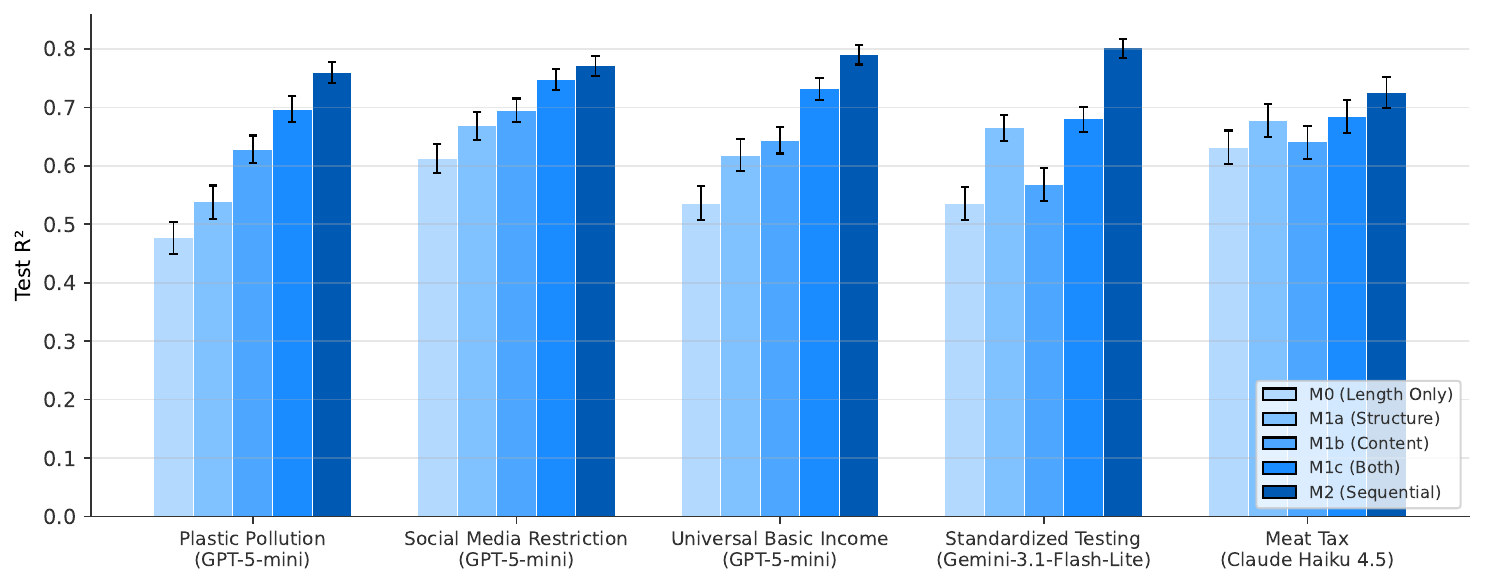}
  \caption{Predictability of argument quality from controller actions across argument topics and LLM judges.
  Each panel shows the performance (R$^2$ including 95\% bootstrap CIs) on the held-out test set (40\% of data).}
  \label{fig:argument-attribution-main}
  \end{figure}

\textbf{Results:} We apply the attribution framework of Section~\ref{sec:association-analysis} to reasoning trajectories for argument generation (Section~\ref{sec:experiments-argument-generation}).
Across all topics and judges, the sequential model (M2) substantially outperforms the presence-based baseline (M1a-c) in predicting the effectiveness of arguments out-of-sample (Figure~\ref{fig:argument-attribution-main}).
% In other words, the temporal structure of controller decisions carries predictive information about output quality.
% In other words, not only which actions the controller takes, but when it takes them and how they combine within a step, carries predictive information about output quality.
In other words, predictive information comes not only from which actions the controller takes, but also from when it takes them and how they combine within a step.
We present additional experimental details and ablations in Appendix~\ref{app:argument-generation-additional-results}.

\subsubsection{Validating LLM-as-a-Judge with Human Preferences}
\label{sec:results-human-validation}

\textbf{Setup:} The attribution results above demonstrate the mechanistic finding that action \emph{sequences} predict argument quality as measured by an LLM judge.
This finding holds for any consistent preference signal since our machinery is agnostic to the source of the outcome variable.
To determine whether the LLM judge also captures meaningful human preferences, we conducted a validation study with 288 Prolific participants (Appendix~\ref{app:human-study}).
The study focused on one topic (single-use plastics ban) and one judge (GPT-5-mini), using arguments drawn from the same pool evaluated in our attribution analysis.
Each participant evaluated 10 argument pairs in a forced binary choice, drawn from a pool of 50 pairs.
For our evaluation, we draw on the ALT test \citep{calderon2025alternativeannotatortestllmasajudge}, a leave-one-out procedure that compares how well the LLM judge aligns with the annotator pool relative to each individual human annotator (Section~\ref{sec:alt-test}).

\textbf{Results:} The ALT test's advantage probability reaches 0.841, meaning the LLM judge matches or exceeds an excluded human's alignment with the remaining pool 84.1\% of the time.
The LLM agrees with the human majority vote 72.0\% of the time, and its choices correlate significantly with human preference rates.\footnote{See Appendix~\ref{app:human-study} for more details.}
For context, human pairwise agreement on this task is only 56.0\% (Krippendorff's $\alpha = 0.118$), reflecting the inherent subjectivity of persuasiveness judgments \citep{rescala-etal-2024-language}.
Importantly, even if the LLM judge were poorly aligned with human preferences, the attribution methodology we present would remain valid. \state learns to predict whatever preference signal is provided, and the same pipeline applies with human respondents as the outcome source.
We use an LLM judge because it provides cheap, fast, and reproducible evaluations at scale, but the human study confirms that, for this topic and judge, the LLM signal is well aligned with human judgment.
Full study details, including design, metrics, and stratified results, are provided in Appendix~\ref{app:human-study}.

\subsubsection{Discovering Promising Unexplored Action Sequences}
\label{sec:results-targeted-trajectory}

\textbf{Setup:} We test whether M2's learned coefficients generalize beyond observed trajectories and can guide search toward high-quality, previously unseen regions of the action space.
Concretely, we score unseen trajectories with M2, generate targeted arguments from top-ranked trajectories, and compare them against random exploration and simpler topic-presence guidance (M1b).
To mitigate length confounding, we evaluate all comparisons on length-matched sets.\footnote{
For each targeted argument, we find the closest-length baseline argument within $\pm5$ characters,
using each baseline argument at most once.
}

\begin{table}[htbp]
\centering
\small
\begin{tabular}{l r c r r}
\toprule
Baseline & Win (T) (\%) & 95\% CI & Top-10 & Top-100 \\
\midrule
Random & 78.7 & [73.9, 83.2] & 8/10 & 78/100 \\
M1b (Topic Presence) & 63.3 & [57.3, 69.2] & 6/10 & 57/100 \\
Original Top 5\% & 68.0 & [60.9, 74.8] & 9/10 & 68/100 \\
\bottomrule
\end{tabular}
\caption{Targeted trajectory exploration vs.\ baselines ($N{=}204$--$354$ length-matched arguments, 5{,}000 pairwise comparisons each).
\textbf{Win (T)}: share of pairwise wins by targeted arguments, with 95\% bootstrap CIs ($B{=}10{,}000$); see Appendix~\ref{app:targeted-trajectory-details}.
\textbf{Top-10/Top-100}: targeted arguments among the top $n$ by Bradley--Terry score.}
\label{tab:targeted-trajectory-results}
\end{table}

\textbf{Results:} In Table~\ref{tab:targeted-trajectory-results} we show that targeted arguments substantially outperform the random baseline (78.7\% win rate), the topic-presence baseline (63.3\% win rate), and the original top 5\% baseline (68.0\% win rate).
This confirms that M2's trajectory rankings identify genuinely promising regions of the action space, more so than a simpler presence-based approach, analogous to a topic model.
We provide additional details and results in Appendix~\ref{app:targeted-trajectory-details}.

\section{Discussion}
\label{sec:discussion}

We developed \STATeOfThoughts\ (\state) as a controllable inference-time compute framework that makes step-level decisions explicit and auditable (Section \ref{sec:methods}). 
On NoveltyBench (Section \ref{sec:experiments-noveltybench}), \state not only produces substantially higher semantic diversity but also improves output quality, demonstrating that intervention-based branching can produce diverse candidates without the typical quality degradation associated with high-temperature sampling.
Furthermore, \state opens up ITC as a tool for exploring what makes open-ended writing effective or ineffective.
In our controllability study, we find that \state's interventions reliably manifest in both intermediate reasoning and final outputs (Section \ref{sec:results-argument-generation-controllability}).
When evaluating predictive power, we show that action \emph{sequences} (not just action presence) improve outcome predictions (Section \ref{sec:results-argument-generation-predictability}).
Crucially, we also show that these learned associations can be operationalized: by scoring and targeting previously unseen trajectories, \state can systematically explore under-visited regions of the controllable feature space and surface strong candidates, rather than repeatedly sampling near-duplicates (Section \ref{sec:results-targeted-trajectory}).
Taken together, these results position \state as a practical method to (1) generate diverse yet high-quality texts, (2) understand which writing strategies drive quality, and (3) discover and target promising new strategies.

\section{Limitations}

\state has several practical limitations.
First, our method relies on prefilling for interventions, but modern closed-source APIs (e.g., GPT, Claude, Gemini) do not expose this functionality.
Second, our action--outcome analysis is associative rather than causal, as the design introduces sequential confounding that our current attribution models do not address.\footnote{
    Notably, our inability to make causal claims does not stem from issues with the \state framework, but rather from how we process trajectories in our study (Section \ref{sec:results-argument-generation-predictability}).
}
Third, \state's interventions strictly involve adding a new reasoning step to an existing trajectory, which limits its expressivity.
\state does not support interventions that affect final output generation, nor does it support interventions that \emph{alter} rather than \emph{extend} existing content.
Fourth, the synthesis step that converts reasoning traces into final outputs introduces a control--quality trade-off.
Concretely, strict synthesis preserves reasoning faithfulness and enables high predictability but can produce stilted prose, whereas flexible synthesis improves fluency at the cost of faithfulness and predictability.
% strict synthesis preserves reasoning faithfulness and enables high predictability but can produce stilted prose, whereas flexible synthesis produces opposite effects.
Finally, the framework strictly supports single-turn interactions and does not support external tool-calls (e.g., RAG \citep{lewis_etal_2020_rag}, code execution \citep{karpas2022mrklsystemsmodularneurosymbolic}, etc.).
As a result, in tasks like argument generation, we cannot ground argumentative claims in retrieved context, nor validate whether claims are hallucinated.
We provide an expanded limitations discussion in Appendix~\ref{app:limitations}.

\section{Future Work}

\state's ability to balance diversity and quality (Section~\ref{sec:experiments-noveltybench}) and the associations we identify between controller actions and output performance (Section~\ref{sec:experiments-argument-generation}) motivate a shift from association toward explicit causal claims about how reasoning patterns shape downstream outcomes.
We can therefore model action trajectories as sequential treatments, and use randomized interventions to identify per-step causal effects (Appendix~\ref{app:future-work-causal}). 
We can then use this framework in large-scale studies that measure belief change and induced actions after exposure to generated arguments to study the causal effects of complex rhetorical strategies (Appendix~\ref{app:future-work-human}).

An equally important direction is optimizing \state itself.
First, replacing fixed beam search with more sophisticated tree-search methods such as Monte Carlo Tree Search \citep{kocsis_and_szepesvari_2006_mcts, coulom_2006_mcts, browne_etal_2012_mcts, hao-etal-2023-reasoning, silver2016mastering, silver_etal_2018_alphazero} could adapt exploration toward high-performing regions of the action space under constrained evaluation budgets (Appendix~\ref{app:future-work-search}).
Second, weight-based optimization via reinforcement learning (e.g., PPO \citep{schulman2017ppo}, GRPO \citep{shao2024deepseekmathpushinglimitsmathematical}) could train the controller, generator, or evaluator to improve downstream performance (Appendix~\ref{app:future-work-rl}).
Third, prompt-optimization pipelines like GEPA \citep{agrawal2025gepareflectivepromptevolution} could refine the instructions and demonstrations used by each module (Appendix~\ref{app:future-work-prompt-opt}).

\section{Ethical Implications}

Argument generation systems can be misused to manipulate at scale by generating misleading, deceitful, or otherwise harmful messages.
Prior work shows that LLM-generated arguments can affect human beliefs and preferences in public policy \citep{bai_voelkel_muldowney_eichstaedt_willer_2025, hackenburg_etal_2025_the_levers_of_political_persuasion}, support harmful narratives (e.g., conspiratorial content; \citealp{costello2026conspiracies}), coerce LLMs into performing harmful requests \citep{zeng-etal-2024-johnny}, and draft convincing phishing or social engineering messages \citep{qi_etal_2025_spearbot, lynch2025agenticmisalignmentllmsinsider}.
\state is particularly capable of such manipulation, since it can search over a diverse yet high-quality collection of arguments, and identify the one most likely to sway behavioral outcomes.
By interacting with or simulating an audience \citep{park_etal_2023_simulacra, park2024generativeagentsimulations1000}, \state can uncover the rhetorical patterns that systematically increase target audience susceptibility.
In adversarial hands, such micro-targeting \citep{salvi2024conversationalpersuasivenesslargelanguage} can potentially persuade people to vote against their interests, purchase unsuitable products, or adopt harmful beliefs.

However, persuasion is not inherently manipulative; it is the mechanism by which individuals and institutions communicate urgency, build trust, and motivate action.
In public health, well-intentioned guidance often fails to account for patients' specific fears or cultural context, and more tailored communication could improve adherence and outcomes \citep{brown2024chatbot, hou2025hpv}.
Improved persuasion can serve prosocial goals, from increasing vaccine uptake to encouraging charitable giving and democratic participation.

Persuasion attempts, both prosocial and adversarial, will only increase as LLMs become more widely available.
% By making the mechanisms that drive persuasion interpretable and measurable, \state can help identify argumentative patterns underlying rhetorical fallacies \citep{sourati_etal_2023_logical_fallacies, mouchel-etal-2025-logical}, misinformation, and manipulation.
Researchers can use \state to uncover argumentative patterns that are emotionally abusive or associated with misuse, and steer LLMs away from employing them.
Rigorous and transparent tools for analyzing persuasion are a prerequisite for defending against its misuse.

\section{Disclosure of LLM Use}

We used Claude, ChatGPT, and AI code-editors to assist in writing our LaTeX code for this paper.
We used LLMs to produce certain tables and figures, and then verified the values in these artifacts against the raw results we kept untouched. 
We used LLMs (GPT and Claude) to get feedback on drafts.
We used an LLM-as-a-Judge in our argument-generation experiments (Section \ref{sec:experiments-argument-generation}).

\section{Acknowledgments}

Funded by the European Union (ERC, Convey, 101078158). Views and opinions expressed are however those of the author(s) only and do not necessarily reflect those of the European Union or the European Research Council Executive Agency. Neither the European Union nor the granting authority can be held responsible for them.
This work was supported in part by the Israel Science Foundation (grant 3123/25).
We thank the Princeton Laboratory for Artificial Intelligence for providing computational resources and the Princeton Data-Driven Social Science Initiative for feedback and support.
We also thank OpenAI for providing additional resources through the Researcher Access Program.
We are grateful to Omar Khattab for his ongoing support of this project and for reviewing our paper at its early stages. 
We also thank Justin Grimmer, Yamil Velez, Allen Roushe, Devin Gonier, John Hines, Matthew Salganik, Susan Murphy, Dongling Zeng, Inbal Billie Nahum-Shani, and Queenie Luo for providing helpful comments and feedback.
Finally, we appreciate the discussions, support, and feedback of our colleagues at Princeton, Technion, and HUJI. 

\newpage

\bibliographystyle{colm2026_conference}
\bibliography{custom}
\newpage
\appendix

\section{Related Work}
\label{app:related-work}

\subsection{Social Science Experiments with Text}
\label{app:related-work-social-science}

Persuasion is central to human communication, spanning political discourse \citep{bai_voelkel_muldowney_eichstaedt_willer_2025, hackenburg_etal_2025_scaling_language_models_for_political_persuasion, hackenburg_etal_2025_the_levers_of_political_persuasion}, human-AI interaction \citep{salvi2024conversationalpersuasivenesslargelanguage, costello2026conspiracies, durmus2024persuasion}, and misinformation correction \citep{costello2025justthefacts, boissin2025conspiracybeliefs}.
Computational social science increasingly formalizes persuasion research by treating text as a treatment variable to study how linguistic features causally affect downstream behaviors \citep{grimmer2022textasdata, feder-etal-2022-causal}.
Traditional approaches focus on identifying content themes across document corpora and assessing how these themes affect outcomes \citep{fong-grimmer-2016-discovery, roberts2014structural}.
For example, \citet{saenger-etal-2024-autopersuade} use topic modeling to discover persuasive themes in argument collections, while \citet{egami_etal_2022_causal_inferences_using_texts} analyze how different framings affect bureaucratic responsiveness.
Recently, researchers have examined how conversations with LLMs affect beliefs \citep{costello_etal_2024_conspiracy_theories, costello2026conspiracies, salvi2024conversationalpersuasivenesslargelanguage}, identifying consistent patterns in effective messaging, such as emphasizing facts and evidence \citep{costello2025justthefacts}.

However, empirical methods in computational social science face limitations in studying fine-grained textual features.
Topic modeling approaches \citep{blei2012ProbablisticTopicModels, grimmer2022textasdata, saenger-etal-2024-autopersuade} naturally capture content themes but struggle with structural and stylistic variation.
These methods typically identify latent features \emph{ex-post} from existing corpora, constraining analysis to features already present in the data and making it difficult to systematically explore novel feature combinations.
Such text-as-treatment experiments ideally manipulate specific features, rhetorical structure \citep{stab-gurevych-2014-annotating, hidey-etal-2017-analyzing, chakrabarty-etal-2019-ampersand, wachsmuth-etal-2018-argumentation} (e.g., whether arguments begin with concessions or lead with strong claims), stylistic choices \citep{deri_etal_2018_coloring_the_links, wachsmuth-etal-2018-argumentation, persuasive_power_of_llms_2024, el-baff-etal-2024-improving} (e.g., formality, tone, pragmatic objective), and content themes, while maintaining coherence \citep{durmus-etal-2019-role} and logical soundness.
However, these features are difficult to control systematically in text generation \citep{saenger-etal-2024-autopersuade}, and are therefore rarely analyzed at scale.
Moreover, most prior work examines feature presence (whether a theme appears) rather than sequential ordering (when in a message a feature appears), limiting insights into how narrative structure affects argument quality. 
\state offers a framework through which to study the effects of granular decision sequences on downstream outcomes.

\subsection{Inference-Time-Compute}
\label{app:related-work-itc}

Inference-time compute (ITC) methods augment LLM generation by allocating additional computation \emph{after} training, either by extending the reasoning depth of individual trajectories or by generating many candidates and selecting among them.
These two axes, depth and breadth, are complementary, and many modern systems combine them.
The unifying motivation is the empirical finding that the quality of reasoning often scales with test-time computation even when the model weights are held fixed \citep{openai2024openaio1card, deepseekai2025deepseekr1incentivizingreasoningcapability, beeching2024scalingtesttimecompute}.
\state belongs to this family of methods and specifically extends Tree-of-Thoughts-style search with an \emph{explicit action space} over reasoning strategies.

\subsubsection{Depth-oriented ITC}
\label{app:related-work-itc-depth}

Chain-of-Thought (CoT) prompting \citep{wei2022chain, kojima_etal_2022_zero_shot_reasoners} scales reasoning \emph{depth} by eliciting intermediate steps before the final answer.
This seemingly simple change yields substantial gains on arithmetic, symbolic reasoning, and commonsense tasks, suggesting that the reasoning process itself carries value beyond the final token \citep{sprague2024cotcotchainofthoughthelps}.
The rationale behind CoT can be understood through the lens of hidden computation: additional tokens allow the model to perform iterative refinement that a single forward pass cannot \citep{pfau2024lets}.
 
Despite these benefits, CoT reasoning is not always faithful to the underlying inference process \citep{admoni2025largelanguagemodelsselfconsistent, anthropic_tracing_thoughts_2025, guan2025monitoringmonitorability}.
\citet{turpin2023language} demonstrate that CoT explanations are frequently post-hoc rationalizations: when models are biased toward incorrect answers through
prompt manipulation, they generate superficially coherent but misleading rationales, causing model performance to drop.
This brittleness raises serious concerns for settings where the chain of thought is meant to serve as an auditable record of model reasoning.
 
A separate line of work asks how to \emph{train} models to reason more effectively.
\citet{zelikman_etal_2022_star} introduce the Self-Taught Reasoner (STaR), which iteratively fine-tunes a model on its own correct rationales, bootstrapping reasoning capability without requiring large annotated rationale datasets.
Reinforcement Learning from Verifiable Rewards (RLVR) takes this further: rather than relying on human-curated signal, the model receives reward based on objective correctness criteria such as code compilation or arithmetic verification.
\citet{shao2024deepseekmathpushinglimitsmathematical} introduce Group Relative Policy Optimization (GRPO), a memory-efficient variant of PPO, and show that it substantially improves mathematical reasoning.
\citet{deepseekai2025deepseekr1incentivizingreasoningcapability} then demonstrate that pure RL training without supervised warm-start can induce emergent reasoning behaviors
such as self-reflection, backtracking, and extended chains of thought---matching OpenAI o1 on competitive mathematics benchmarks.
Similarly, \citet{openai2024openaio1card} and \citet{muennighoff2025s1} show that models explicitly optimized for long-horizon reasoning substantially amplify the benefits of depth-oriented ITC.
 
\state is complementary to this line of work.
Where depth-oriented methods focus on optimizing \emph{how long} a model reasons, \state focuses on \emph{what} it reasons about at each step.
By conditioning generation on explicit action templates, \state makes high-level decisions in a reasoning trajectory auditable and manipulable in a way that standard CoT, even when faithful, does not support.

\subsubsection{Breadth-oriented ITC}
\label{app:related-work-itc-breadth}

Breadth-oriented methods generate multiple candidate responses and select among them according to an external criterion, improving robustness \footnote{
    For example, the model may fail to create a valid generation due to a refusal, exceeding the context limit, or failing to adhere to a structured output schema.
} by reducing reliance on a single reasoning chain \citep{brown_etal_2020_gpt3, stiennon_etal_2020_learning_to_summarize}.
For example, Self-Consistency \citep{wang2022self, chen2024universal, taubenfeld-etal-2025-confidence} samples multiple candidate reasoning paths and then selects an answer by majority voting.
The central challenge is inducing \emph{meaningful diversity} across candidates rather than many near-duplicates of the same response.
 
The standard approach is high-temperature sampling, which expands the vocabulary distribution over the next token.
More principled truncation strategies have been proposed to improve the quality--diversity trade-off.
Nucleus (top-$p$) sampling \citep{Holtzman2020The} truncates the distribution to the smallest set of tokens whose cumulative probability exceeds $p$, preventing catastrophically low-probability tokens at modest quality cost.
Top-$k$ sampling \citep{fan-etal-2018-hierarchical} truncates to the top-$k$ tokens by probability mass, offering a simpler but less adaptive alternative.
More recently, min-$p$ sampling \citep{minh2025turning} introduces a dynamic threshold that scales the cutoff by the top token's probability, effectively widening the candidate
set when the model is uncertain and narrowing it when the model is confident.
These token-level strategies share a common limitation: they operate on the logit distribution at each decoding step and therefore do not control the \emph{semantic} content or rhetorical strategy of the generated response.
 
At higher levels of abstraction, prompt-based diversity methods attempt to elicit variation through the input rather than the decoding algorithm.
\citet{zhang2025verbalized_sampling} propose Verbalized Sampling (VS), a training-free prompting strategy that asks the model to jointly generate a set of responses and
verbalize a probability distribution over them.
By surfacing the model's internal uncertainty as explicit text, VS bypasses the typicality bias introduced by post-training alignment and recovers diversity that was suppressed
during RLHF.
VS achieves 1.6--2.1$\times$ diversity gains over direct prompting on creative writing tasks without sacrificing quality.
However, VS is fundamentally bounded by a single LLM generation: to produce $n$ diverse responses, the entire batch must be generated within one context window.
This constraint makes VS poorly suited to large $n$.
 
\state addresses the diversity bottleneck at a higher level of abstraction.
Rather than modifying the decoding algorithm or asking the model to self-sample a distribution, \state precomputes an explicit set of \emph{action templates}---discrete, interpretable specifications of rhetorical strategy---and uses a reranker controller to select top-$n$ distinct actions for each branching step.
This guarantees that each branch explores a semantically distinct region of the reasoning space without requiring high temperature or long-context self-sampling.
Diversity is therefore a structural property of the search procedure rather than a statistical side-effect of decoding.

\subsubsection{Tree-of-Thoughts}
\label{app:related-work-itc-tot}

Tree-of-Thoughts (ToT) \citep{yao2023tree} unifies depth and breadth by recasting LLM inference as search over a tree of partial reasoning states.
At each layer, the model branches into multiple candidate thoughts, an evaluator scores them, and low-value branches are pruned, preventing exponential growth and error
propagation.
\citet{hao-etal-2023-reasoning} formalize this connection by treating LLM inference as planning in a world model, while \citet{hao2024llm}, \citet{beeching2024scalingtesttimecompute}, and \citet{shalevshwartz2025reasoningsuperintelligencesearchtheoreticperspective} explore the performance of depth-first versus breadth-first search for complex reasoning tasks.
 
Several extensions enrich the basic ToT framework with more principled search algorithms. 
Monte Carlo Tree Search (MCTS) \citep{kocsis_and_szepesvari_2006_mcts, coulom_2006_mcts, browne_etal_2012_mcts} balances exploration and exploitation via upper-confidence bounds, enabling adaptive allocation of the evaluation budget toward high-value regions of the reasoning tree.
\citet{zhang2024restmcts} integrate MCTS with process reward models to guide search and simultaneously generate high-quality training data for policy and reward model improvement, outperforming both Best-of-$n$ and standard ToT under the same computation budget.
Chain of Preference Optimization \citep{zhang2024chain} uses the preference signal implicit in the ToT search tree---which branches were kept versus pruned---to fine-tune the model with DPO, achieving CoT-level inference cost at ToT-level quality.
These RL-flavored formulations are natural: ToT can be read as a form of tree-structured policy search, where each branching action is sampled from a policy $\pi_\theta$, intermediate states receive process rewards from a value function $V$, and the goal is to maximize the reward of the final leaf state \citep{schulman2017ppo, shao2024deepseekmathpushinglimitsmathematical}.
 
Despite this expressiveness, standard ToT implementations share two important limitations that \state addresses.
First, existing methods rely exclusively on stochastic temperature sampling to differentiate branches.
Because sampling operates at the token level, branches in the same tree often converge on semantically similar content \citep{jiang2025artificial, zhang2025noveltybenchevaluatinglanguagemodels}, undermining the exploration benefit that motivates tree search in the first place.
Second, because reasoning decisions are implicit in the decoding process, it is difficult to attribute differences in output quality to specific choices made at specific reasoning steps.
This opacity limits the interpretive utility of ToT: researchers can observe that some trajectories outperform others, but not \emph{why}.
 
\state replaces token-level stochasticity with an explicit, structured action space, making every branching decision both interpretable and auditable.
The controller selects a named action from a fixed vocabulary---specifying, for example, which rhetorical structure and content theme to employ at each step---and the generator prefills that action as a textual intervention before sampling the continuation.
This design decouples \emph{what to reason about} (the action) from \emph{how to express it} (the generated token sequence), a separation that standard
ToT conflates.
As a result, each path through the \state tree corresponds to a logged, human-readable action sequence that can be subjected to formal attribution analysis.

\newpage

\section{DSPy Background}
\label{app:dspy-background}

\state is built on \textbf{DSPy} \citep{khattab2023dspy}, which provides a modular, declarative approach to programming LLMs.
DSPy separates \emph{what} a task does (expressed as a \textbf{Signature}) from \emph{how} it is executed (determined by a \textbf{Module}, \textbf{Adapter}, and \textbf{Language Model}).
This separation of concerns makes components independently configurable and composable, and enables automatic prompt optimization without manual re-engineering of prompts.

\subsection{Core DSPy Primitives}
\label{app:dspy-primitives}

\paragraph{Fields.}
The fundamental building blocks in DSPy are \emph{Fields}, which define the input/output schema of a task through typed annotations and natural-language descriptions.
\texttt{InputField} objects describe the variables a module expects; \texttt{OutputField} objects specify what the module should produce.
For example:

\begin{lstlisting}[style=dspypython]
topic: str = InputField(desc='Debate topic')
stance: Literal['PRO', 'ANTI'] = InputField(desc='Stance to argue')
argument: str = OutputField(desc='Generated argument')
\end{lstlisting}

\paragraph{Signatures.}
A \emph{Signature} is a declarative task specification: it bundles a set of Fields together with a task-level docstring instruction, defining \emph{what} the module should do without specifying any prompt template.
Signatures are defined as Python classes that subclass \texttt{dspy.Signature}:

\begin{lstlisting}[style=dspypython]
class GenerateArgument(dspy.Signature):
    '''Generate an argument for the given topic and stance.'''
    topic: str = dspy.InputField(desc='Debate topic')
    stance: Literal['PRO', 'ANTI'] = dspy.InputField(desc='Stance to argue')
    argument: str = dspy.OutputField(desc='Generated argument')
\end{lstlisting}

\paragraph{Modules.}
A \emph{Module} is a parameterized layer that executes a Signature.
The basic DSPy module \texttt{dspy.Predict} takes a Signature and, at inference time, calls the configured Language Model to produce predictions.
Modules are composable: larger pipelines can be assembled from multiple modules, each handling a distinct subtask (e.g., planning, generation, evaluation).

\paragraph{Adapters.}
An \emph{Adapter} bridges a Signature and a Language Model by formatting the inputs and the signature's instruction into a concrete prompt string, and by parsing the LLM's raw textual response back into typed, structured field values.
DSPy ships with several built-in adapters (e.g., \texttt{ChatAdapter}, \texttt{JSONAdapter}); \state uses custom adapters (\texttt{LocalVLLMAdapter} for generative models and \texttt{LocalVLLMScoringAdapter} for reranker models) that extend these to support tool-call formatting, prefill injection, and query--document scoring.
After the Adapter parses the output, it type-checks each \texttt{OutputField} value---raising an error if the response does not conform to the declared type (e.g., a \texttt{Literal} constraint).

\subsection{Instantiation and Forward Pass}
\label{app:dspy-instantiation}

\paragraph{Instantiation phase.}
A Module is created by combining a \textbf{Signature} (task definition: \emph{what} to do) with a \textbf{Language Model} (executes prompts: \emph{which} LLM) and an \textbf{Adapter} (formats prompts, parses outputs).

\begin{figure}[h!]
\centering
\begin{tikzpicture}[>=stealth, font=\small, node distance=1.2cm]
  \node[draw, rounded corners=4pt, fill=yellow!25, minimum width=3.6cm,
        minimum height=1.2cm, align=center] (sig)
        {\textbf{Signature}\\task definition: \emph{what to do}};
  \node[draw, rounded corners=4pt, fill=blue!15, minimum width=3.6cm,
        minimum height=1.2cm, align=center, right=1.8cm of sig] (lm)
        {\textbf{Language Model}\\executes prompts};
  \node[draw, rounded corners=4pt, fill=green!20, minimum width=3.6cm,
        minimum height=1.2cm, align=center, right=1.8cm of lm] (adapter)
        {\textbf{Adapter}\\formats prompts \&\\parses outputs};
  \node[draw, rounded corners=4pt, fill=gray!15, minimum width=3.6cm,
        minimum height=1.2cm, align=center, below=1.5cm of lm] (module)
        {\textbf{dspy.Module}};

  \draw[->, thick] (sig.south) .. controls +(0,-0.8) and +(-1.2,0) .. (module.west);
  \draw[->, thick] (lm.south) -- (module.north);
  \draw[->, thick] (adapter.south) .. controls +(0,-0.8) and +(1.2,0) .. (module.east);
\end{tikzpicture}
\caption{A \texttt{dspy.Module} is assembled from a Signature (task specification), a Language Model (inference backend), and an Adapter (prompt formatting and output parsing).}
\label{fig:dspy-instantiation}
\end{figure}

\paragraph{Forward (inference) phase.}
When a Module is called with an \emph{Example} (a dictionary of input field values matching the Signature), the pipeline proceeds as follows:
(1) the Adapter formats the Signature instruction, field descriptions, and input values into a prompt string;
(2) the Language Model generates a textual response;
(3) the Adapter extracts and type-checks values for each \texttt{OutputField} from the response, returning a \textbf{Prediction} object with structured field values.

\begin{figure}[h!]
\centering
\begin{tikzpicture}[>=stealth, font=\small, node distance=0.8cm and 2.6cm]
  \node[draw, rounded corners=4pt, fill=gray!10,
        minimum width=2.8cm, minimum height=0.9cm, align=center] (user) {User / Caller};

  \node[draw, rounded corners=4pt, fill=gray!15,
        minimum width=2.8cm, minimum height=0.9cm, align=center,
        right=2.8cm of user] (module) {\textbf{dspy.Module}};

  \node[draw, rounded corners=4pt, fill=green!20,
        minimum width=2.8cm, minimum height=0.9cm, align=center,
        below=1.1cm of module] (adapter) {\textbf{Adapter}};

  \node[draw, rounded corners=4pt, fill=blue!15,
        minimum width=2.8cm, minimum height=0.9cm, align=center,
        below=1.1cm of adapter] (lm) {\textbf{Language Model}};

  \draw[->, thick]
    (user.east) -- 
    node[above, font=\scriptsize, align=center, text width=2.4cm, fill=white, inner sep=1pt]
    {Example\\(topic=\ldots, stance=\ldots)}
    (module.west);

  \draw[->, thick]
    (module.south) -- 
    node[right, font=\scriptsize, fill=white, inner sep=1pt]
    {Sig.\ + inputs}
    (adapter.north);

  \draw[->, thick]
    (adapter.south) -- 
    node[right, font=\scriptsize, fill=white, inner sep=1pt]
    {formatted prompt}
    (lm.north);

  \draw[->, thick]
    (lm.west) .. controls +(-1,0) and +(-1,0) ..
    node[left, font=\scriptsize, fill=white, inner sep=1pt]
    {raw response}
    (adapter.west);

  \draw[->, thick]
    (adapter.west) .. controls +(-1,0) and +(-1,0) ..
    node[left, font=\scriptsize, fill=white, inner sep=1pt]
    {parsed fields}
    (module.west);

  \draw[->, thick]
    (module.north) .. controls +(0,0.6) and +(0,0.6) ..
    node[above, font=\scriptsize, align=center, text width=2.2cm, fill=white, inner sep=1pt]
    {Prediction\\(argument=\ldots)}
    (user.north);
\end{tikzpicture}
\caption{The Module delegates prompt formatting to the Adapter, which calls the Language Model and parses its response into a Prediction.}
\label{fig:dspy-forward}
\end{figure}

\subsection{Prompt Optimization with DSPy}
\label{app:dspy-optimization}

A key advantage of the DSPy abstraction is that, because all prompt logic is encapsulated inside Signatures and Adapters, the \emph{textual content} of prompts can be treated as learnable parameters and optimized automatically.
DSPy optimizers, like the two examples below, search over the space of candidate prompts and select those that maximize a user-defined metric on a development set.

\textbf{MIPROv2 \citep{opsahl-ong-etal-2024-optimizing}.}
MIPROv2 (Multi-prompt Instruction PRoposal Optimizer v2) is a Bayesian optimization-based prompt optimizer.
It jointly optimizes two components of each DSPy module:
(i)~\textbf{Signature instructions}---the free-text preamble in the system message that describes the task to the LLM;
and (ii)~\textbf{few-shot demonstrations} (``demos'')---a small set of example (input, output) pairs prepended to the user message to guide in-context learning.
MIPROv2 uses a surrogate model over the candidate-prompt space to propose high-quality instruction rewrites and select informative demo subsets, significantly reducing the number of LLM calls required compared to brute-force search.

\textbf{GEPA \citep{agrawal2025gepareflectivepromptevolution}.}
GEPA (\textbf{Ge}netic \textbf{Pa}reto) frames prompt optimization as a reflective evolution process.
An LLM ``editor'' is given the current instruction, recent failed examples (coupled with their scores and textual feedback), and is asked to propose revised instructions.
These proposals are evaluated on a held-out set, and the best-performing candidates are carried forward.
GEPA complements MIPROv2 by enabling more targeted, semantically-informed instruction rewrites, and is particularly well-suited to iterative refinement when the optimization landscape is smooth.

\newpage

\section{\state Modules}
\label{app:state-modules}

\begin{figure}[ht]
  \centering
  \resizebox{\textwidth}{!}{%
    \begin{tikzpicture}[
        >=stealth,
        font=\footnotesize,
        box/.style={draw=borderblue, rounded corners=4pt, align=center,
                    text width=4.2cm, minimum height=2cm, inner sep=6pt,
                    fill=white, line width=0.8pt},
        band/.style={inner sep=6pt, rounded corners=6pt},
        arr/.style={->, thick, color=borderblue},
        lbl/.style={font=\scriptsize, midway}
      ]
      \node[box] (plan) at (0,0)
        {\textbf{Plan}\\Controller $C$\\selects actions from $\mathcal{A}$};
      \node[box] (gen)  at (5.4,0)
        {\textbf{Generate}\\Generator $G$ produces\\candidates conditioned on actions};
      \node[box] (eval) at (10.8,0)
        {\textbf{Evaluate}\\Evaluator $V$\\scores candidates};
      \node[box] (sel)  at (16.2,0)
        {\textbf{Select}\\Beam Search\\keeps top-$k$ candidates};

      \begin{pgfonlayer}{background}
        \node[band, fit=(plan),  fill=bandblue]   {};
        \node[band, fit=(gen),   fill=bandgreen]  {};
        \node[band, fit=(eval),  fill=bandorange] {};
        \node[band, fit=(sel),   fill=bandred]    {};
      \end{pgfonlayer}

      \draw[arr] (plan.east) -- (gen.west);
      \draw[arr] (gen.east)  -- (eval.west);
      \draw[arr] (eval.east) -- (sel.west);
      \draw[arr] (sel.south) .. controls +(0,-1.4) and +(0,-1.4) ..
        node[below, yshift=-2pt]{\small next layer} (plan.south);
    \end{tikzpicture}
    }
  \caption{ \state module loop used in tree search: controller planning, candidate generation, evaluation, and beam selection, followed by expansion to the next layer.}
\end{figure}

\begin{figure}[h!]
  \centering
  \includegraphics[width=\textwidth]{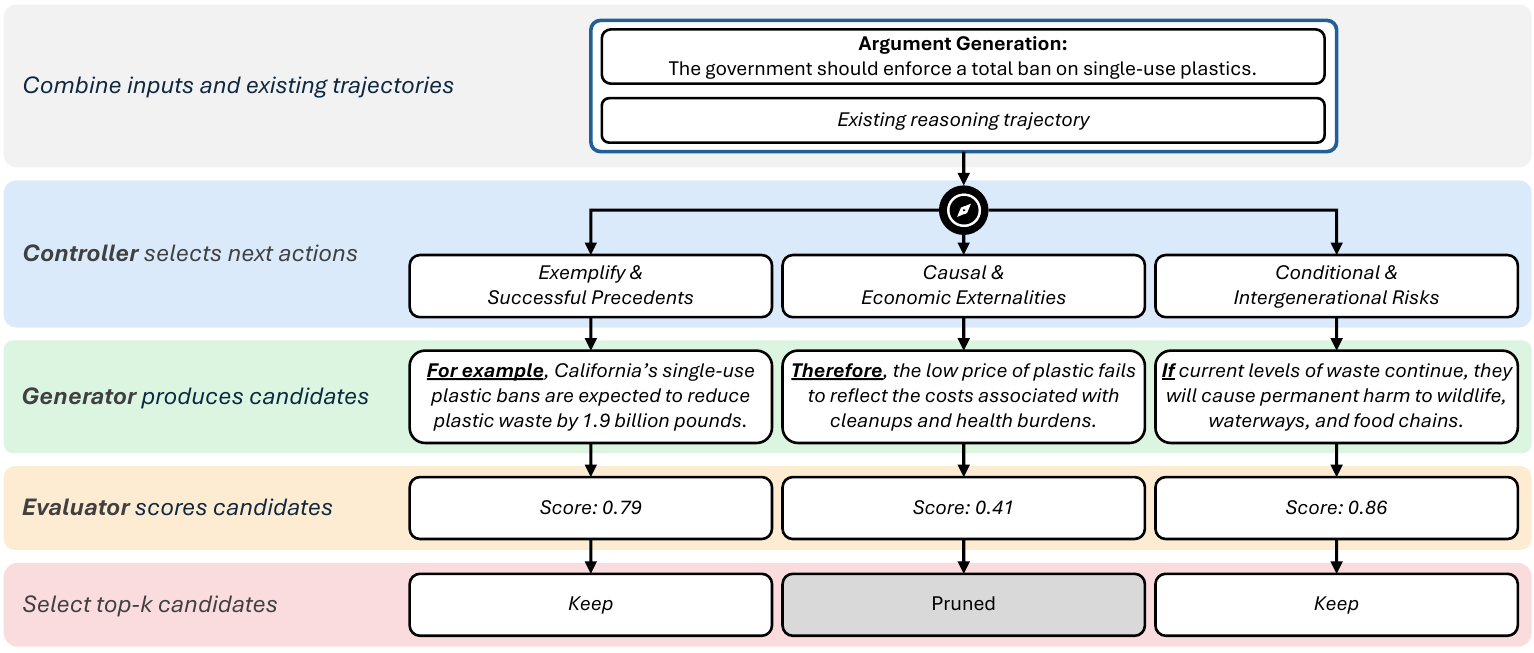}
  \caption{Stylized example of \state's Plan$\rightarrow$Generate$\rightarrow$Evaluate$\rightarrow$Select loop.
  The controller plans which actions to explore, the generator expands candidate trajectories, the evaluator scores them, and beam selection retains the top-$k$ states.}
  \label{fig:architecture-overview}
  \end{figure}

\state's Controller, Generator, and Evaluator (Section \ref{sec:methods-tree-of-thoughts-components}) are implemented with DSPy \citep{khattab2023dspy} Modules (Appendix \ref{app:dspy-background}).
Figure~\ref{fig:architecture-overview} illustrates how these components interlock across layers of the search tree.
At each layer $i$, the \textbf{controller} observes the current beam of states and, for each state $s_{i-1}$, selects up to $n$ actions from the action space $\mathcal{A}$.
Each selected action is executed to produce a \texttt{ReasoningIntervention} (a prefix and context guidance), which the \textbf{generator} injects into the LLM's assistant message as a prefill before sampling a continuation.
This produces a set of candidate child states; each child is either an intermediate reasoning state $s_i = [s_{i-1}, z_i]$ or a final-answer state $s_i = [s_{i-1}, y]$, depending on whether the controller signalled early stopping.
Intermediate children are passed to the \textbf{evaluator}, which scores them with a Process Reward Model $V_\text{PRM}$; final-answer children are scored with an Outcome Reward Model $V_\text{ORM}$ and collected for the terminal selection step.
\textbf{Beam selection} then retains the top-$k$ intermediate states by score, forming the beam for layer $i{+}1$.
This Plan$\rightarrow$Generate$\rightarrow$Evaluate$\rightarrow$Select cycle repeats until the maximum depth $d$ is reached or all active branches have terminated via early stopping, at which point the highest-scoring final-answer state is returned.
In practice, LLM calls from all components are parallelized across all nodes in a given layer using vLLM \citep{kwon2023efficient}, significantly reducing end-to-end latency compared to sequential decoding.

\begin{table}[h!]
  \centering
  \small
  \setlength{\tabcolsep}{5pt}
  \begin{tabular}{p{0.31\linewidth} p{0.31\linewidth} p{0.31\linewidth}}
  \toprule
  \textbf{First step} & \textbf{Intermediate step} & \textbf{Final step} \\
  \midrule
  \footnotesize\ttfamily
  \textless thinking\textgreater\newline
  \textless step\textgreater\newline
    \#\# internal\_reasoning\newline
  \internalr{I should evaluate how plastics persist for centuries in the environment, meaning to- day's convenience imposes long-term burdens on future generations who had no say in their creation.}\newline
  \#\# claim\newline
  \prefixex{If} \cont{current levels of plastic waste continue, they will cause permanent harm to marine ecosystems\ldots}\newline
  &
  \footnotesize\ttfamily
  \textless thinking\textgreater\newline
  \textless step\textgreater\newline
  \#\# internal\_reasoning\newline
  \internalr{I should identify risks\ldots}\newline
  \#\# claim\newline
  \prefixex{If} \cont{current levels of plastic waste continue\ldots}\newline
  \textless/step\textgreater\newline
  \ldots\newline
  \textless step\textgreater\newline
  \#\# internal\_reasoning\newline
  \internalr{I should examine case studies from Rwanda, the EU, Kenya, and various US states showing that bans are enforceable and produce measurable reductions in pollution.}\newline
  \#\# claim\newline
  \prefixex{For example,} \cont{California's single-use plastic bans are expected to reduce plastic waste by 1.9 billion pounds\ldots}\newline
  &
  \footnotesize\ttfamily
  \textless thinking\textgreater\newline
  \textless step\textgreater\newline
  \#\# internal\_reasoning\newline
  \internalr{I should identify risks\ldots}\newline
  \#\# claim\newline
  \prefixex{If} \cont{current levels of plastic waste continue\ldots}\newline
  \textless/step\textgreater\newline
  \ldots\newline
  \textless step\textgreater\newline
  \#\# internal\_reasoning\newline
  \internalr{I should evaluate\ldots}\newline
  \#\# claim\newline
  \prefixex{For example,} \cont{California's single-use plastic ban\ldots}\newline
  \textless/step\textgreater\newline
  \ldots\newline
  \textless/thinking\textgreater\newline
  \textless answer\textgreater\\
  \bottomrule
  \end{tabular}
  \caption{Illustrative interventions for argument generation example in favor of single-use plastics ban.
  Templates are in black, \internalr{internal reasoning} in teal, \prefixex{prefixes} in blue, and the \cont{model continuation} in orange.
  Each column shows the generation state at different stages: first step (single claim), intermediate (multiple claims), and final (complete reasoning with answer delimiter). The action space used here is presented in Appendix~\ref{app:action-spaces-argument-generation}.}
  \label{tab:prefix-examples}
\end{table}

\newpage

\subsection{Controller}
\label{app:state-modules-controller}

We treat each action as a \emph{tool call}.
Selecting an action corresponds to choosing a tool name from a fixed set of action templates (Appendix \ref{app:action-spaces}) and providing values for the tool's arguments.
Executing the tool returns a structured intervention, i.e., a prefix and internal reasoning guidance, that is injected into the next generation step.
In our argument generation example (Figure~\ref{fig:workflow} and Figure \ref{fig:architecture-overview}), the actions encode \emph{structure} and \emph{content} dimensions, per \citet{wachsmuth-etal-2018-argumentation}.
Each argument in a tool has a fixed set of choices, which are dictated by an action-space's ``dimensions'' (Appendix~\ref{app:action-space-guidance}).
This mirrors the iterative tool-use paradigm in ReAct \citep{yao2023react}, with two key differences:
(i) we allow \emph{branching} by selecting multiple tools per input state, and (ii) our tools are lightweight prefix interventions rather than external capabilities such as retrieval or code execution.
The controller's action selection process supports \emph{early stopping}: if the controller determines that reasoning is sufficient, it selects a dedicated \texttt{FINISH} action, signifying that the generator should produce a final answer.\footnote{
    This is represented in our controller as an additional tool that takes no arguments.
}
We implement two kinds of controllers: one that uses a generative LLM to produce tool calls (Appendix~\ref{app:state-modules-generative-controller}), and another that uses a reranker LLM \citep{qwen3embedding} to pick among all possible tool calls (Appendix~\ref{app:state-modules-reranker-controller}).

The performance of a controller matters most when the action space for the task is large, and relatively few actions from the action space would enable a high-quality final response.
For example, in a math task, where the action space consists of dozens or hundreds of math operations, but less than a handful are valid or productive, a poor-quality controller would steer generation towards flawed responses.
On the other hand, open-ended tasks like argument generation, where many actions from the action space could yield high-quality responses, are less-dependent on the controller's quality.
Our experiments in this paper focus on open-ended tasks with relatively small action-spaces. 
There are relatively few cases in which a sequence of choices by the controller would explicitly steer the response in a poor direction (e.g., starting with an example or specification in argument generation).

\subsubsection{Generative controller}
\label{app:state-modules-generative-controller}

The generative controller prompts an LLM to \emph{propose} a tool call \citep{yao2023react, karpas2022mrklsystemsmodularneurosymbolic} from action space $\mathcal{A}$ conditioned on the current state: it chooses which tool to use (which action template) and provides values for the tool's arguments (permitted choices are specified through a \texttt{Literal} type). 
In this setup, the same generator model that produces thoughts can also decide what to do next. 
A key advantage is that the controller can utilize ITC methods like CoT, and produce \emph{natural-language rationales} for why a tool call is appropriate given the chain so far.\footnote{
    Preceding tool call outputs with reasoning enables more informed decision-making and easier debugging.
} 
A key limitation is \emph{low action diversity}: because tool-call generation is itself sampled from an LLM, the controller can collapse to repeatedly predicting the same high-probability action, reducing the benefit of branching.
While verbalized sampling \citep{zhang2025verbalized_sampling} can help if $n$ (the number of actions the controller must select) is small, it is not effective if $n$ is large (Appendix \ref{app:related-work-itc-breadth}). 

\begin{figure}[ht]
  \centering
  \begin{tikzpicture}[
      >=stealth,
      font=\footnotesize,
      node distance=1.3cm,
      box/.style={draw=borderblue, rounded corners=4pt, align=center,
                  text width=4.2cm, inner sep=6pt, fill=white, line width=0.8pt},
      wide/.style={draw=borderblue, rounded corners=4pt, align=center,
                   text width=5cm, inner sep=6pt, fill=white, line width=0.8pt},
      arr/.style={->, thick, color=borderblue},
      lbl/.style={font=\scriptsize, midway, color=black}
    ]

    % Nodes first (to compute positions for bands)
    \node[box]  (state)  {State\\(input + reasoning)};
    \node[box, right=2cm of state] (aspace) {Action Space\\(tool definitions)};

    \node[wide, below=1.8cm of $(state.south)!0.5!(aspace.south)$]
      (prompt) {Controller prompt};
    \node[wide, below=of prompt]
      (llm)    {Generative LLM};
    \node[wide, below=of llm]
      (tool)   {Tool Call\\(name + arguments)};
    \node[wide, below=of tool]
      (interv) {Interventions\\internal reasoning + prefix};

    % Background bands (drawn behind nodes)
    \begin{pgfonlayer}{background}
      % Controller band: prompt, llm, tool (not state/aspace — those are inputs)
      \node[fit=(prompt)(tool), inner sep=12pt,
            fill=bandblue, rounded corners=8pt] {};
      % Output band: interventions (feeds generator)
      \node[fit=(interv), inner sep=12pt,
            fill=bandgreen, rounded corners=8pt] {};
    \end{pgfonlayer}

    % Arrows
    \coordinate (mergeY) at ($(state.south)!0.5!(prompt.north)$);
    \draw[arr] (state.south) -- (state.south |- mergeY) -| (prompt.north);
    \draw[-,  thick, color=borderblue] (aspace.south) -- (aspace.south |- mergeY) -| (prompt.north);

    \draw[arr] (prompt.south) -- node[lbl, right]{call LLM} (llm.north);
    \draw[arr] (llm.south)    -- node[lbl, right]{parse}    (tool.north);
    \draw[arr] (tool.south)   -- node[lbl, right]{execute}  (interv.north);

  \end{tikzpicture}
  \caption{ Generative controller pipeline. Blue band = controller, green band = output (feeds generator).}
\end{figure}

\begin{tcolorbox}[
  after skip=2pt,
  colback=gray!8, colframe=gray!55, arc=4pt,
  title={\small\bfseries\sffamily System Message},
  fonttitle=\color{white}, coltitle=white, colbacktitle=gray!55,
  left=6pt, right=6pt, top=4pt, bottom=4pt,
]
\begin{lstlisting}[
  basicstyle=\ttfamily\scriptsize,
  breaklines=true,
  moredelim={**[is][\color{teal}]{@T}{T@}},
  moredelim={**[is][\color{blue}]{@B}{B@}},
]
@TGenerate an argument which takes the provided stance towards the provided topic.T@

You are given `@TtopicT@` and `@TstanceT@` and your goal is to finish with `@TargumentT@`.
To accomplish this goal, you will need to reason about the problem step by step
rather than generating `@TargumentT@` directly.
You have up to `number_of_additional_reasoning_steps` additional steps to reason
about the problem before generating `@TargumentT@`.
Refer to the existing reasoning steps under the `reasoning` header to inform
your next step. Reasoning steps are ordered sequentially, and each one includes
a `@TclaimT@` header above the content of the step itself.
Choose a tool to use from the following options:
(1) `@Tintervene_on_next_reasoning_stepT@`:
        * Description:
                @TDetermine the best choice in each dimension to improve the quality
                of the next reasoning step. You must select one choice for each
                of the provided dimensions.T@
        * Arguments:
                - @Tcausal_stylesT@: Literal['@Tfigurative_language,
                  statistical_and_data_driven, narrative_and_anecdote,
                  expert_and_authoritative_voice, repetition_and_parallelism,
                  contrast_and_antithesis, measured_and_authoritative_tone,
                  passionate_and_urgent_tone, direct_engagement, scope_and_framingT@']
                        @TForces the next reasoning step to adopt a specific rhetorical
                        style or expressive technique.T@
                - @Tcausal_subtopicsT@: Literal['@Tcost_benefit_and_impact_analysis,
                  rights_and_liberties, justice_and_fairness, ethical_principles,
                  governance_and_accountability, risk_and_unintended_consequences,
                  feasibility_and_implementation, incentives_and_power_dynamics,
                  precedent_and_long_term_effects, stakeholder_responsibilityT@']
                        @TForces the next reasoning step to analyze the issue through
                        a specific argumentative lens or topical framework.T@
(2) `finish`:
        * Description:
                Signals that the reasoning so far is sufficient for producing a
                high-quality response. If selected, the generator will produce
                the final output rather than reasoning further.

Your inputs will be:
1. `@TtopicT@` (str): @TThe topic to generate an argument aboutT@
2. `@TstanceT@` (Literal['PRO', 'ANTI']): @TThe stance to take on the topicT@
3. `reasoning` (str): @TThe existing reasoning steps towards producing `@TargumentT@`.
   Each step's content is under the `@TclaimT@` header.T@
4. `number_of_additional_reasoning_steps` (int): @TThe maximum number of additional
   reasoning steps you can take before you must produce `@TargumentT@`.T@

Your goal is to produce the following outputs:
1. `action` (str): The selected action (tool) to guide the next reasoning step.
2. `action_arguments` (dict[str, Any]): The input arguments for the selected action.

Please provide your response with each output field under its own header using
the format:
## field_name
Your response for that field here

NOTE: Line breaks, capitalization, and punctuation (i.e., '##' followed by a
space) are important. If you do not follow these guidelines, your response will
be rejected.
\end{lstlisting}
\end{tcolorbox}
\promptcaptionnote{lst:gen-controller-system}{Generative Controller}{\textcolor{teal}{Teal}: task signature and action-space configuration. \textcolor{blue}{Blue}: concrete runtime value.}
\begin{tcolorbox}[
  after skip=2pt,
  colback=gray!8, colframe=gray!55, arc=4pt,
  title={\small\bfseries\sffamily User Message},
  fonttitle=\color{white}, coltitle=white, colbacktitle=gray!55,
  left=6pt, right=6pt, top=4pt, bottom=4pt,
]
\begin{lstlisting}[
  basicstyle=\ttfamily\scriptsize,
  breaklines=true,
  moredelim={**[is][\color{teal}]{@T}{T@}},
  moredelim={**[is][\color{blue}]{@B}{B@}},
]
## @TtopicT@
@BShould single-use plastics be banned?B@

## @TstanceT@
@BPROB@

## reasoning
@B<thinking>
<step>
## internal_reasoning
Opening with env harm.
## claim
Single-use plastics are a leading contributor to ocean pollution.
</step>
</thinking>B@

## number_of_additional_reasoning_steps
@B2B@

Respond with the corresponding output fields, starting with the field
`## @TactionT@`, then `## @Taction_argumentsT@` (must be formatted as a valid
Python dict[str, Any])
\end{lstlisting}
\end{tcolorbox}
\promptcaption{lst:gen-controller-user}{Generative Controller User Message}

\newpage
\subsubsection{Reranker controller}
\label{app:state-modules-reranker-controller}

To introduce reliable diversity, we instead use a discriminative reranker controller that scores
\emph{all} candidate action-argument combinations and selects the top-$n$. 
We formulate action selection by measuring the relevance of a query and a document using a cross-encoder \citep{reimers-gurevych-2019-sentence,thakur2021beir,santhanam-etal-2022-colbertv2,laban-etal-2022-summac,sun-etal-2023-chatgpt,qwen3embedding}: 
the \emph{query} contains the input $x$ and the reasoning chain $Z_i$, while each \emph{document} is a description of the effect of the given tool and parameter (e.g., ``introduce a new claim that expands on financial impacts''). 
A reranker assigns a relevance score to each document, yielding a distribution over actions, and we take the top-$n$ scoring actions for expansion. 
This design supports diverse branching (by selecting different high-scoring actions) and enables efficient enumeration when $\mathcal{A}$ is finite and structured.

\begin{figure}[ht]
  \centering
  \begin{tikzpicture}[
      >=stealth,
      font=\footnotesize,
      node distance=1.3cm,
      box/.style={draw=borderblue, rounded corners=4pt, align=center,
                  text width=4.2cm, inner sep=6pt, fill=white, line width=0.8pt},
      wide/.style={draw=borderblue, rounded corners=4pt, align=center,
                   text width=5cm, inner sep=6pt, fill=white, line width=0.8pt},
      arr/.style={->, thick, color=borderblue},
      lbl/.style={font=\scriptsize, midway, color=black}
    ]

    % Nodes
    \node[box]  (state)  {State\\(input + reasoning)};
    \node[box, right=2cm of state] (aspace) {Action Space\\(all combinations)};

    \node[box, below=of state]  (query) {Query};
    \node[box, below=of aspace] (docs)  {Documents\\(one per combination)};

    \node[wide, below=1.8cm of $(query.south)!0.5!(docs.south)$]
      (scoring) {Reranker LLM\\score each (query, doc)};
    \node[wide, below=of scoring]
      (topn)    {Top-$n$ actions\\(sorted by score)};
    \node[wide, below=of topn]
      (interv)  {Interventions\\internal reasoning + prefix};

    % Background bands
    \begin{pgfonlayer}{background}
      % Controller band: query, docs, scoring, topn (not state/aspace — those are inputs)
      \node[fit=(query)(docs)(topn), inner sep=12pt,
            fill=bandblue, rounded corners=8pt] {};
      % Output band: interventions (feeds generator)
      \node[fit=(interv), inner sep=12pt,
            fill=bandgreen, rounded corners=8pt] {};
    \end{pgfonlayer}

    % Arrows
    \draw[arr] (state.south)  -- (query.north);
    \draw[arr] (aspace.south) -- (docs.north);

    \coordinate (mergeY) at ($(query.south)!0.5!(scoring.north)$);
    \draw[arr] (query.south) -- (query.south |- mergeY) -| (scoring.north);
    \draw[-,  thick, color=borderblue] (docs.south) -- (docs.south |- mergeY) -| (scoring.north);

    \draw[arr] (scoring.south) -- node[lbl, right]{top-$n$}       (topn.north);
    \draw[arr] (topn.south)    -- node[lbl, right]{execute each}   (interv.north);

  \end{tikzpicture}
  \caption{ Reranker controller pipeline. Blue band = controller, green band = output (feeds generator).}
\end{figure}

\begin{tcolorbox}[
  after skip=2pt,
  colback=gray!8, colframe=gray!55, arc=4pt,
  title={\small\bfseries\sffamily System Message},
  fonttitle=\color{white}, coltitle=white, colbacktitle=gray!55,
  left=6pt, right=6pt, top=4pt, bottom=4pt,
]
\begin{lstlisting}[
  basicstyle=\ttfamily\scriptsize,
  breaklines=true,
]
Judge whether the Document meets the requirements based on the Query and the
Instruct provided. Note that the answer can only be `yes' or `no'.
\end{lstlisting}
\end{tcolorbox}
\promptcaptionnote{lst:reranker-controller-system}{Reranker Controller}{Each action is scored as a query--document pair \citep{qwen3embedding}. The system message constrains output to ``yes''/``no''; the user message carries the instruction, query (inputs + reasoning), and document (action description). \textcolor{teal}{Teal}: task signature. \textcolor{blue}{Blue}: concrete runtime value.}
\begin{tcolorbox}[
  after skip=2pt,
  colback=gray!8, colframe=gray!55, arc=4pt,
  title={\small\bfseries\sffamily User Message},
  fonttitle=\color{white}, coltitle=white, colbacktitle=gray!55,
  left=6pt, right=6pt, top=4pt, bottom=4pt,
]
\begin{lstlisting}[
  basicstyle=\ttfamily\scriptsize,
  breaklines=true,
  moredelim={**[is][\color{teal}]{@T}{T@}},
  moredelim={**[is][\color{blue}]{@B}{B@}},
  escapeinside={(*@}{@*)},
]
<Instruct>: Your objective is to decide what action to take for the next
reasoning step for a user-assigned task.

The user provided the following task:
(*@\char34@*)@TGenerate an argument which takes the provided stance towards the provided topic.T@(*@\char34@*)

This task requires taking a sequence of reasoning steps to reach a solution.
You must determine what action to take next.
You will find the inputs for this task under the (*@\#@*) Inputs header in the Query.
You will find the intermediate reasoning trajectory towards solving this problem
under the (*@\#@*) Reasoning header in the Query.
You will find the action under consideration under the (*@\#@*) Action heading in the
Document.

Judge whether the provided action is likely to be a good next step for
addressing the user's task.
NOTE: You have @B2B@ actions remaining before you must return a final answer.

<Query>: (*@\#@*) Inputs
@Btopic: Should single-use plastics be banned?B@
@Bstance: PROB@

# Reasoning
@B<thinking>\n<step>\n## claim\nSingle-use plastics are a leading contributor...\n</step>\n</thinking>B@

<Document>: (*@\#@*) Action
@Bstructure: causal_reasoning | subtopic: environmental_impactB@
@BForces the next step to state causes, effects, and consequences.B@
@BI should analyze environmental causes and cascading ecological effects.B@
\end{lstlisting}
\end{tcolorbox}
\promptcaption{lst:reranker-controller-user}{Reranker Controller User Message}

\subsection{Generator}
\label{app:state-modules-generator}

The Generator expands the search tree by producing candidate reasoning steps or final answers.
Responses are structured using XML tags that create natural stopping points: each intermediate reasoning step is enclosed in \texttt{<step>...</step>}, and the final answer in \texttt{<answer>...</answer>}.
Generation is terminated via \emph{stop tokens}: \texttt{</step>} for intermediate steps and \texttt{</answer>} for final answers.
The choice of stop token depends on whether the controller has signalled to continue reasoning (\texttt{continue\_reasoning=True}) or to produce a final answer (\texttt{continue\_reasoning=False}).

The generator injects controller guidance into the LLM through \emph{assistant prefilling} via vLLM's \texttt{continue\_final\_message} mechanism.
Given an action $a_i^j$ that returns a \texttt{ReasoningIntervention} with fields \texttt{internal\_reasoning} (context guidance) and \texttt{prefix} (text that opens the next step), the adapter builds a prefill string of the form:
\begin{lstlisting}
## internal_reasoning
{guidance}
## {reasoning_field}
{prefix}
\end{lstlisting}
This is appended to the existing assistant message (i.e., prior reasoning steps if they exist), and the model continues generating from the prefix.
The prefill therefore simultaneously injects high-level guidance (via \texttt{internal\_reasoning}) and steers the surface form of the generation (via \texttt{prefix}).

Figure~\ref{fig:generator-pipeline} illustrates the full pipeline from state and intervention to the child state.

\begin{figure}[ht]
  \centering
  \begin{tikzpicture}[
      >=stealth,
      font=\footnotesize,
      node distance=1.1cm and 1.5cm,
      box/.style={draw=borderblue, rounded corners=4pt, align=center,
                  minimum width=3cm, minimum height=0.9cm, inner sep=6pt,
                  fill=white, line width=0.8pt},
      decision/.style={draw=borderblue, diamond, aspect=2.2,
                       minimum width=3cm, minimum height=0.8cm,
                       align=center, fill=white, inner sep=3pt, line width=0.8pt},
      arr/.style={->, thick, color=borderblue},
      lbl/.style={font=\scriptsize, midway, color=black}
    ]

    % Nodes
    % Inputs (outside bands)
    \node[box]  (state)  {State $s_{i-1}$};
    \node[box, right=1.5cm of state]
      (action) {Action $a_i^j$\\(intervention)};

    % Controller decision: FINISH or continue? (blue band)
    \node[decision, below=1.5cm of $(state)!0.5!(action)$]
      (dec) {\footnotesize\texttt{FINISH?}};

    % Generator: two branches (green band)
    \node[box, below left=1.5cm and 1.2cm of dec]
      (prefill) {Prefill builder\\
                 \texttt{internal\_reasoning}\\
                 $+$\; \texttt{prefix}};
    \node[box, below right=1.5cm and 1.2cm of dec]
      (synth) {Prefill builder\\prepares final output\\(answer)};

    \node[box, below=0.9cm of prefill]
      (step)   {vLLM\\stop: \texttt{</step>}};
    \node[box, below=0.9cm of synth]
      (answer) {vLLM\\stop: \texttt{</answer>}};

    % Output states (outside bands)
    \node[box, below=0.9cm of step]
      (child)  {Child state $s_i^j$};
    \node[box, below=0.9cm of answer]
      (final)  {Final state $s_i^j$};

    % Background bands
    \begin{pgfonlayer}{background}
      % Controller band: decision node
      \node[fit=(dec), inner sep=14pt,
            fill=bandblue, rounded corners=8pt] {};
      % Generator band: prefill, synth, vLLM (not output states)
      \node[fit=(prefill)(synth)(step)(answer), inner sep=12pt,
            fill=bandgreen, rounded corners=8pt] {};
    \end{pgfonlayer}

    % Arrows: inputs merge into decision
    \coordinate (mergeY) at ($(state.south)!0.5!(dec.north)$);
    \draw[arr] (state.south)  -- (state.south |- mergeY) -| (dec.north);
    \draw[-,  thick, color=borderblue] (action.south) -- (action.south |- mergeY) -| (dec.north);

    % Decision branches
    \draw[arr] (dec) -- node[lbl, left]  {Continue}  (prefill);
    \draw[arr] (dec) -- node[lbl, right] {FINISH}    (synth);
    \draw[arr] (prefill.south) -- (step.north);
    \draw[arr] (synth.south)   -- (answer.north);
    \draw[arr] (step.south)    -- (child.north);
    \draw[arr] (answer.south)  -- (final.north);

  \end{tikzpicture}
  \caption{ Generator pipeline. The controller decides whether to continue reasoning or finish (blue band). The generator executes accordingly (green band).}
  \label{fig:generator-pipeline}
\end{figure}

\paragraph{Intermediate reasoning steps.}
Once the controller selects actions $\{a^1_i, \ldots, a^n_i\}$ for a given parent state $s_{i-1}$, each action $a_i^j$ is executed to obtain text guidance $a_i^j()$, which is appended to the LLM's assistant message as a prefill.
Given the parent state $s_{i-1}=[x, Z_{i-1}]$, where $Z_{i-1} = [z_1, \ldots, z_{i-1}]$ represents the reasoning steps so far, we sample a continuation for each action as:
\begin{equation}
z_i^j \sim
G\bigl(
  z \mid x, \texttt{prefill}({Z_{i-1}, a_i^j()});\texttt{temp}
\bigr)\,[:\texttt{</step>}]
\end{equation}
Each generated thought $z_i^j$ is combined with the current state to form a child state $s_i^j = [s_{i-1}, z_i^j]$.
The continuation is terminated by the stop token \texttt{</step>}, per the prompt described in Prompt \ref{lst:generator-system}.
Table~\ref{tab:prefix-examples} illustrates what these interventions look like for our argument generation task.

\paragraph{Final answer generation.}
A final answer is always generated by selecting the \texttt{FINISH} action: either the controller chooses it explicitly (early stopping) or it is forced when maximum depth $d$ is reached.
In both cases, the formalism is the same: we condition on the reasoning trace and the \texttt{FINISH} action's prefill:
\begin{equation}
y^j \sim G(y \mid x, \texttt{prefill}(Z_{i-1}, a_i^j()); \texttt{temp})\,[:{\texttt{</answer>}}]
\end{equation}
where $a_i^j$ is the \texttt{FINISH} action, and the stop token is \texttt{</answer>}.

\newpage

\begin{tcolorbox}[
  after skip=2pt,
  colback=gray!8, colframe=gray!55, arc=4pt,
  title={\small\bfseries\sffamily System Message},
  fonttitle=\color{white}, coltitle=white, colbacktitle=gray!55,
  left=6pt, right=6pt, top=4pt, bottom=4pt,
]
\begin{lstlisting}[
  basicstyle=\ttfamily\scriptsize,
  breaklines=true,
  moredelim={**[is][\color{teal}]{@T}{T@}},
  moredelim={**[is][\color{blue}]{@B}{B@}},
]
# Instructions

@TGenerate an argument which takes the provided stance towards the provided topic.T@

Your inputs will be:
1. `topic` (str): @TThe topic to generate an argument aboutT@
2. `stance` (Literal['PRO', 'ANTI']): @TThe stance to take on the topicT@

Your goal is to produce the following output:
`argument` (str): @TThe generated argumentT@

When solving this problem, you must break down your solution into a series of
reasoning steps, followed by a final answer.
Each step towards the answer should be encased within <step>...</step> tags,
and contain a `@TclaimT@` that advances the solution towards producing `@TargumentT@`.
@TYour final answer must remain highly faithful to the reasoning steps and their
underlying ideas.

- Preserve the full set of reasoning steps and their original order.
- You may lightly rephrase for clarity and readability, but the meaning must remain unchanged.
- Structure and sequence should closely follow the original reasoning steps.
- Do NOT introduce new ideas, arguments, facts, or examples.
- Do NOT remove or significantly alter any existing reasoning.

Your goal is to produce a clear synthesis that respects both the content and
structure of the original reasoning, while allowing minimal refinement.T@

Your reasoning process should follow the rules below:
- Each `@TclaimT@` (of type `str`) entails @Ta component of the argument that advocates
  for the given stance towards the topicT@.
- Before writing a new `@TclaimT@`, start with some internal reasoning which
  discusses and guides what to do with the next `@TclaimT@`.

## Response Format

Once a user provides `@TtopicT@` and `@TstanceT@`, your response must follow this
exact template:

<thinking>
<step>
## internal_reasoning
Your internal reasoning about the first `@TclaimT@`
## @TclaimT@
The first reasoning step towards producing `@TargumentT@`
</step>
<step>
## internal_reasoning
Your internal reasoning about the second `@TclaimT@`
## @TclaimT@
The second reasoning step towards producing `@TargumentT@`
</step>
...
<step>
## internal_reasoning
Your internal reasoning about the final `@TclaimT@`
## @TclaimT@
The final reasoning step towards producing `@TargumentT@`
</step>
</thinking>
<answer>
## @TargumentT@
Your response for `@TargumentT@` here
</answer>
\end{lstlisting}
\end{tcolorbox}
\promptcaptionnote{lst:generator-system}{Generator --- System Message}{Instantiated for the argument generation running example (topic: \emph{Should single-use plastics be banned?}, stance: \emph{PRO}). \textcolor{teal}{Teal}: task signature and hyperparameters. \textcolor{blue}{Blue}: concrete inference-time value.}
\begin{tcolorbox}[
  after skip=2pt,
  colback=gray!8, colframe=gray!55, arc=4pt,
  title={\small\bfseries\sffamily User Message},
  fonttitle=\color{white}, coltitle=white, colbacktitle=gray!55,
  left=6pt, right=6pt, top=4pt, bottom=4pt,
]
\begin{lstlisting}[
  basicstyle=\ttfamily\scriptsize,
  breaklines=true,
  moredelim={**[is][\color{teal}]{@T}{T@}},
  moredelim={**[is][\color{blue}]{@B}{B@}},
]
@TGenerate an argument which takes the provided stance towards the provided topic.T@

## @TtopicT@
@BShould single-use plastics be banned?B@

## @TstanceT@
@BPROB@
\end{lstlisting}
\end{tcolorbox}
\promptcaption{lst:generator-user}{Generator --- User Message}
\begin{tcolorbox}[
  after skip=2pt,
  colback=gray!8, colframe=gray!55, arc=4pt,
  title={\small\bfseries\sffamily Assistant Message (prefill)},
  fonttitle=\color{white}, coltitle=white, colbacktitle=gray!55,
  left=6pt, right=6pt, top=4pt, bottom=4pt,
]
\begin{lstlisting}[
  basicstyle=\ttfamily\scriptsize,
  breaklines=true,
  moredelim={**[is][\color{teal}]{@T}{T@}},
  moredelim={**[is][\color{blue}]{@B}{B@}},
]
<thinking>
@B<step>
## internal_reasoning
I'll open with the core environmental harm caused by single-use plastics.
## claim
Single-use plastics are a leading contributor to ocean pollution, harming
marine ecosystems and entering the food chain.
</step>B@
<step>
## internal_reasoning
@BI should address the causal chain between plastic production and environmental
degradation, using concrete data.B@
## @TclaimT@
@BBanning single-use plastics wouldB@
\end{lstlisting}
\end{tcolorbox}
\promptcaption{lst:generator-prefill}{Generator --- Assistant Message (prefill)}

\newpage

\subsubsection{Synthesis modes}
\label{app:synthesis-modes}
Once a final answer is triggered, the generator instantiates one of four \emph{synthesis modes} that determine how tightly the output must mirror the intermediate reasoning trace.
The modes span a continuum from verbatim preservation to unconstrained generation, trading off action attribution strength against output fluency.
The synthesis mode is injected into the Generator's system prompt at inference time, so the same generator can operate in any of the four synthesis modes without retraining.
The system-prompt instructions for each mode are shown in Prompts~\ref{lst:synthesis-strict}--\ref{lst:synthesis-conclusion}.

\begin{tcolorbox}[
  after skip=2pt,
  colback=gray!8, colframe=gray!55, arc=4pt,
  title={\small\bfseries\sffamily Final Output Instruction --- Strict},
  fonttitle=\color{white}, coltitle=white, colbacktitle=gray!55,
  left=6pt, right=6pt, top=4pt, bottom=4pt,
]
\begin{lstlisting}[basicstyle=\ttfamily\scriptsize, breaklines=true]
Your final answer must include the full text from all reasoning steps,
copied nearly word-for-word and in sequential order.

- Preserve the exact wording, phrasing, structure, and examples.
- Maintain the original order and logical flow exactly as provided.
- You may add only: A brief introduction/conclusion, short transitions.
- Do NOT rewrite, paraphrase, summarize, or restructure.
- Do NOT add new ideas, arguments, facts, or examples.
\end{lstlisting}
\end{tcolorbox}
\promptcaptionnote{lst:synthesis-strict}{Final Output Instruction (Strict)}{Preserves exact wording from reasoning steps.}

\begin{tcolorbox}[
  after skip=2pt,
  colback=gray!8, colframe=gray!55, arc=4pt,
  title={\small\bfseries\sffamily Final Output Instruction --- Faithful},
  fonttitle=\color{white}, coltitle=white, colbacktitle=gray!55,
  left=6pt, right=6pt, top=4pt, bottom=4pt,
]
\begin{lstlisting}[basicstyle=\ttfamily\scriptsize, breaklines=true]
Your final answer must remain highly faithful to the reasoning steps and
their underlying ideas.

- Preserve the full set of reasoning steps and their original order.
- You may lightly rephrase for clarity and readability, but the meaning
  must remain unchanged.
- Structure and sequence should closely follow the original reasoning steps.
- Do NOT introduce new ideas, arguments, facts, or examples.
- Do NOT remove or significantly alter any existing reasoning.

Your goal is to produce a clear synthesis that respects both the content
and structure of the original reasoning, while allowing minimal refinement.
\end{lstlisting}
\end{tcolorbox}
\promptcaptionnote{lst:synthesis-faithful}{Final Output Instruction (Faithful)}{Allows light rephrasing while preserving meaning.}
\begin{tcolorbox}[
  after skip=2pt,
  colback=gray!8, colframe=gray!55, arc=4pt,
  title={\small\bfseries\sffamily Final Output Instruction --- Restructured},
  fonttitle=\color{white}, coltitle=white, colbacktitle=gray!55,
  left=6pt, right=6pt, top=4pt, bottom=4pt,
]
\begin{lstlisting}[basicstyle=\ttfamily\scriptsize, breaklines=true]
Your final answer should preserve the same core ideas and reasoning from
the steps provided, while improving clarity and coherence.

- Maintain the essential arguments and logical intent.
- You may rephrase, reorganize, and restructure the content for better
  flow and readability.
- The overall set of ideas should remain the same, but the presentation
  may differ.
- Do NOT introduce new ideas or factual content beyond what appears in
  the reasoning steps.

Your goal is to produce a well-structured synthesis that faithfully
reflects the original reasoning while optimizing expression and
organization.
\end{lstlisting}
\end{tcolorbox}
\promptcaptionnote{lst:synthesis-restructured}{Final Output Instruction (Restructured)}{Allows free reorganisation while preserving core ideas.}

\begin{tcolorbox}[
  after skip=2pt,
  colback=gray!8, colframe=gray!55, arc=4pt,
  title={\small\bfseries\sffamily Final Output Instruction --- Conclusion},
  fonttitle=\color{white}, coltitle=white, colbacktitle=gray!55,
  left=6pt, right=6pt, top=4pt, bottom=4pt,
]
\begin{lstlisting}[basicstyle=\ttfamily\scriptsize, breaklines=true]
Your final answer must be a standalone response to the user's task and
instructions.

- Focus on producing a clear, logically consistent, and high-quality
  final answer.
- You are not required to preserve the structure, wording, or order from
  the reasoning steps (between <thinking>...</thinking> tags).
- Use the reasoning steps only as internal guidance; do NOT mention them
  or refer to them.
- The user will *not* have access to the reasoning steps you wrote, so
  referencing them is confusing and unhelpful.
- Do NOT explain what you are going to do; just produce the final
  deliverable.
- If the task requires strict formatting (math, code, etc.), follow those
  requirements exactly in the final output.

Your goal is to output only the final answer content that satisfies the
user's instructions.
\end{lstlisting}
\end{tcolorbox}
\promptcaptionnote{lst:synthesis-conclusion}{Final Output Instruction (Conclusion)}{Treats reasoning as internal guidance only, producing a standalone answer.}

\newpage

\subsection{Evaluator}
\label{app:state-modules-evaluator}

\subsubsection{Generative Evaluator}
\label{app:state-modules-generative-evaluator}

The generative evaluator prompts an LLM to act as a judge and score the reasoning (PRM) or output (ORM) at hand. 
With this setup, the same generator model that produces thoughts and outputs is used to score them. 
In practice, our evaluators follow a \emph{multi-item rubric}: a list of domain-dependent criteria \footnote{
    For example, in argument generation tasks, the criteria should be centered on the argument's persuasiveness, logical rigor, and effective structure \citep{wachsmuth-etal-2017-computational}.
} the judge should check (e.g., constraint satisfaction, correctness, coherence, style).
Importantly, we allow rubric items to carry different importance weights, so that violations of high-priority requirements dominate the score.
This is implemented via weighted, multi-dimensional scoring (each rubric item has a weight, and the final score is a weighted aggregate).

In the original ToT paper, \citet{yao2023tree} proposed a generative LLM-as-a-Judge system that utilized majority voting over an ensemble of judges. 
This allowed the evaluator in ToT to calibrate its score given other candidates, rather than rate each one blindly.
While we support an ensemble of scoring judges, we do not support voting as a PRM or ORM evaluation mechanism.
We deliberately avoid the voting mechanism, since it does not scale well in generative models if the number of candidates to evaluate (in our case, either $n$ in the first layer or $n \times k$ in subsequent layers) is too large. 
We also built a generative Evaluator that utilizes many pairwise comparisons, and then ranks candidates according to Bradley-Terry (BT) scores. 
Even without covering all possible candidate pairs, BT allows us to estimate the relative quality of candidates in a non-isolated fashion, and without exceedingly large contexts.

\newpage
\begin{tcolorbox}[
  after skip=2pt,
  colback=gray!8, colframe=gray!55, arc=4pt,
  title={\small\bfseries\sffamily System Message (PRM)},
  fonttitle=\color{white}, coltitle=white, colbacktitle=gray!55,
  left=6pt, right=6pt, top=4pt, bottom=4pt,
]
\begin{lstlisting}[
  basicstyle=\ttfamily\scriptsize,
  breaklines=true,
  moredelim={**[is][\color{teal}]{@T}{T@}},
  moredelim={**[is][\color{blue}]{@B}{B@}},
]
@TJudge the quality of reasoning steps for a problem-solving process.
The task requires producing `argument` given `topic` and `stance`.
Reasoning steps towards producing `argument` are provided in
`reasoning_steps`.

Since you are evaluating intermediate reasoning, don't score based on
completeness, and instead score based on the rubric items below:
- persuasiveness: Persuasiveness score (an int between 1 and 7).
- coherence: Coherence score (an int between 1 and 7).
- relevance: Relevance score (an int between 1 and 7).T@

Your inputs will be:
1. `@TtopicT@` (str): @TThe topic to generate an argument aboutT@
2. `@TstanceT@` (Literal['PRO', 'ANTI']): @TThe stance to take on the topicT@
3. `@Treasoning_stepsT@` (list[str]): @TList of `claim`s to evaluate toward
producing `argument`T@

Your goal is to produce the following outputs:
1. `@TpersuasivenessT@` (int): @TPersuasiveness scoreT@
        Constraints: >= 1 and <= 7
2. `@TcoherenceT@` (int): @TCoherence scoreT@
        Constraints: >= 1 and <= 7
3. `@TrelevanceT@` (int): @TRelevance scoreT@
        Constraints: >= 1 and <= 7

Please provide your response with each output field under its own header using
the format:
## field_name
Your response for that field here

NOTE: Line breaks, capitalization, and punctuation (i.e., '##' followed by a
space) are important. If you do not follow these guidelines, your response will
be rejected.
\end{lstlisting}
\end{tcolorbox}
\promptcaptionnote{lst:gen-evaluator-system-prm}{Generative Evaluator --- System Message (PRM)}{\textcolor{teal}{Teal}: task signature and rubric as rendered by the adapter. \textcolor{blue}{Blue}: concrete runtime value.}
\begin{tcolorbox}[
  after skip=2pt,
  colback=gray!8, colframe=gray!55, arc=4pt,
  title={\small\bfseries\sffamily User Message (PRM)},
  fonttitle=\color{white}, coltitle=white, colbacktitle=gray!55,
  left=6pt, right=6pt, top=4pt, bottom=4pt,
]
\begin{lstlisting}[
  basicstyle=\ttfamily\scriptsize,
  breaklines=true,
  moredelim={**[is][\color{teal}]{@T}{T@}},
  moredelim={**[is][\color{blue}]{@B}{B@}},
]
## @TtopicT@
@BShould single-use plastics be banned?B@

## @TstanceT@
@BPROB@

## @Treasoning_stepsT@
@B['Single-use plastics are a leading contributor to ocean pollution, harming
marine ecosystems and entering the food chain.', 'Banning them would
accelerate adoption of reusable and biodegradable alternatives while reducing
long-term cleanup costs.']B@

Respond with the corresponding output fields, starting with the field
`## @TpersuasivenessT@` (must be formatted as a valid Python int), then
`## @TcoherenceT@` (must be formatted as a valid Python int), then
`## @TrelevanceT@` (must be formatted as a valid Python int)
\end{lstlisting}
\end{tcolorbox}
\promptcaption{lst:gen-evaluator-user-prm}{Generative Evaluator --- User Message (PRM)}

\newpage
\begin{tcolorbox}[
  after skip=2pt,
  colback=gray!8, colframe=gray!55, arc=4pt,
  title={\small\bfseries\sffamily System Message (ORM)},
  fonttitle=\color{white}, coltitle=white, colbacktitle=gray!55,
  left=6pt, right=6pt, top=4pt, bottom=4pt,
]
\begin{lstlisting}[
  basicstyle=\ttfamily\scriptsize,
  breaklines=true,
  moredelim={**[is][\color{teal}]{@T}{T@}},
  moredelim={**[is][\color{blue}]{@B}{B@}},
]
@TJudge the quality of a response for the provided task.
The task requires producing `argument` given `topic` and `stance`.

Evaluate the response using the rubric items below and assign numeric scores
to each:
- persuasiveness: Persuasiveness score (an int between 1 and 7).
- coherence: Coherence score (an int between 1 and 7).
- relevance: Relevance score (an int between 1 and 7).T@

Your inputs will be:
1. `@TtopicT@` (str): @TThe topic to generate an argument aboutT@
2. `@TstanceT@` (Literal['PRO', 'ANTI']): @TThe stance to take on the topicT@
3. `@TargumentT@` (str): @TThe generated argumentT@

Your goal is to produce the following outputs:
1. `@TpersuasivenessT@` (int): @TPersuasiveness scoreT@
        Constraints: >= 1 and <= 7
2. `@TcoherenceT@` (int): @TCoherence scoreT@
        Constraints: >= 1 and <= 7
3. `@TrelevanceT@` (int): @TRelevance scoreT@
        Constraints: >= 1 and <= 7

Please provide your response with each output field under its own header using
the format:
## field_name
Your response for that field here

NOTE: Line breaks, capitalization, and punctuation (i.e., '##' followed by a
space) are important. If you do not follow these guidelines, your response will
be rejected.
\end{lstlisting}
\end{tcolorbox}
\promptcaptionnote{lst:gen-evaluator-system}{Generative Evaluator --- System Message (ORM)}{\textcolor{teal}{Teal}: task signature and rubric as rendered by the adapter. \textcolor{blue}{Blue}: concrete runtime value.}
\begin{tcolorbox}[
  after skip=2pt,
  colback=gray!8, colframe=gray!55, arc=4pt,
  title={\small\bfseries\sffamily User Message (ORM)},
  fonttitle=\color{white}, coltitle=white, colbacktitle=gray!55,
  left=6pt, right=6pt, top=4pt, bottom=4pt,
]
\begin{lstlisting}[
  basicstyle=\ttfamily\scriptsize,
  breaklines=true,
  moredelim={**[is][\color{teal}]{@T}{T@}},
  moredelim={**[is][\color{blue}]{@B}{B@}},
]
## @TtopicT@
@BShould single-use plastics be banned?B@

## @TstanceT@
@BPROB@

## @TargumentT@
@BThe proliferation of single-use plastics represents one of the most pressing
environmental crises of our time. Each year, millions of tonnes of plastic waste
enter our oceans, choking marine life and entering the food chain. A
comprehensive ban would catalyse the development of sustainable alternatives
while sending a clear signal to industry and consumers alike.B@

Respond with the corresponding output fields, starting with the field
`## @TpersuasivenessT@` (must be formatted as a valid Python int), then
`## @TcoherenceT@` (must be formatted as a valid Python int), then
`## @TrelevanceT@` (must be formatted as a valid Python int)
\end{lstlisting}
\end{tcolorbox}
\promptcaption{lst:gen-evaluator-user}{Generative Evaluator --- User Message (ORM)}

\newpage
\subsubsection{Reranker Evaluator}
\label{app:state-modules-reranker-evaluator}

Analogous to the reranker controller (Appendix~\ref{app:state-modules-reranker-controller}), we can use a cross-encoder to score candidate states rather than candidate actions.
Here, the \emph{query} contains the input $x$ and the evaluation criteria (e.g., coherence, correctness, constraint satisfaction), while each \emph{document} is a candidate reasoning chain $Z_i$ (for intermediate evaluation) or final output $y$ (for outcome evaluation).
The reranker assigns a relevance score to each candidate, which we interpret as a quality estimate.
This approach is particularly efficient when the number of candidates is large, as cross-encoders can score all candidates in a single batched forward pass without requiring the longer generations that LLM-as-a-Judge evaluators produce.
Further, this resolves the sycophancy issues of the generative evaluator, which tends to award perfect scores to all good candidates \citep{sharma2024towards}.

Unlike the generative evaluator, which produces explicit integer scores that can be directly weighted and summed, the reranker model returns a single relevance logit.
\state extracts this logit by computing the log-probability of the token \texttt{yes} (vs.\ \texttt{no}) from the model's output distribution and applies a softmax to convert logit differences into a scalar in $(0, 1)$.
This scalar serves directly as the candidate's quality estimate for beam selection.

\begin{tcolorbox}[
  after skip=2pt,
  colback=gray!8, colframe=gray!55, arc=4pt,
  title={\small\bfseries\sffamily System Message},
  fonttitle=\color{white}, coltitle=white, colbacktitle=gray!55,
  left=6pt, right=6pt, top=4pt, bottom=4pt,
]
\begin{lstlisting}[
  basicstyle=\ttfamily\scriptsize,
  breaklines=true,
]
Judge whether the Document meets the requirements based on the Query and the
Instruct provided. Note that the answer can only be `yes' or `no'.
\end{lstlisting}
\end{tcolorbox}
\promptcaptionnote{lst:reranker-evaluator-system}{Reranker Evaluator --- System Message}{Uses the same Qwen3 query--document format as the reranker controller \citep{qwen3embedding}. The static system prompt is shared across PRM and ORM; the user message carries the instruction, rubric, query (inputs), and document (candidate output or reasoning trajectory). \textcolor{teal}{Teal}: task signature and rubric. \textcolor{blue}{Blue}: concrete runtime value.}
\begin{tcolorbox}[
  after skip=2pt,
  colback=gray!8, colframe=gray!55, arc=4pt,
  title={\small\bfseries\sffamily User Message (PRM)},
  fonttitle=\color{white}, coltitle=white, colbacktitle=gray!55,
  left=6pt, right=6pt, top=4pt, bottom=4pt,
]
\begin{lstlisting}[
  basicstyle=\ttfamily\scriptsize,
  breaklines=true,
  moredelim={**[is][\color{teal}]{@T}{T@}},
  moredelim={**[is][\color{blue}]{@B}{B@}},
  escapeinside={(*@}{@*)},
]
<Instruct>: Your objective is to judge a reasoning trajectory towards solving a
user-assigned task.

The user provided the following task:
(*@\char34@*)@TGenerate an argument which takes the provided stance towards the provided
topic.T@(*@\char34@*)

Since this is a reasoning task, we are interested not only in the final
output, but also in the reasoning process that leads to it.
You will find the inputs for this task under the (*@\#@*) Inputs header in the
Query.
You will find the intermediate reasoning trajectory towards solving this
problem under the (*@\#@*) Reasoning header in the Document.

@TJudge whether the provided reasoning trajectory is a strong partial solution
for the task given the inputs, using the rubric below.
Rubric:
- persuasiveness: Persuasiveness score
- coherence: Coherence score
- relevance: Relevance scoreT@

<Query>: (*@\#@*) Inputs
@Btopic: Should single-use plastics be banned?B@
@Bstance: PROB@

<Document>: (*@\#@*) Reasoning
@B<thinking>
<step>
## claim
Single-use plastics are a leading contributor to ocean pollution, harming
marine ecosystems and entering the food chain.
</step>
</thinking>B@
\end{lstlisting}
\end{tcolorbox}
\promptcaption{lst:reranker-evaluator-user-prm}{Reranker Evaluator --- User Message (PRM)}
\begin{tcolorbox}[
  after skip=2pt,
  colback=gray!8, colframe=gray!55, arc=4pt,
  title={\small\bfseries\sffamily User Message (ORM)},
  fonttitle=\color{white}, coltitle=white, colbacktitle=gray!55,
  left=6pt, right=6pt, top=4pt, bottom=4pt,
]
\begin{lstlisting}[
  basicstyle=\ttfamily\scriptsize,
  breaklines=true,
  moredelim={**[is][\color{teal}]{@T}{T@}},
  moredelim={**[is][\color{blue}]{@B}{B@}},
  escapeinside={(*@}{@*)},
]
<Instruct>: Your objective is to judge an output for a user-assigned task.

The user provided the following task:
(*@\char34@*)@TGenerate an argument which takes the provided stance towards the provided topic.T@(*@\char34@*)

You will find the inputs for this task under the (*@\#@*) Inputs header in the Query.
You will find the output under consideration under the (*@\#@*) Output heading in the
Document.

@TJudge whether the provided output is a strong final output for the task given
the inputs, using the rubric below.
Rubric:
- persuasiveness: Persuasiveness score
- coherence: Coherence score
- relevance: Relevance scoreT@

<Query>: (*@\#@*) Inputs
@Btopic: Should single-use plastics be banned?B@
@Bstance: PROB@

<Document>: (*@\#@*) Output
@Bargument: The proliferation of single-use plastics represents one of the most
pressing environmental crises of our time. Each year, millions of tonnes of
plastic waste enter our oceans, choking marine life and entering the food chain.
A comprehensive ban would catalyse the development of sustainable alternatives
while sending a clear signal to industry and consumers alike.B@
\end{lstlisting}
\end{tcolorbox}
\promptcaption{lst:reranker-evaluator-user}{Reranker Evaluator --- User Message (ORM)}

\subsubsection{Programmatic Evaluator}
\label{app:state-modules-programmatic-evaluator}

In some domains, intermediate and final states admit \emph{programmatic} evaluation: a deterministic procedure can verify correctness, constraint satisfaction, or structural validity without invoking an LLM \citep{lambert2025tulu3pushingfrontiers, gao2024designingeffectiverlreward, kimiteam2025kimik15scalingreinforcement}.
This setting crucially assumes an \emph{additive action space}, where each reasoning step produces text that is concatenated to all previous steps, so that a partial trajectory $Z_i = [z_1,\ldots,z_i]$ represents a prefix of a well-formed candidate solution.
Many instruction-following and mathematical tasks satisfy this property, as successive steps monotonically construct a single output whose validity can be checked incrementally (e.g., JSON well-formedness, exact string constraints, section counts, or arithmetic consistency).
This stands in contrast to \emph{metacognitive} action spaces -- such as self-reflection, critique, or targeted editing of an existing draft -- where actions do not compose into a single executable artifact and intermediate states cannot be interpreted as partial answers.
As a result, programmatic evaluators are inherently task-dependent and cannot be assumed to exist for all domains.
Formally, we define a \emph{Programmatic Evaluator} as a deterministic scoring function conditioned on the input $x$:\footnote{
    It is also possible to condition the ORM on $\text{concat}(x, Z_i)$ as opposed to just conditioning on $x$.
}
\begin{align}
    V^*_{\text{PRM}}(\text{concat}(Z_i) \mid x) \rightarrow [0,1] \\
    V^*_{\text{ORM}}(y \mid x) \rightarrow [0,1]
\end{align}

which evaluates whether the concatenated reasoning ($\operatorname{concat}(Z_i)$) or answer ($y$) satisfies all constraints induced by the task, $x$.\footnote{
    In practice, the concatenation operation ignores intermediate or internal reasoning fields (e.g., chain-of-thought or planning annotations), retaining only the externally visible output fields.
    Including internal reasoning in the concatenation would often invalidate otherwise correct partial outputs and interfere with reliable programmatic grading.
}
In tasks where constraint satisfaction is prefix-monotonic, $V^*$ can be used interchangeably as both a Process Reward Model and an Outcome Reward Model, i.e.,
\[
V^*_{\text{PRM}} \equiv V^*_{\text{ORM}} \equiv V^*,
\]
allowing invalid trajectories to be pruned immediately upon violation.
When available, programmatic evaluators eliminate judge variance, avoid sycophancy effects, and provide exact credit assignment over the action space.
However, their applicability fundamentally relies on additive action spaces and reliable prefix-level validation; extending programmatic evaluation to non-additive or revision-based action spaces remains an open challenge.

\section{Beam-Search Complexity}
\label{app:beam_search_call_complexity}

\state supports both full-tree exploration (no pruning) and heuristic-based tree search \citep{newell_and_simon_1976_symbols_and_search} in the form of Beam Search \citep{lowerre1976harpy}.
This section analyzes the search-space complexity under both regimes and discusses the implementation-level effects of batching with vLLM \citep{kwon2023efficient}.

\paragraph{Notation.}
Let $d$ denote the maximum reasoning depth, $n$ the branching factor, and $k$ the beam width.
Let $L_i$ denote the frontier after $i$ reasoning steps, with $b_i := |L_i|$ and $b_0 = 1$ (contains root only).

\paragraph{One layer.}
At each non-terminal layer $i < d$, every state in $L_i$ is expanded into $n$ children,\footnote{
    It is possible that some generations fail, and that in practice there are fewer than $n$ children for a given parent. For our complexity analysis, which concerns an upper bound, we assume zero failures
} producing a candidate pool of size $b_i \cdot n$.
If pruning is disabled, all candidates survive to the next layer. If pruning is enabled, the evaluator scores the candidates and the selector retains the top-$k$.

We first analyze the unpruned case to establish the baseline exponential growth, and then show how beam-search pruning reduces this scaling.

\paragraph{Full-tree expansion (no pruning).}
Without pruning, every candidate survives, so the frontier grows as $b_i = n^i$.
Each layer requires $b_i$ controller invocations (to select $n$ actions per state) and $b_i \cdot n$ generator completions (one per child), giving a total intermediate generation cost of
\begin{equation}
    N_{\mathrm{gen}}^{\mathrm{inter}}
    =
    \sum_{i=0}^{d-1} n^{i+1}
    =
    \begin{cases}
        dn,                        & n = 1, \\[4pt]
        \dfrac{n(n^d - 1)}{n - 1}, & n > 1.
    \end{cases}
\end{equation}
The controller cost matches the generator cost for a generative controller ($n$ calls per state), but may be higher for a reranker controller that scores all $|\mathcal{A}_i|$ actions per state (see below).
Our implementation does \emph{not} score non-terminal nodes when pruning is disabled; intermediate PRM-style evaluation is skipped, and only nodes containing final outputs are evaluated.
The practical implication is that disabling pruning converts the method from beam search into explicit tree expansion with exponential growth in depth.

\paragraph{Beam search (with pruning).}
With pruning, the frontier is truncated to at most $k$ states after each layer, so $b_i \le k$ for all $i \ge 1$.
% Therefore,
% \begin{equation}
%     \sum_{i=0}^{d-1} b_i
%     \le
%     1 + (d-1)\,k.
% \end{equation}
Under a generative controller, all three components (generator, evaluator, controller) scale with the same leading term for intermediate layers:
\begin{equation}
    N_{\mathrm{gen}}^{\mathrm{inter}},\;
    N_{\mathrm{eval}}^{\mathrm{inter}},\;
    N_{\mathrm{ctrl}}^{\mathrm{gen}}
    \;\le\;
    n \sum_{i=0}^{d-1} b_i
    \le
    n\bigl(1 + (d-1)\,k\bigr).
\end{equation}
Under pruning, the logical search cost thus grows \emph{linearly} in depth and beam width, rather than exponentially in depth.
This is the standard computational advantage of beam search, now instantiated over reasoning trajectories.

\paragraph{Reranker controller cost.}
The bounds above assume a generative controller, which produces $n$ action choices per state.
In contrast, a reranker controller scores \emph{all} admissible actions individually, so its cost at layer $i$ is $b_i \cdot m_i$ where $m_i \le |\mathcal{A}_i|$ is the number of admissible actions.
Hence, reranker-controller complexity is governed not only by beam width but also by action-space size:
\begin{equation}
    N_{\mathrm{ctrl}}^{\mathrm{rank}}
    =
    \sum_{i=0}^{d-1} b_i \, m_i,
    \qquad m_i \le |\mathcal{A}_i|.
\end{equation}

\paragraph{Final synthesis step.}
At maximum depth $d$, the implementation sets every remaining frontier state to \texttt{FINISH} and generates $n_{\mathrm{final}}$ outputs per surviving trajectory.
The final generation and evaluation costs are
\begin{equation}
    N_{\mathrm{gen}}^{\mathrm{final}}
    = n_{\mathrm{final}} \, b_d,
    \qquad
    N_{\mathrm{eval}}^{\mathrm{final}}
    = n_{\mathrm{final}} \, b_d.
\end{equation}
Under pruning, $b_d \le k$, so both costs are at most $n_{\mathrm{final}} \cdot k$.
Without pruning, $b_d = n^d$ in the worst case.

\paragraph{Batched implementation calls.}
The expressions above count \emph{logical} state-level interactions.
In practice, the implementation batches all states in a frontier into a single vLLM call per layer.
In the worst case (no early stopping), there are at most
\begin{equation}
    B_{\mathrm{ctrl}} \le d,
    \qquad
    B_{\mathrm{gen}} \le d+1,
    \qquad
    B_{\mathrm{eval}} \le d+1
\end{equation}
batched forward passes for the controller, generator, and evaluator, respectively.
The extra pass comes from the final synthesis step; the controller is not called during that step since \texttt{FINISH} is set deterministically.
These bounds explain why batch inference remains practical even when the logical number of trajectories is large: the number of batched model calls grows linearly with depth and is independent of beam width at the level of batched calls.

\paragraph{Early stopping and special cases.}
When early stopping is enabled, the controller may emit \texttt{FINISH} before depth $d$, which can only reduce the realized cost relative to the bounds above.
All expressions in this section should therefore be interpreted as worst-case upper bounds.

Several common decoding procedures emerge as boundary cases.
With $d=1$ and no pruning, the procedure reduces to Best-of-$n$ over a single reasoning step.
With $d=1$ and $k<n$, it becomes a one-step beam search.
With larger $d$ and pruning, it yields multi-step beam search over partial trajectories.

Overall, \state does not introduce a new asymptotic search primitive; rather, it instantiates beam search over \emph{structured reasoning actions and partial trajectories}.
The resulting complexity is controlled by the same core quantities as classical beam search (depth $d$, branching factor $n$, beam width $k$), together with three \state-specific factors: the first-layer beam width $k_1$ (when set independently of $k$), the action-space size $|\mathcal{A}|$ (for reranker controllers), and the number of final responses $n_{\mathrm{final}}$.

\newpage

\section{Complete Experiments}
\label{app:complete-results}

We conduct all runs with vLLM \citep{kwon2023efficient} in offline mode to enable efficient layer-wise batching across tree expansions. 
Baseline methods (including ToT) require 1 GPU to generate outputs, while \state requires 2 GPUs.
When using 2 GPUs for \state, one is meant for the generative LLM that serves the Generator (Appendix \ref{app:state-modules-generator}) and the other is meant for the reranker LLM that serves the Controller (Appendix \ref{app:state-modules-reranker-controller}) and Reranker (Appendix \ref{app:state-modules-reranker-evaluator}).

\subsection{NoveltyBench}
\label{app:noveltybench}

We evaluate on the curated NoveltyBench set of 100 prompts spanning four categories: \emph{randomness}, \emph{factual knowledge} with underspecified queries, \emph{creative writing}, and \emph{subjectivity}.
Because NoveltyBench provides only a single ``test'' split, we use the first 10 prompts as a development subset to refine prompts and system settings, and report final results on the remaining 90 prompts.

We compare Best-of-$n$ baselines (I/O and CoT), ToT, and \state across seven models from four families: Qwen3 \citep{yang2025qwen3technicalreport}, Gemma-3 \citep{gemmateam2025gemma3technicalreport}, Nemotron-3 \citep{nvidia2025nvidianemotron3efficient}, and Ministral-3 \citep{liu2026ministral3}.
For \state, we use Qwen3-8B-Reranker \citep{qwen3embedding} for both the reranker controller (Appendix~\ref{app:state-modules-reranker-controller}) and reranker evaluator (Appendix~\ref{app:state-modules-reranker-evaluator}).
We run depth-1 trees with wide branching ($n=8$, $k=8$), so each candidate includes one reasoning step and one final answer, and repeat each configuration over 10 random seeds.
We focus on a single branching operation since deeper heuristic search optimizes for evaluator-aligned scores rather than frontier diversity, and deeper trajectories often share parent states (which introduces overlapping reasoning). 
Therefore, diversity as a function of tree depth is outside this experiment's scope.

To isolate the impact of controller-guided interventions, we also include baselines that expose the same action spaces directly in prompts (as DSPy input fields, described in Appendix \ref{app:dspy-background}).
This setup evaluates diversity as a function of (1) choice of reasoning template (I/O vs. CoT vs. ToT), (2) including the action space as an explicit prompt input, and (3) enabling controller-guided interventions.
The action space combines two dimensions with 5 choices each, yielding 25 action combinations per step (Appendix~\ref{app:action-spaces-noveltybench}): \emph{personality traits} (following the Big Five model; \citealp{goldberg1990alternative}) and \emph{target audience} (demographic age to appeal to).
Each action provides internal reasoning guidance (Appendix~\ref{app:state-modules-generator}) that steers generation toward the selected persona or audience.
In this configuration, \state's internal evaluator does not affect search or final selection because all outputs are returned.
We sweep low-, medium-, and high-temperature regimes per model: $T \in \{0.5, 0.7, 1.0\}$ for most models and $T \in \{0.1, 0.3, 0.5\}$ for Ministral-3 (consistent with provider recommendations).

\paragraph{Metrics.}
Diversity is measured as Mean Distinct$_k$ \citep{zhang2025noveltybenchevaluatinglanguagemodels}, which counts the number of meaningfully distinct responses a model generates in $k$ samples (where two responses are considered distinct if a user would benefit from seeing both).
\citet{zhang2025noveltybenchevaluatinglanguagemodels} measure diversity through clustering responses with a fine-tuned DeBERTa model \citep{he2021deberta}.
NoveltyBench also computes the ``quality'' of a response through an LLM-as-a-Judge score \citep{liu2026humanai}, which is assigned to each individual response.
Utility is measured as Mean Utility$_k$ \citep{zhang2025noveltybenchevaluatinglanguagemodels}, which formalizes the cumulative benefit to a user who sequentially inspects $k$ generations and only gains value from a response when it is distinct from all previously seen ones.
Formally, $\text{Utility}_k = \sum_{i=1}^{k} q_i \cdot \mathbbm{1}[\text{response } i \text{ is distinct from responses } 1, \ldots, i{-}1]$, where $q_i$ is the quality score of the $i$-th response, and responses are distinct if they are assigned to different clusters (in our case, by DeBERTa embeddings).
Utility thus jointly penalizes low quality \emph{and} redundancy: a method can only accumulate high utility by generating responses that are both good and genuinely new.
Tables~\ref{tab:noveltybench-curated-test-diversity}, \ref{tab:noveltybench-curated-test-raw-judge}, and \ref{tab:noveltybench-curated-test-utility} report all three metrics across all model and temperature configurations.

\subsubsection{Generalizability Across Models}
\label{app:noveltybench-generalizability}

\state achieves the highest diversity across all seven models and all three temperature regimes (Table~\ref{tab:noveltybench-curated-test-diversity}).
Relative to Best-of-$n$ with I/O prompting, the simplest and most common ITC method, \state roughly doubles diversity at medium temperature across most models.
\state's gains range from $+49\%$ for Qwen3-8B ($4.63$ vs.\ $3.11$) to $+153\%$ for Gemma-3-27B ($4.35$ vs.\ $1.72$), with a mean improvement of $+95\%$ across the seven models.
These gains reflect a systematic limitation of I/O sampling: without structural interventions, repeated high-temperature sampling collapses toward the same few high-probability completions.

Relative to the strongest non-\state baseline (which varies by model), the margin is more modest but consistent.
Gains are largest for models in the mid-capability range---$+37\%$ for Qwen3-30B ($4.57$ vs.\ $3.33$ for CoT w/ Action Space), $+35\%$ for Gemma-3-4B ($3.77$ vs.\ $2.79$), $+33\%$ for Gemma-3-27B ($4.35$ vs.\ $3.26$)---and smaller for models where action-space-augmented baselines already achieve strong diversity on their own: $+7\%$ for Nemotron-3-30B ($5.39$ vs.\ $5.05$ for ToT w/ Action Space) and $+5\%$ for Ministral-3-14B ($4.86$ vs.\ $4.64$ for CoT w/ Action Space).
The consistent pattern across all models is that controller-guided prefix interventions provide a meaningful additional boost beyond simply exposing the action space in the prompt: ``CoT w/ Action Space'' and ``ToT w/ Action Space'' always
trail \state despite having access to the same action vocabulary, confirming that it is the \emph{enforcement} of actions via prefilling---not their mere presence in the context---that drives diversity.

\begin{table*}[h!]
\centering
\scriptsize
\setlength{\tabcolsep}{6pt}
\begin{tabular}{llccc}
\toprule
\textbf{Model} & \textbf{Method} & \textbf{Low} & \textbf{Medium} & \textbf{High} \\
\midrule
 \textbf{Ministral 3 14B} & Baseline & 1.96\,$\pm$\,0.05 & 2.76\,$\pm$\,0.08 & 3.52\,$\pm$\,0.07 \\
   & Baseline CoT & 3.16\,$\pm$\,0.09 & 3.79\,$\pm$\,0.06 & 4.28\,$\pm$\,0.12 \\
   & Baseline w/ Action Space & 2.56\,$\pm$\,0.07 & 3.55\,$\pm$\,0.1 & 4.5\,$\pm$\,0.08 \\
   & Baseline CoT w/ Action Space & \underline{4.08\,$\pm$\,0.1} & \underline{4.64\,$\pm$\,0.1} & 4.95\,$\pm$\,0.11 \\
   & Baseline ToT & 2.91\,$\pm$\,0.09 & 3.73\,$\pm$\,0.1 & 4.53\,$\pm$\,0.09 \\
   & Baseline ToT w/ Action Space & 3.67\,$\pm$\,0.1 & 4.48\,$\pm$\,0.09 & \underline{5.2\,$\pm$\,0.12} \\
   & STATe of Thoughts & \textbf{4.66\,$\pm$\,0.12} & \textbf{4.86\,$\pm$\,0.07} & \textbf{5.29\,$\pm$\,0.08} \\
\midrule
 \textbf{Qwen3 4B} & Baseline & 1.89\,$\pm$\,0.06 & 2.26\,$\pm$\,0.08 & 2.73\,$\pm$\,0.08 \\
   & Baseline CoT & 2.68\,$\pm$\,0.11 & 2.87\,$\pm$\,0.09 & 3.25\,$\pm$\,0.11 \\
   & Baseline w/ Action Space & 1.85\,$\pm$\,0.06 & 2.16\,$\pm$\,0.06 & 2.66\,$\pm$\,0.08 \\
   & Baseline CoT w/ Action Space & \underline{3.18\,$\pm$\,0.1} & \underline{3.38\,$\pm$\,0.09} & \underline{3.76\,$\pm$\,0.09} \\
   & Baseline ToT & 2.31\,$\pm$\,0.06 & 2.65\,$\pm$\,0.06 & 3.11\,$\pm$\,0.1 \\
   & Baseline ToT w/ Action Space & 2.54\,$\pm$\,0.09 & 2.91\,$\pm$\,0.11 & 3.41\,$\pm$\,0.06 \\
   & STATe of Thoughts & \textbf{3.73\,$\pm$\,0.08} & \textbf{4.1\,$\pm$\,0.13} & \textbf{4.67\,$\pm$\,0.06} \\
\midrule
 \textbf{Qwen3 8B} & Baseline & 2.91\,$\pm$\,0.07 & 3.11\,$\pm$\,0.12 & 3.24\,$\pm$\,0.11 \\
   & Baseline CoT & 3.14\,$\pm$\,0.08 & 3.27\,$\pm$\,0.1 & 3.46\,$\pm$\,0.06 \\
   & Baseline w/ Action Space & \underline{3.19\,$\pm$\,0.07} & 3.32\,$\pm$\,0.1 & 3.58\,$\pm$\,0.06 \\
   & Baseline CoT w/ Action Space & 3.19\,$\pm$\,0.1 & 3.38\,$\pm$\,0.09 & 3.58\,$\pm$\,0.1 \\
   & Baseline ToT & 2.47\,$\pm$\,0.09 & 2.83\,$\pm$\,0.06 & 3.32\,$\pm$\,0.12 \\
   & Baseline ToT w/ Action Space & 3.16\,$\pm$\,0.08 & \underline{3.6\,$\pm$\,0.11} & \underline{4.22\,$\pm$\,0.1} \\
   & STATe of Thoughts & \textbf{4.31\,$\pm$\,0.1} & \textbf{4.63\,$\pm$\,0.07} & \textbf{5.28\,$\pm$\,0.07} \\
\midrule
 \textbf{Qwen3 30B} & Baseline & 1.68\,$\pm$\,0.05 & 1.98\,$\pm$\,0.03 & 2.41\,$\pm$\,0.05 \\
   & Baseline CoT & 2.31\,$\pm$\,0.06 & 2.59\,$\pm$\,0.09 & 3.0\,$\pm$\,0.1 \\
   & Baseline w/ Action Space & 1.94\,$\pm$\,0.05 & 2.26\,$\pm$\,0.1 & 2.84\,$\pm$\,0.09 \\
   & Baseline CoT w/ Action Space & \underline{2.98\,$\pm$\,0.09} & \underline{3.33\,$\pm$\,0.12} & \underline{3.76\,$\pm$\,0.1} \\
   & Baseline ToT & 1.97\,$\pm$\,0.05 & 2.27\,$\pm$\,0.05 & 2.78\,$\pm$\,0.11 \\
   & Baseline ToT w/ Action Space & 2.38\,$\pm$\,0.06 & 2.76\,$\pm$\,0.08 & 3.29\,$\pm$\,0.11 \\
   & STATe of Thoughts & \textbf{4.24\,$\pm$\,0.11} & \textbf{4.57\,$\pm$\,0.13} & \textbf{4.94\,$\pm$\,0.1} \\
\midrule
 \textbf{Gemma 3 4B} & Baseline & 1.58\,$\pm$\,0.03 & 1.78\,$\pm$\,0.09 & 2.14\,$\pm$\,0.05 \\
   & Baseline CoT & 2.25\,$\pm$\,0.08 & 2.47\,$\pm$\,0.07 & 2.72\,$\pm$\,0.08 \\
   & Baseline w/ Action Space & 1.85\,$\pm$\,0.06 & 2.17\,$\pm$\,0.06 & 2.7\,$\pm$\,0.08 \\
   & Baseline CoT w/ Action Space & \underline{2.53\,$\pm$\,0.09} & \underline{2.79\,$\pm$\,0.08} & \underline{3.12\,$\pm$\,0.08} \\
   & Baseline ToT & 1.86\,$\pm$\,0.05 & 2.16\,$\pm$\,0.06 & 2.6\,$\pm$\,0.08 \\
   & Baseline ToT w/ Action Space & 2.12\,$\pm$\,0.09 & 2.5\,$\pm$\,0.08 & 2.92\,$\pm$\,0.09 \\
   & STATe of Thoughts & \textbf{3.55\,$\pm$\,0.07} & \textbf{3.77\,$\pm$\,0.05} & \textbf{4.09\,$\pm$\,0.08} \\
\midrule
 \textbf{Gemma 3 27B} & Baseline & 1.54\,$\pm$\,0.04 & 1.72\,$\pm$\,0.05 & 2.04\,$\pm$\,0.05 \\
   & Baseline CoT & 2.35\,$\pm$\,0.07 & 2.55\,$\pm$\,0.11 & 2.86\,$\pm$\,0.09 \\
   & Baseline w/ Action Space & 1.81\,$\pm$\,0.04 & 2.05\,$\pm$\,0.08 & 2.47\,$\pm$\,0.1 \\
   & Baseline CoT w/ Action Space & \underline{3.03\,$\pm$\,0.09} & \underline{3.26\,$\pm$\,0.07} & \underline{3.59\,$\pm$\,0.05} \\
   & Baseline ToT & 2.13\,$\pm$\,0.07 & 2.49\,$\pm$\,0.08 & 2.88\,$\pm$\,0.08 \\
   & Baseline ToT w/ Action Space & 2.69\,$\pm$\,0.08 & 3.0\,$\pm$\,0.09 & 3.44\,$\pm$\,0.09 \\
   & STATe of Thoughts & \textbf{4.21\,$\pm$\,0.08} & \textbf{4.35\,$\pm$\,0.09} & \textbf{4.66\,$\pm$\,0.09} \\
\midrule
 \textbf{Nemotron 3 30B} & Baseline & 2.74\,$\pm$\,0.08 & 3.25\,$\pm$\,0.07 & 4.32\,$\pm$\,0.13 \\
   & Baseline CoT & 3.28\,$\pm$\,0.09 & 3.58\,$\pm$\,0.1 & 4.33\,$\pm$\,0.12 \\
   & Baseline w/ Action Space & 3.82\,$\pm$\,0.1 & 4.73\,$\pm$\,0.15 & 5.91\,$\pm$\,0.12 \\
   & Baseline CoT w/ Action Space & 4.38\,$\pm$\,0.09 & 4.82\,$\pm$\,0.14 & 5.71\,$\pm$\,0.15 \\
   & Baseline ToT & 3.25\,$\pm$\,0.1 & 3.82\,$\pm$\,0.09 & 4.78\,$\pm$\,0.07 \\
   & Baseline ToT w/ Action Space & \underline{4.49\,$\pm$\,0.11} & \underline{5.05\,$\pm$\,0.15} & \underline{6.05\,$\pm$\,0.09} \\
   & STATe of Thoughts & \textbf{4.81\,$\pm$\,0.11} & \textbf{5.39\,$\pm$\,0.12} & \textbf{6.28\,$\pm$\,0.12} \\
\bottomrule
\end{tabular}
\caption{NoveltyBench Mean Distinct (diversity) for all models (mean$\pm$std over seeds). Low, medium, and high temperature correspond to $T=0.1$, $0.3$, $0.5$ for Ministral~3~14B, and $T=0.5$, $0.7$, $1.0$ for others. Best and second best per model per column are \textbf{bolded} and \underline{underlined}.}
\label{tab:noveltybench-curated-test-diversity}
\end{table*}

\subsubsection{Quality and Utility}
\label{app:noveltybench-diversity-quality}

\paragraph{Quality.}
\state achieves the highest raw quality score (Table~\ref{tab:noveltybench-curated-test-raw-judge}) across all seven models and all temperature settings without exception, demonstrating that diversity gains do not come
at the expense of individual response quality.
Relative to I/O Best-of-$n$, quality improvements at medium temperature range from $+13\%$ for Qwen3-8B ($3.29$ vs.\ $2.90$) to $+89\%$ for Gemma-3-27B ($3.46$ vs.\ $1.83$), with a mean of $+60\%$ across models.
This is particularly notable because I/O sampling is optimized to maximize the quality of the single most likely response, yet \state's structurally diversified outputs surpass it on quality as well.
When we introduce an inductive bias in the form of our action space, quality tends to improve on both baselines and \state.
% Notably, \state utilizes the action space directly as part of generation rather than including it in the system prompt.

\begin{table*}[h!]
\centering
\scriptsize
\setlength{\tabcolsep}{6pt}
\begin{tabular}{llccc}
\toprule
\textbf{Model} & \textbf{Method} & \textbf{Low} & \textbf{Medium} & \textbf{High} \\
\midrule
 \textbf{Ministral 3 14B} & Baseline & 2.09\,$\pm$\,0.05 & 2.9\,$\pm$\,0.09 & 3.74\,$\pm$\,0.07 \\
   & Baseline CoT & 3.12\,$\pm$\,0.09 & 3.74\,$\pm$\,0.05 & 4.23\,$\pm$\,0.13 \\
   & Baseline w/ Action Space & 2.39\,$\pm$\,0.06 & 3.17\,$\pm$\,0.11 & 3.84\,$\pm$\,0.08 \\
   & Baseline CoT w/ Action Space & \underline{3.84\,$\pm$\,0.11} & \underline{4.37\,$\pm$\,0.12} & \underline{4.66\,$\pm$\,0.14} \\
   & Baseline ToT & 2.9\,$\pm$\,0.07 & 3.74\,$\pm$\,0.11 & 4.53\,$\pm$\,0.12 \\
   & Baseline ToT w/ Action Space & 3.05\,$\pm$\,0.13 & 3.66\,$\pm$\,0.07 & 4.24\,$\pm$\,0.09 \\
   & STATe of Thoughts & \textbf{4.81\,$\pm$\,0.12} & \textbf{5.01\,$\pm$\,0.1} & \textbf{5.4\,$\pm$\,0.1} \\
\midrule
 \textbf{Qwen3 4B} & Baseline & 1.6\,$\pm$\,0.05 & 1.89\,$\pm$\,0.07 & 2.2\,$\pm$\,0.06 \\
   & Baseline CoT & 2.25\,$\pm$\,0.09 & 2.4\,$\pm$\,0.09 & 2.71\,$\pm$\,0.08 \\
   & Baseline w/ Action Space & 1.44\,$\pm$\,0.05 & 1.66\,$\pm$\,0.04 & 1.98\,$\pm$\,0.06 \\
   & Baseline CoT w/ Action Space & \underline{2.58\,$\pm$\,0.09} & \underline{2.74\,$\pm$\,0.1} & \underline{3.0\,$\pm$\,0.1} \\
   & Baseline ToT & 1.96\,$\pm$\,0.05 & 2.24\,$\pm$\,0.08 & 2.56\,$\pm$\,0.1 \\
   & Baseline ToT w/ Action Space & 1.94\,$\pm$\,0.07 & 2.21\,$\pm$\,0.07 & 2.52\,$\pm$\,0.07 \\
   & STATe of Thoughts & \textbf{2.67\,$\pm$\,0.09} & \textbf{2.87\,$\pm$\,0.11} & \textbf{3.23\,$\pm$\,0.06} \\
\midrule
 \textbf{Qwen3 8B} & Baseline & 2.69\,$\pm$\,0.09 & 2.9\,$\pm$\,0.12 & 3.01\,$\pm$\,0.1 \\
   & Baseline CoT & \underline{2.89\,$\pm$\,0.12} & \underline{2.97\,$\pm$\,0.11} & \underline{3.14\,$\pm$\,0.07} \\
   & Baseline w/ Action Space & 2.6\,$\pm$\,0.05 & 2.69\,$\pm$\,0.09 & 2.82\,$\pm$\,0.08 \\
   & Baseline CoT w/ Action Space & 2.73\,$\pm$\,0.09 & 2.9\,$\pm$\,0.08 & 3.08\,$\pm$\,0.13 \\
   & Baseline ToT & 1.98\,$\pm$\,0.06 & 2.25\,$\pm$\,0.07 & 2.62\,$\pm$\,0.08 \\
   & Baseline ToT w/ Action Space & 1.9\,$\pm$\,0.09 & 2.17\,$\pm$\,0.08 & 2.52\,$\pm$\,0.08 \\
   & STATe of Thoughts & \textbf{3.14\,$\pm$\,0.09} & \textbf{3.29\,$\pm$\,0.08} & \textbf{3.64\,$\pm$\,0.09} \\
\midrule
 \textbf{Qwen3 30B} & Baseline & 1.67\,$\pm$\,0.05 & 1.9\,$\pm$\,0.04 & 2.25\,$\pm$\,0.05 \\
   & Baseline CoT & 2.13\,$\pm$\,0.06 & 2.31\,$\pm$\,0.08 & 2.66\,$\pm$\,0.11 \\
   & Baseline w/ Action Space & 1.69\,$\pm$\,0.04 & 1.91\,$\pm$\,0.1 & 2.37\,$\pm$\,0.09 \\
   & Baseline CoT w/ Action Space & \underline{2.59\,$\pm$\,0.08} & \underline{2.9\,$\pm$\,0.1} & \underline{3.23\,$\pm$\,0.1} \\
   & Baseline ToT & 1.72\,$\pm$\,0.06 & 1.99\,$\pm$\,0.06 & 2.4\,$\pm$\,0.08 \\
   & Baseline ToT w/ Action Space & 1.99\,$\pm$\,0.06 & 2.32\,$\pm$\,0.06 & 2.7\,$\pm$\,0.12 \\
   & STATe of Thoughts & \textbf{3.36\,$\pm$\,0.09} & \textbf{3.52\,$\pm$\,0.08} & \textbf{3.73\,$\pm$\,0.09} \\
\midrule
 \textbf{Gemma 3 4B} & Baseline & 1.47\,$\pm$\,0.04 & 1.63\,$\pm$\,0.05 & 1.96\,$\pm$\,0.05 \\
   & Baseline CoT & 1.92\,$\pm$\,0.08 & 2.08\,$\pm$\,0.09 & 2.28\,$\pm$\,0.05 \\
   & Baseline w/ Action Space & 1.58\,$\pm$\,0.06 & 1.84\,$\pm$\,0.04 & 2.26\,$\pm$\,0.07 \\
   & Baseline CoT w/ Action Space & \underline{2.1\,$\pm$\,0.12} & \underline{2.34\,$\pm$\,0.06} & \underline{2.6\,$\pm$\,0.07} \\
   & Baseline ToT & 1.56\,$\pm$\,0.06 & 1.78\,$\pm$\,0.06 & 2.1\,$\pm$\,0.08 \\
   & Baseline ToT w/ Action Space & 1.69\,$\pm$\,0.09 & 1.95\,$\pm$\,0.06 & 2.24\,$\pm$\,0.07 \\
   & STATe of Thoughts & \textbf{2.53\,$\pm$\,0.08} & \textbf{2.69\,$\pm$\,0.06} & \textbf{2.84\,$\pm$\,0.08} \\
\midrule
 \textbf{Gemma 3 27B} & Baseline & 1.67\,$\pm$\,0.04 & 1.83\,$\pm$\,0.05 & 2.13\,$\pm$\,0.05 \\
   & Baseline CoT & 2.18\,$\pm$\,0.07 & 2.35\,$\pm$\,0.1 & 2.63\,$\pm$\,0.08 \\
   & Baseline w/ Action Space & 1.76\,$\pm$\,0.03 & 2.01\,$\pm$\,0.07 & 2.35\,$\pm$\,0.07 \\
   & Baseline CoT w/ Action Space & \underline{2.77\,$\pm$\,0.08} & \underline{3.01\,$\pm$\,0.07} & \underline{3.35\,$\pm$\,0.06} \\
   & Baseline ToT & 1.94\,$\pm$\,0.07 & 2.24\,$\pm$\,0.05 & 2.58\,$\pm$\,0.06 \\
   & Baseline ToT w/ Action Space & 2.41\,$\pm$\,0.07 & 2.67\,$\pm$\,0.06 & 2.99\,$\pm$\,0.07 \\
   & STATe of Thoughts & \textbf{3.39\,$\pm$\,0.1} & \textbf{3.46\,$\pm$\,0.1} & \textbf{3.75\,$\pm$\,0.08} \\
\midrule
 \textbf{Nemotron 3 30B} & Baseline & 2.48\,$\pm$\,0.07 & 2.87\,$\pm$\,0.07 & 3.6\,$\pm$\,0.09 \\
   & Baseline CoT & 2.97\,$\pm$\,0.08 & 3.2\,$\pm$\,0.1 & 3.86\,$\pm$\,0.09 \\
   & Baseline w/ Action Space & 2.87\,$\pm$\,0.1 & 3.47\,$\pm$\,0.09 & 3.74\,$\pm$\,0.15 \\
   & Baseline CoT w/ Action Space & \underline{3.39\,$\pm$\,0.12} & 3.66\,$\pm$\,0.13 & 4.04\,$\pm$\,0.17 \\
   & Baseline ToT & 2.66\,$\pm$\,0.11 & 3.14\,$\pm$\,0.07 & 3.76\,$\pm$\,0.08 \\
   & Baseline ToT w/ Action Space & 3.38\,$\pm$\,0.12 & \underline{3.79\,$\pm$\,0.1} & \underline{4.1\,$\pm$\,0.1} \\
   & STATe of Thoughts & \textbf{3.7\,$\pm$\,0.11} & \textbf{4.02\,$\pm$\,0.12} & \textbf{4.21\,$\pm$\,0.11} \\
\bottomrule
\end{tabular}
\caption{NoveltyBench Mean Quality (raw LLM-as-a-Judge score) for all models (mean$\pm$std over seeds). Low, medium, and high temperature correspond to $T=0.1$, $0.3$, $0.5$ for Ministral~3~14B, and $T=0.5$, $0.7$, $1.0$ for others. Best and second best per model per column are \textbf{bolded} and \underline{underlined}.}
\label{tab:noveltybench-curated-test-raw-judge}
\end{table*}

\paragraph{Utility.}
Utility results (Table~\ref{tab:noveltybench-curated-test-utility}) tell a similar story.
Relative to I/O Best-of-$n$ at medium temperature, \state improves utility by $+33\%$ for Ministral-3-14B ($6.53$ vs.\ $4.90$), $+33\%$ for Qwen3-30B ($5.12$ vs.\ $3.84$), $+30\%$ for Gemma-3-27B ($5.08$ vs.\ $3.90$), and $+19\%$ for Nemotron-3-30B ($5.62$ vs.\ $4.73$).
\state ranks first on five of the seven models across most temperatures.

The two exceptions are the smallest models in each family.
For Qwen3-4B, CoT w/ Action Space achieves higher utility at low and medium temperature ($4.37$ and $4.51$ vs.\ \state's $4.27$ and $4.40$), and for Gemma-3-4B the margin is negligible at high temperature ($4.47$ vs.\ $4.45$).
Both cases share a likely cause: \state's reranker controller orders responses by proximity to its top-ranked action preference.
As a result, high-ranked controller outputs tend to cluster around a dominant action choice (e.g., most actions preferring the same target audience).
Because utility accumulates sequentially and is discounted for redundant responses, this ordering effect can understate \state's utility relative to methods that return responses in arbitrary order.
Crucially, the raw quality tables show \state is strongest everywhere, confirming this is an ordering artifact rather than a genuine quality deficit.

\begin{table*}[h!]
\centering
\scriptsize
\setlength{\tabcolsep}{6pt}
\begin{tabular}{llccc}
\toprule
\textbf{Model} & \textbf{Method} & \textbf{Low} & \textbf{Medium} & \textbf{High} \\
\midrule
 \textbf{Ministral 3 14B} & Baseline & 4.12\,$\pm$\,0.04 & 4.9\,$\pm$\,0.08 & 5.67\,$\pm$\,0.06 \\
   & Baseline CoT & 5.01\,$\pm$\,0.09 & 5.59\,$\pm$\,0.06 & 5.99\,$\pm$\,0.11 \\
   & Baseline w/ Action Space & 4.11\,$\pm$\,0.06 & 4.84\,$\pm$\,0.11 & 5.42\,$\pm$\,0.09 \\
   & Baseline CoT w/ Action Space & \underline{5.62\,$\pm$\,0.12} & \underline{6.03\,$\pm$\,0.08} & 6.28\,$\pm$\,0.14 \\
   & Baseline ToT & 4.85\,$\pm$\,0.1 & 5.62\,$\pm$\,0.14 & \underline{6.32\,$\pm$\,0.14} \\
   & Baseline ToT w/ Action Space & 4.8\,$\pm$\,0.14 & 5.32\,$\pm$\,0.07 & 5.78\,$\pm$\,0.12 \\
   & STATe of Thoughts & \textbf{6.34\,$\pm$\,0.11} & \textbf{6.53\,$\pm$\,0.1} & \textbf{6.88\,$\pm$\,0.14} \\
\midrule
 \textbf{Qwen3 4B} & Baseline & 3.52\,$\pm$\,0.05 & 3.79\,$\pm$\,0.08 & 4.07\,$\pm$\,0.07 \\
   & Baseline CoT & 4.12\,$\pm$\,0.07 & 4.24\,$\pm$\,0.1 & 4.55\,$\pm$\,0.07 \\
   & Baseline w/ Action Space & 3.23\,$\pm$\,0.06 & 3.47\,$\pm$\,0.05 & 3.75\,$\pm$\,0.07 \\
   & Baseline CoT w/ Action Space & \textbf{4.37\,$\pm$\,0.08} & \textbf{4.51\,$\pm$\,0.1} & \underline{4.72\,$\pm$\,0.09} \\
   & Baseline ToT & 3.84\,$\pm$\,0.07 & 4.12\,$\pm$\,0.08 & 4.39\,$\pm$\,0.1 \\
   & Baseline ToT w/ Action Space & 3.78\,$\pm$\,0.08 & 4.03\,$\pm$\,0.08 & 4.3\,$\pm$\,0.07 \\
   & STATe of Thoughts & \underline{4.27\,$\pm$\,0.1} & \underline{4.4\,$\pm$\,0.08} & \textbf{4.75\,$\pm$\,0.06} \\
\midrule
 \textbf{Qwen3 8B} & Baseline & 4.61\,$\pm$\,0.07 & 4.78\,$\pm$\,0.1 & 4.87\,$\pm$\,0.09 \\
   & Baseline CoT & \textbf{4.77\,$\pm$\,0.13} & \underline{4.81\,$\pm$\,0.09} & \underline{4.99\,$\pm$\,0.08} \\
   & Baseline w/ Action Space & 4.34\,$\pm$\,0.1 & 4.46\,$\pm$\,0.1 & 4.52\,$\pm$\,0.1 \\
   & Baseline CoT w/ Action Space & 4.57\,$\pm$\,0.09 & 4.7\,$\pm$\,0.08 & 4.88\,$\pm$\,0.13 \\
   & Baseline ToT & 3.83\,$\pm$\,0.06 & 4.1\,$\pm$\,0.08 & 4.43\,$\pm$\,0.09 \\
   & Baseline ToT w/ Action Space & 3.57\,$\pm$\,0.1 & 3.82\,$\pm$\,0.08 & 4.14\,$\pm$\,0.09 \\
   & STATe of Thoughts & \underline{4.74\,$\pm$\,0.09} & \textbf{4.88\,$\pm$\,0.09} & \textbf{5.16\,$\pm$\,0.09} \\
\midrule
 \textbf{Qwen3 30B} & Baseline & 3.64\,$\pm$\,0.06 & 3.84\,$\pm$\,0.05 & 4.17\,$\pm$\,0.07 \\
   & Baseline CoT & 4.05\,$\pm$\,0.05 & 4.21\,$\pm$\,0.08 & 4.53\,$\pm$\,0.12 \\
   & Baseline w/ Action Space & 3.56\,$\pm$\,0.07 & 3.72\,$\pm$\,0.08 & 4.17\,$\pm$\,0.09 \\
   & Baseline CoT w/ Action Space & \underline{4.39\,$\pm$\,0.12} & \underline{4.68\,$\pm$\,0.12} & \underline{4.96\,$\pm$\,0.11} \\
   & Baseline ToT & 3.61\,$\pm$\,0.08 & 3.88\,$\pm$\,0.07 & 4.27\,$\pm$\,0.07 \\
   & Baseline ToT w/ Action Space & 3.79\,$\pm$\,0.08 & 4.13\,$\pm$\,0.05 & 4.47\,$\pm$\,0.09 \\
   & STATe of Thoughts & \textbf{4.98\,$\pm$\,0.08} & \textbf{5.12\,$\pm$\,0.08} & \textbf{5.3\,$\pm$\,0.1} \\
\midrule
 \textbf{Gemma 3 4B} & Baseline & 3.42\,$\pm$\,0.05 & 3.57\,$\pm$\,0.05 & 3.88\,$\pm$\,0.05 \\
   & Baseline CoT & 3.81\,$\pm$\,0.07 & 3.95\,$\pm$\,0.1 & 4.14\,$\pm$\,0.03 \\
   & Baseline w/ Action Space & 3.44\,$\pm$\,0.08 & 3.69\,$\pm$\,0.06 & 4.1\,$\pm$\,0.06 \\
   & Baseline CoT w/ Action Space & \underline{3.98\,$\pm$\,0.13} & \underline{4.2\,$\pm$\,0.08} & \textbf{4.47\,$\pm$\,0.1} \\
   & Baseline ToT & 3.46\,$\pm$\,0.07 & 3.71\,$\pm$\,0.06 & 3.98\,$\pm$\,0.09 \\
   & Baseline ToT w/ Action Space & 3.56\,$\pm$\,0.08 & 3.8\,$\pm$\,0.07 & 4.06\,$\pm$\,0.07 \\
   & STATe of Thoughts & \textbf{4.16\,$\pm$\,0.08} & \textbf{4.33\,$\pm$\,0.04} & \underline{4.45\,$\pm$\,0.08} \\
\midrule
 \textbf{Gemma 3 27B} & Baseline & 3.73\,$\pm$\,0.05 & 3.9\,$\pm$\,0.05 & 4.2\,$\pm$\,0.05 \\
   & Baseline CoT & 4.15\,$\pm$\,0.08 & 4.3\,$\pm$\,0.13 & 4.56\,$\pm$\,0.07 \\
   & Baseline w/ Action Space & 3.68\,$\pm$\,0.04 & 3.93\,$\pm$\,0.07 & 4.24\,$\pm$\,0.09 \\
   & Baseline CoT w/ Action Space & \underline{4.69\,$\pm$\,0.08} & \underline{4.89\,$\pm$\,0.09} & \underline{5.21\,$\pm$\,0.09} \\
   & Baseline ToT & 3.88\,$\pm$\,0.05 & 4.17\,$\pm$\,0.05 & 4.51\,$\pm$\,0.06 \\
   & Baseline ToT w/ Action Space & 4.26\,$\pm$\,0.09 & 4.49\,$\pm$\,0.08 & 4.77\,$\pm$\,0.06 \\
   & STATe of Thoughts & \textbf{4.97\,$\pm$\,0.09} & \textbf{5.08\,$\pm$\,0.1} & \textbf{5.33\,$\pm$\,0.07} \\
\midrule
 \textbf{Nemotron 3 30B} & Baseline & 4.36\,$\pm$\,0.07 & 4.73\,$\pm$\,0.08 & 5.36\,$\pm$\,0.11 \\
   & Baseline CoT & 4.85\,$\pm$\,0.1 & 5.06\,$\pm$\,0.08 & \underline{5.62\,$\pm$\,0.11} \\
   & Baseline w/ Action Space & 4.57\,$\pm$\,0.11 & 5.08\,$\pm$\,0.11 & 5.14\,$\pm$\,0.17 \\
   & Baseline CoT w/ Action Space & 5.05\,$\pm$\,0.13 & 5.28\,$\pm$\,0.12 & 5.51\,$\pm$\,0.18 \\
   & Baseline ToT & 4.52\,$\pm$\,0.11 & 4.95\,$\pm$\,0.06 & 5.46\,$\pm$\,0.11 \\
   & Baseline ToT w/ Action Space & \underline{5.08\,$\pm$\,0.12} & \underline{5.4\,$\pm$\,0.09} & 5.55\,$\pm$\,0.11 \\
   & STATe of Thoughts & \textbf{5.33\,$\pm$\,0.13} & \textbf{5.62\,$\pm$\,0.1} & \textbf{5.77\,$\pm$\,0.12} \\
\bottomrule
\end{tabular}
\caption{NoveltyBench Mean Utility for all models (mean$\pm$std over seeds). Low, medium, and high temperature correspond to $T=0.1$, $0.3$, $0.5$ for Ministral~3~14B, and $T=0.5$, $0.7$, $1.0$ for others. Best and second best per model per column are \textbf{bolded} and \underline{underlined}.}
\label{tab:noveltybench-curated-test-utility}
\end{table*}

\newpage

\subsubsection{Token Usage}
Table \ref{tab:noveltybench-compute-profile} reveals that \textit{input-token} usage increases with prompt complexity: I/O uses the simplest prompt, whereas ToT and \state require substantially more context.
Including the action space in the generator prompt further increases input-token usage, particularly when the action space is large.
The CoT variants generate the most explicit \textit{reasoning-output} tokens, with CoT + Actions producing substantially more than the search-based methods.
Much of the iterative computation performed by ToT and \state instead appears as reasoning-stage input, as intermediate search states are repeatedly expanded and evaluated.
\state generates only slightly more reasoning-output tokens than ToT and fewer than ToT + Actions.
Finally, \textit{output-token} usage shows that \state's improvements over the action-conditioned baselines do not result from longer answers.
\state produces fewer final-output tokens than both I/O + Actions and CoT + Actions.
This is particularly relevant given the known length bias in LLM-as-a-Judge evaluations \citep{dubois2024length}: \state's quality and utility advantages cannot be attributed to verbosity.

\begin{table*}[h!]
\centering
\scriptsize
\setlength{\tabcolsep}{2.4pt}
\renewcommand{\arraystretch}{1.12}
\begin{tabular*}{\textwidth}{@{\extracolsep{\fill}}lccrcr@{}}
\toprule
& \multicolumn{1}{c}{Problem}
& \multicolumn{2}{c}{Reasoning}
& \multicolumn{2}{c}{Final} \\
\cmidrule(lr){2-2}
\cmidrule(lr){3-4}
\cmidrule(l){5-6}
Method / preset
& \shortstack{Input\\cached / total (hit rate)}
& \shortstack{Input\\cached / total (hit rate)}
& \shortstack{Output\\tokens}
& \shortstack{Input\\cached / total (hit rate)}
& \shortstack{Output\\tokens} \\
\midrule
I/O
& 149.3 / 160.3 (93.2\%)
& N/A
& N/A
& N/A
& 328.4 \\
I/O + Actions
& 401.1 / 427.3 (93.9\%)
& N/A
& N/A
& N/A
& 988.5 \\
\addlinespace[2pt]
CoT
& 175.5 / 184.3 (95.2\%)
& N/A
& 955.0
& N/A
& 200.3 \\
CoT + Actions
& 414.4 / 451.3 (91.8\%)
& N/A
& 1565.3
& N/A
& 557.5 \\
\addlinespace[2pt]
ToT
& 462.2 / 479.3 (96.4\%)
& 5951.3 / 6437.7 (92.4\%)
& 311.0
& 3829.9 / 3935.6 (97.3\%)
& 232.1 \\
ToT + Actions
& 704.5 / 750.3 (93.9\%)
& 9506.3 / 10806.0 (88.0\%)
& 591.0
& 6222.8 / 6349.0 (98.0\%)
& 550.3 \\
\addlinespace[2pt]
\state
& 568.2 / 646.4 (87.9\%)
& 4018.3 / 4517.6 (88.9\%)
& 378.4
& 5475.9 / 5609.8 (97.6\%)
& 494.0 \\
\bottomrule
\end{tabular*}
\caption{
    \textbf{Per-example usage on NoveltyBench with Qwen3-30B.}
    Input cells report cached tokens, total input tokens, and cache-hit rate in parentheses.
    Reasoning-output and final-output columns report generated tokens.
    ``N/A'' denotes a stage that is not instantiated by the corresponding method. All entries are means over the profiling run.
    ``Actions'' denotes configurations in which the action space is included directly in the generator prompt.
}
\label{tab:noveltybench-compute-profile}
\end{table*}

\begin{figure}[h!]
\centering
\includegraphics[width=0.7\linewidth]{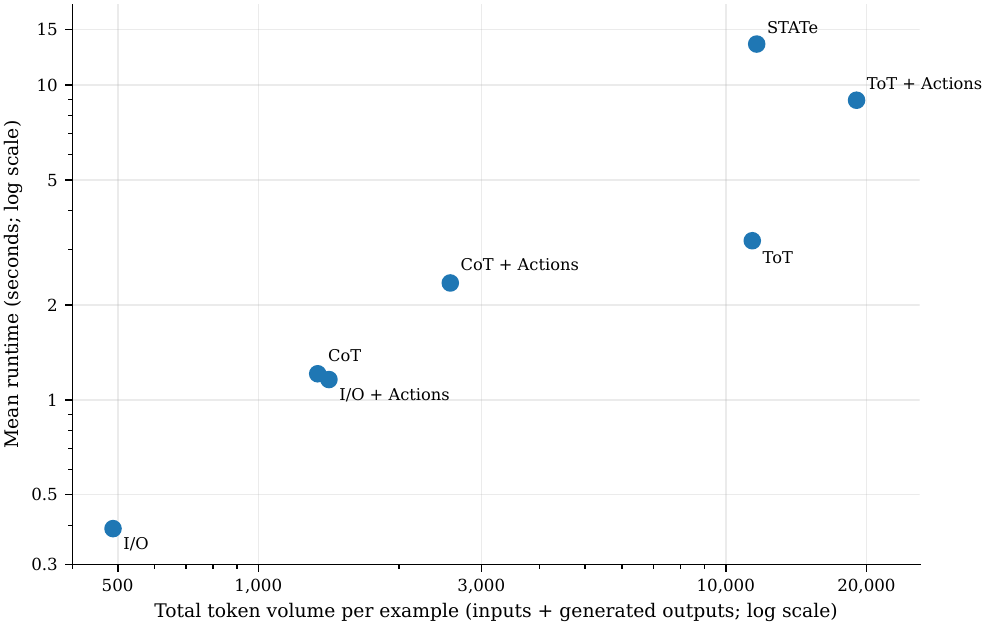}
\caption{\textbf{Runtime versus token volume on NoveltyBench.} Each point denotes an inference method or preset. Total token volume is the sum of all input tokens and generated reasoning and final-output tokens per example. Both axes use logarithmic scales.}
\label{fig:noveltybench-runtime-vs-token-volume}
\end{figure}

Figure~\ref{fig:noveltybench-runtime-vs-token-volume} relates wall-clock runtime to total token volume, defined as the sum of all input tokens and generated reasoning and final-output tokens.
% Both axes are logarithmic because the methods span more than an order of magnitude in both quantities.
Runtime generally increases with token volume, but token volume does not fully explain the latency differences between methods.
In particular, \state and ToT process similar total token volumes, approximately $11.6$K and $11.4$K tokens per example, respectively, yet their mean runtimes are $13.48$ and $3.20$ seconds.
The additional latency for \state arises primarily from its reranker controller.
% For each frontier state, the controller scores the available actions and selects which action should guide the next expansion.
ToT and the other baselines do not invoke this separate controller, so \state incurs controller-inference latency beyond the cost represented by generator-side token volume alone.

Prompt caching nevertheless eliminates most repeated input processing.
Approximately $93.4\%$ of \state's input tokens are cache hits.
This is close to ToT's aggregate cache-hit rate of approximately $94.4\%$ and higher than the approximately $91.8\%$ rate obtained by ToT + Actions.
At the stage level, \state obtains cache-hit rates of $87.9\%$ for the problem input, $88.9\%$ for reasoning input, and $97.6\%$ for final input.
The near-complete cache reuse during final synthesis indicates that \state efficiently reuses its long shared prefixes.
Consequently, \state's additional runtime is better explained by its controller calls than by broadly inefficient prompt-cache use.

\newpage

\subsection{Controllability Evaluation}
\label{app:controllability-evaluation}

We evaluate controllability on a subset of 1,000 arguments generated on a fixed topic.
An LLM judge (GPT-5-mini; \citealp{openai_gpt5_systemcard_2025}) performs up to 13 boolean checks per argument, split across two separate calls to prevent cross-contamination between the reasoning trace and the final output.

\subsubsection{Step-Level Evaluation Prompt}

The step-level call receives only the three reasoning steps (the final argument is withheld).
For each step, it checks two properties: (i) whether the step exhibits its prescribed discourse structure (verified via a prescribed prefix word) and (ii) whether the step discusses its prescribed subtopic (verified via the guidance description).
This yields 6 boolean checks in total.

\begin{tcolorbox}[
  after skip=2pt,
  colback=gray!8, colframe=gray!55, arc=4pt,
  title={\small\bfseries\sffamily System Message},
  fonttitle=\color{white}, coltitle=white, colbacktitle=gray!55,
  left=6pt, right=6pt, top=4pt, bottom=4pt,
]
\begin{lstlisting}[
  basicstyle=\ttfamily\scriptsize,
  breaklines=true,
]
You are an expert evaluator of argumentative text. You will be given text
along with descriptions of prescribed properties. For each check, determine
whether the text exhibits the described property.

Each property may include:
- A 'prefix': the prescribed opening word(s) that the text was instructed
  to start with (e.g., 'However', 'Therefore').
- A 'guidance': the internal instruction given to the model describing what
  angle or lens to reason through. The text should reflect this guidance
  in its content and framing.

Respond with a JSON object containing exactly the keys specified, each with
a boolean value (true or false). Do not include any other text.
\end{lstlisting}
\end{tcolorbox}
\promptcaption{lst:controllability-step-system}{Controllability Step-Level Judge --- System Message}
\begin{tcolorbox}[
  after skip=2pt,
  colback=gray!8, colframe=gray!55, arc=4pt,
  title={\small\bfseries\sffamily User Message},
  fonttitle=\color{white}, coltitle=white, colbacktitle=gray!55,
  left=6pt, right=6pt, top=4pt, bottom=4pt,
]
\begin{lstlisting}[
  basicstyle=\ttfamily\scriptsize,
  breaklines=true,
  moredelim={**[is][\color{teal}]{@T}{T@}},
  moredelim={**[is][\color{blue}]{@B}{B@}},
]
Topic: @B{topic}B@
Stance: @B{stance}B@

=== Step 1 ===
@B{step_1_text}B@

=== Step 2 ===
@B{step_2_text}B@

=== Step 3 ===
@B{step_3_text}B@

=== Evaluation Checks ===
For each check below, respond true if the text clearly exhibits the
described property, or false otherwise.

step_1_has_structure: Does Step 1 exhibit the discourse structure
  '@T{structure_1}T@' -- defined as: '@T{structure_1_def}T@'
  (prescribed prefix: '@T{structure_1_prefix}T@')?
step_1_has_subtopic: Does Step 1 discuss the subtopic
  '@T{subtopic_1}T@' -- defined as: '@T{subtopic_1_def}T@'
  (guidance: '@T{subtopic_1_guidance}T@')?
step_2_has_structure: ...
step_2_has_subtopic: ...
step_3_has_structure: ...
step_3_has_subtopic: ...

Respond with a JSON object containing exactly these 6 keys, each
with a boolean value:
step_1_has_structure, step_1_has_subtopic, step_2_has_structure,
step_2_has_subtopic, step_3_has_structure, step_3_has_subtopic
\end{lstlisting}
\end{tcolorbox}
\promptcaptionnote{lst:controllability-step-user}{Controllability Step-Level Judge --- User Message}{\textcolor{teal}{Teal}: action-space configuration (structure definitions, subtopic descriptions, prescribed prefixes). \textcolor{blue}{Blue}: concrete inference-time values.}

\subsubsection{Final-Level Evaluation Prompt}

The final-level call receives only the finished argument (the reasoning steps are withheld).
It checks whether the argument reflects each step's prescribed structure and subtopic, and whether the prescribed step ordering is preserved in the argument.
This yields 7 boolean checks: 6 mirroring the step-level checks (applied to the final text) plus one order-preservation check.

\begin{tcolorbox}[
  after skip=2pt,
  colback=gray!8, colframe=gray!55, arc=4pt,
  title={\small\bfseries\sffamily User Message},
  fonttitle=\color{white}, coltitle=white, colbacktitle=gray!55,
  left=6pt, right=6pt, top=4pt, bottom=4pt,
]
\begin{lstlisting}[
  basicstyle=\ttfamily\scriptsize,
  breaklines=true,
  moredelim={**[is][\color{teal}]{@T}{T@}},
  moredelim={**[is][\color{blue}]{@B}{B@}},
]
Topic: @B{topic}B@
Stance: @B{stance}B@

=== Argument ===
@B{argument_text}B@

=== Prescribed Properties ===
This argument was generated using three sequential reasoning steps. Each
step was prescribed a discourse structure and a subtopic. Evaluate whether
the argument reflects these prescriptions.

Step 1 prescription -- structure: '@T{structure_1}T@' ('@T{structure_1_def}T@',
  prefix: '@T{structure_1_prefix}T@'), subtopic: '@T{subtopic_1}T@'
  ('@T{subtopic_1_def}T@', guidance: '@T{subtopic_1_guidance}T@')
Step 2 prescription -- ...
Step 3 prescription -- ...

=== Evaluation Checks ===
For each check below, respond true if the argument clearly exhibits the
described property, or false otherwise.

final_has_structure_1: Does the argument contain content reflecting the
  structure '@T{structure_1}T@'?
final_has_subtopic_1: Does the argument contain content reflecting the
  subtopic '@T{subtopic_1}T@'?
final_has_structure_2: ...
final_has_subtopic_2: ...
final_has_structure_3: ...
final_has_subtopic_3: ...
final_preserves_order: Does the argument contain all three steps'
  prescribed content, presented in the correct order (Step 1 material
  appears before Step 2, which appears before Step 3)?

Respond with a JSON object containing exactly these 7 keys, each with a
boolean value:
final_has_structure_1, final_has_subtopic_1, final_has_structure_2,
final_has_subtopic_2, final_has_structure_3, final_has_subtopic_3,
final_preserves_order
\end{lstlisting}
\end{tcolorbox}
\promptcaptionnote{lst:controllability-final-user}{Controllability Final-Level Judge --- User Message}{The same system message as Prompt~\ref{lst:controllability-step-system} is reused. \textcolor{teal}{Teal}: action-space configuration. \textcolor{blue}{Blue}: concrete inference-time values.}

\subsubsection{Controllability Results Across Synthesis Modes}
\label{app:controllability_by_synthesis}

The three synthesis modes (Appendix~\ref{app:synthesis-modes}) serve as ablations along a control-quality continuum.
\textbf{Strict} synthesis produces near-verbatim concatenations of reasoning steps, yielding the strongest attribution between actions and output text.
\textbf{Faithful} synthesis permits light rephrasing while preserving the structure and ordering of reasoning steps.
\textbf{Restructured} synthesis allows free reorganization of the reasoning content, providing the weakest action attribution but the best prose fluency.
This gradient reveals how synthesis flexibility affects controllability: as the model is given more freedom to deviate from the reasoning trace, step-level fidelity may be preserved while final-argument fidelity degrades.

\begin{figure}[ht]
  \centering
  \includegraphics[width=\textwidth]{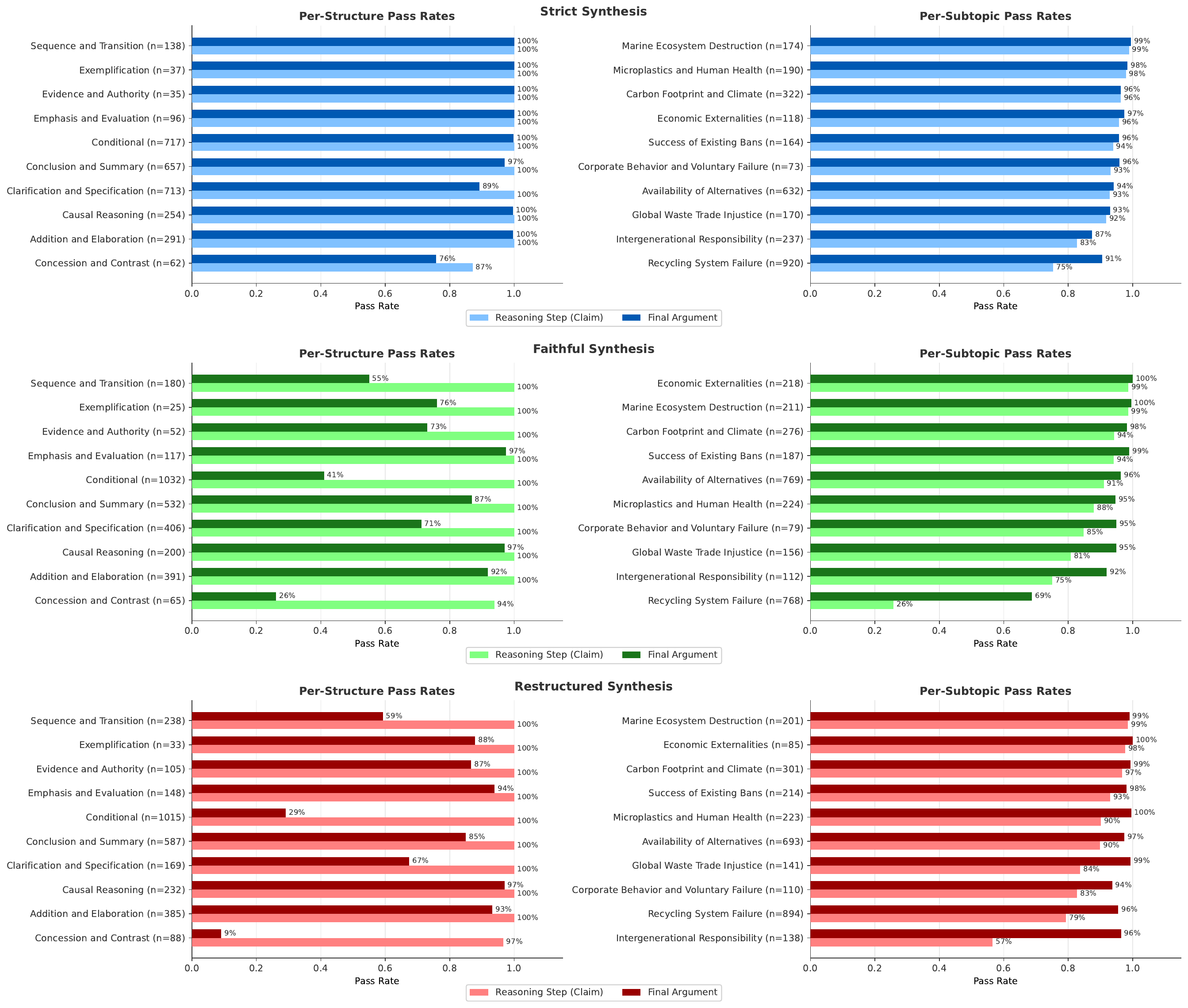}
  \caption{Controllability pass rates across synthesis modes (plastic pollution topic). Each group of bars shows a different evaluation check. Strict synthesis achieves the highest overall pass rate, particularly for final-level structure and order preservation. Restructured synthesis maintains high subtopic adherence but loses structural fidelity at the final-argument level.}
  \label{fig:controllability-unified}
\end{figure}

Table~\ref{tab:controllability-results} extends Table~\ref{tab:controllability-strict} with pass rate breakdowns across all three synthesis modes, and Figure~\ref{fig:controllability-unified} provides further detail by action space dimension. All evaluations use plastic pollution as a case study. Strict synthesis achieves the highest pass rates overall, as expected. However, the envisioned gradient from strict to restructured does not fully manifest: faithful synthesis unexpectedly achieves lower pass rates than restructured synthesis, particularly for the subtopic \textit{recycling system failure}. This suggests that strict synthesis operates reliably, but the distinction between faithful and restructured modes requires further investigation and potential fine-tuning.

\begin{table}[ht]
\centering
\small
\begin{tabular}{@{}l ccc@{}}
\toprule
& \textbf{Strict} & \textbf{Faithful} & \textbf{Restructured} \\
\midrule
\textbf{Overall (all 13 checks)} & \textbf{93.8} { [93.3, 94.4]} & \textbf{81.4} { [80.6, 82.2]} & \textbf{86.2} { [85.6, 86.9]} \\
\midrule
Step structure avg    & 99.7 { [99.5, 99.9]} & 99.9 { [99.7, 100.0]} & 99.9 { [99.8, 100.0]} \\
Step subtopic avg     & 87.8 { [86.6, 89.0]} & 74.4 { [72.9, 75.8]} & 86.4 { [85.0, 87.7]} \\
Final structure avg   & 96.2 { [95.5, 96.9]} & 67.1 { [65.1, 69.1]} & 63.4 { [61.4, 65.3]} \\
Final subtopic avg    & 93.5 { [92.6, 94.4]} & 89.7 { [88.6, 90.8]} & 97.3 { [96.8, 97.9]} \\
Final preserves order & 87.9 { [85.9, 89.9]} & 64.7 { [61.7, 67.6]} & 80.0 { [77.5, 82.5]} \\
\bottomrule
\end{tabular}
\caption{Controllability evaluation results across synthesis modes. An LLM judge (GPT-5-mini) performs 13 boolean checks per argument (6 step-level, 7 final-level) on 1,000 arguments per synthesis type. Values are pass rates (\%) with 95\% bootstrap CIs (resampling arguments, $B = 10{,}000$).}
\label{tab:controllability-results}
\end{table}

\subsubsection{Effect of Subtopic Specificity on Controllability}
\label{app:controllability-subtopic-specificity}

The controllability results above use \emph{topic-specific} subtopics designed for the plastic pollution domain (e.g., \textit{marine ecosystem destruction}, \textit{recycling system failure}; Table~\ref{tab:action-space-content-plastic-waste}).
To test whether action-space specificity itself affects controllability, we repeat the strict-synthesis evaluation with \emph{generic} subtopics, consisting of ten topic-independent argumentative lenses such as \textit{justice and fairness} or \textit{ethical principles} (Table~\ref{tab:action-space-generic-subtopics}), applied to the same plastic pollution topic.
The generic evaluation covers 500 arguments (6,500 total checks), compared with 1,000 arguments (13,000 checks) for the topic-specific evaluation.

\begin{figure}[ht]
  \centering
  \includegraphics[width=\textwidth]{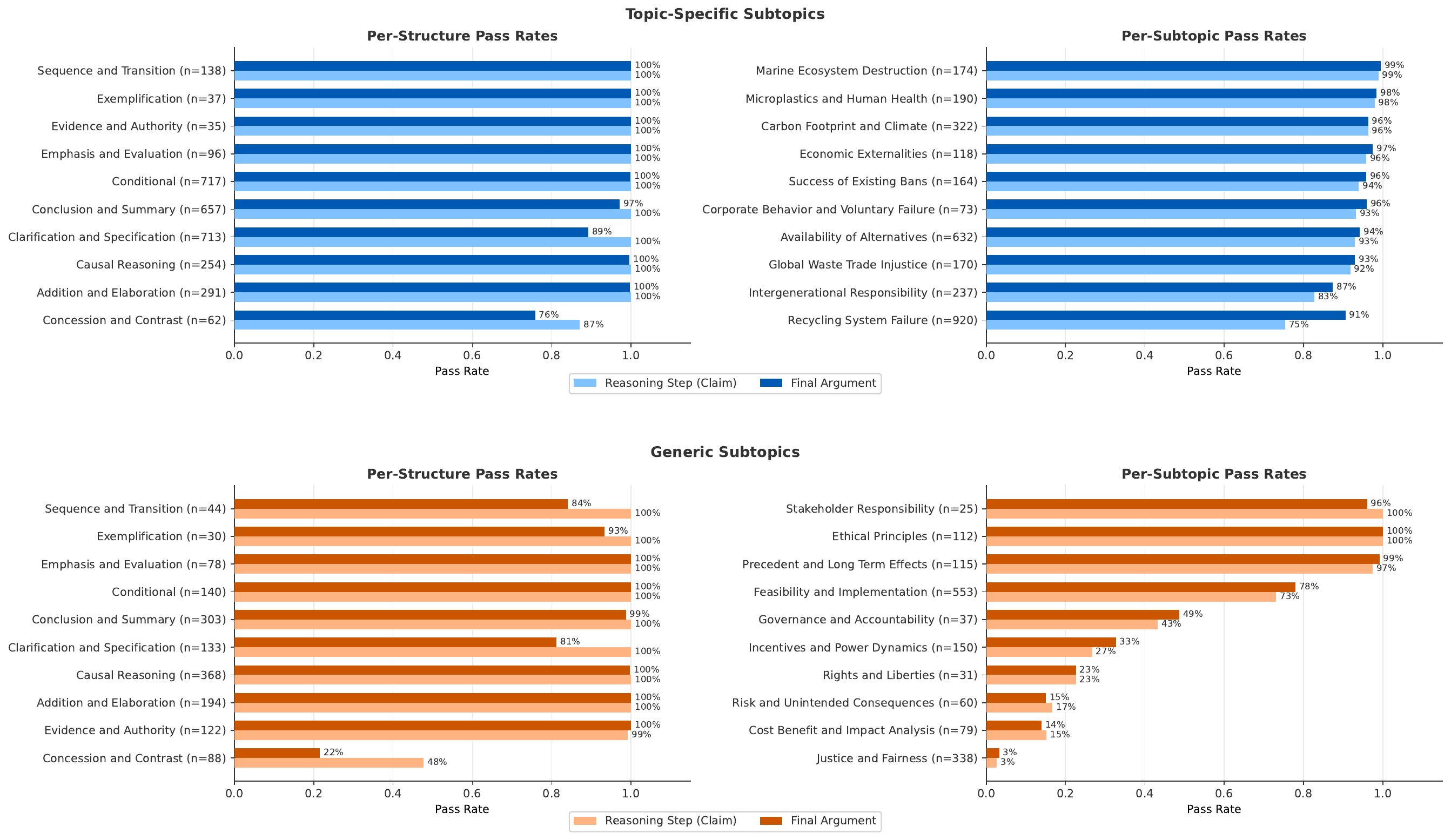}
  \caption{Controllability comparison between topic-specific subtopics (top, blue; $n{=}1{,}000$ arguments) and generic subtopics (bottom, orange; $n{=}500$ arguments), both under strict synthesis on the plastic pollution topic.
  Structure adherence is comparable across both designs, but subtopic adherence drops substantially with generic categories.}
  \label{fig:controllability-generic-vs-specific}
\end{figure}

% Subtopic adherence drops substantially with generic subtopics; from 87.8\% to 49.8\% at the step level and from 93.5\% to 52.4\% at the response level (Figure~\ref{fig:controllability-generic-vs-specific}).
Subtopic adherence drops substantially with generic subtopics: from 87.8\% to 49.8\% at the step level and from 93.5\% to 52.4\% at the response level (Figure~\ref{fig:controllability-generic-vs-specific}).
Structure adherence, by contrast, remains comparable (96.9\% vs.\ 99.7\% step-level; 92.8\% vs.\ 96.2\% response-level), confirming that the prefix mechanism is robust regardless of subtopic design.
Per-subtopic variance is also markedly higher with generic categories. \textit{Justice and fairness} achieves only 2.7\% step-level adherence, while the lowest-performing specific subtopic (\textit{recycling system failure}) still reaches 75.3\%.

We conclude that topic-specific subtopics yield higher controllability because concrete, domain-grounded categories are semantically distinct and easier for the generator to follow and for the judge to detect.
Vague categories such as \textit{justice and fairness} overlap semantically with many arguments about plastic pollution, making adherence harder to enforce and verify.
This finding reinforces the practitioner guidance in Appendix~\ref{app:action-space-guidance}.
When designing content dimensions, practitioners should prefer specific, potentially topic-grounded categories over abstract, generic lenses.

\subsection{Predictability of Argument Quality from Actions}
\label{app:argument-generation-additional-results}

\subsubsection{Argument Generation}
\label{app:argument-evaluation}
The argument generation task is defined by a proposition (topic statement) and stance (pro vs. anti). We evaluate argument quality across five debate topics, each assigned to one of three LLM judges for pairwise comparisons (Table~\ref{tab:topic-overview}).

\begin{table}[ht]
\centering
\small
\footnotesize
\begin{tabular}{p{6.5cm}lll}
\toprule
\textbf{Proposition (stance: PRO)} & \textbf{Topic} & \textbf{LLM Judge} \\
\midrule
The government should enforce a total ban on single-use plastics. & Plastic Pollution & GPT-5-mini \\
Social media platforms should enforce a minimum age restriction of 16. & Social Media Restriction & GPT-5-mini \\
The government should implement a universal basic income program. & Universal Basic Income & GPT-5-mini \\
Standardized testing should be abolished as a primary measure of student performance. & Standardized\ Testing & Gemini-3.1-Flash-Lite \\
A special tax should be imposed on meat products to reduce consumption. & Meat Tax & Claude-Haiku-4.5 \\
\bottomrule
\end{tabular}
\caption{Five debate topics with their assigned LLM judges for pairwise quality evaluation.
The controllability evaluation (Section~\ref{sec:results-argument-generation-controllability}) and targeted trajectory
exploration (Section~\ref{sec:results-targeted-trajectory}) are conducted on the plastic pollution
topic as a detailed case study.}
\label{tab:topic-overview}
\end{table}

The main results in Section~\ref{sec:experiments-argument-generation} use strict synthesis.
Here, we additionally report ablation results for the faithful and restructured synthesis modes introduced in Appendix~\ref{app:synthesis-modes}, to assess the robustness of our attribution findings.
For each mode, we generate 5,000 arguments using 20 random-seed trees (depth 3, beam width 250), then sample 50,000 random argument pairs for judge comparisons and fit a Bradley--Terry model to obtain quality rankings.
We standardize these rankings and use them as our response variable.
Further, we randomly split each dataset into training (60\%) and test (40\%) partitions at the level of individual arguments, so each argument and its trajectory appear in exactly one partition.
The split is not grouped at the tree level.
As each dataset comprises 20 tree-search runs of 250 arguments each, and sibling arguments from the same tree share partial reasoning trajectories, overlapping trajectories can appear on both sides of the split.
Test performance should therefore be read as predictive accuracy for new arguments from the same generation process, rather than generalization to independently grown trees.

The sequential model (M2) uses two action dimensions, content and structure, with features that encode (i) action identities by position, (ii) within-step content-structure interactions, and (iii) within-dimension transitions of length two across steps.
These are compared against a length-only baseline (M0) and presence-based models (M1a--c).
All models include argument length in characters as a feature.
In this multi-dimensional setting, the feature space is high-dimensional and sparse, which increases overfitting risk and numerical instability.
To address this, we use LASSO regression for the M2 models \citep{tibshirani1996regression}, with $\alpha$ selected by grid search and 10-fold cross-validation, so irrelevant coefficients are shrunk toward zero:

\begin{equation}
\hat{\boldsymbol{\beta}} = \arg\min_{\boldsymbol{\beta}} \left\{ \frac{1}{2N} \sum_{i=1}^{N} (Y_i - \mathbf{x}^{\text{sequential}}_i \boldsymbol{\beta})^2 + \alpha \|\boldsymbol{\beta}\|_1 \right\}
\end{equation}

where $\mathbf{x}^{\text{sequential}}_i$ is the sequential feature encodings for the M2 models.
For M0 and M1a--c, we use ordinary least squares.
We report bootstrap 95\% confidence intervals (1,000 resamples of individual test arguments) for the test-set $R^2$.
M2's feature space strictly refines the presence-only features, which enables a direct comparison of whether sequential structure improves predictability.

\begin{table*}[htbp]
\centering
\scriptsize
\begin{tabular}{llcccccc}
\toprule
Topic & Synthesis & M0 R$^2$ & M1a R$^2$ & M1b R$^2$ & M1c R$^2$ & M2 R$^2$ & \\
\midrule
Plastic Pollution & Strict & 0.478\,$\pm$\,0.027 & 0.539\,$\pm$\,0.028 & 0.627\,$\pm$\,0.023 & 0.697\,$\pm$\,0.022 & \textbf{0.760\,$\pm$\,0.018} \\
 & Faithful & 0.345\,$\pm$\,0.034 & 0.363\,$\pm$\,0.035 & 0.610\,$\pm$\,0.025 & 0.615\,$\pm$\,0.025 & \textbf{0.626\,$\pm$\,0.024} \\
 & Restructured & 0.200\,$\pm$\,0.032 & 0.220\,$\pm$\,0.032 & 0.611\,$\pm$\,0.025 & 0.613\,$\pm$\,0.024 & \textbf{0.614\,$\pm$\,0.024}\\
\midrule
Social Media & Strict & 0.613\,$\pm$\,0.025 & 0.669\,$\pm$\,0.024 & 0.695\,$\pm$\,0.020 & 0.748\,$\pm$\,0.018 & \textbf{0.771\,$\pm$\,0.017} \\
 Restriction & Faithful & 0.294\,$\pm$\,0.033 & 0.304\,$\pm$\,0.032 & 0.620\,$\pm$\,0.025 & 0.621\,$\pm$\,0.024 & \textbf{0.634\,$\pm$\,0.024} \\
 & Restructured & 0.321\,$\pm$\,0.033 & 0.320\,$\pm$\,0.034 & 0.608\,$\pm$\,0.025 & 0.608\,$\pm$\,0.025 & \textbf{0.609\,$\pm$\,0.027}  \\
\midrule
Universal Basic & Strict & 0.536\,$\pm$\,0.029 & 0.618\,$\pm$\,0.027 & 0.644\,$\pm$\,0.023 & 0.732\,$\pm$\,0.020 & \textbf{0.790\,$\pm$\,0.017} \\
 Income & Faithful & 0.311\,$\pm$\,0.032 & 0.338\,$\pm$\,0.033 & 0.704\,$\pm$\,0.022 & 0.711\,$\pm$\,0.021 & \textbf{0.734\,$\pm$\,0.021} \\
 & Restructured & 0.232\,$\pm$\,0.028 & 0.276\,$\pm$\,0.031 & 0.772\,$\pm$\,0.016 & 0.773\,$\pm$\,0.016 & \textbf{0.799\,$\pm$\,0.016} \\
\midrule
Standardized & Strict & 0.536\,$\pm$\,0.029 & 0.666\,$\pm$\,0.022 & 0.569\,$\pm$\,0.028 & 0.681\,$\pm$\,0.022 & \textbf{0.802\,$\pm$\,0.016}\\
 Testing & Faithful & 0.416\,$\pm$\,0.027 & 0.482\,$\pm$\,0.027 & 0.532\,$\pm$\,0.026 & 0.584\,$\pm$\,0.025 & \textbf{0.634\,$\pm$\,0.023}  \\
 & Restructured & 0.410\,$\pm$\,0.031 & 0.432\,$\pm$\,0.030 & 0.609\,$\pm$\,0.025 & 0.610\,$\pm$\,0.025 & \textbf{0.625\,$\pm$\,0.024} \\
\midrule
Meat Tax & Strict & 0.632\,$\pm$\,0.029 & 0.678\,$\pm$\,0.028 & 0.641\,$\pm$\,0.028 & 0.684\,$\pm$\,0.028 & \textbf{0.726\,$\pm$\,0.027} \\
 & Faithful & 0.342\,$\pm$\,0.032 & 0.351\,$\pm$\,0.032 & 0.401\,$\pm$\,0.031 & 0.398\,$\pm$\,0.032 & \textbf{0.408\,$\pm$\,0.035} \\
 & Restructured & 0.406\,$\pm$\,0.032 & 0.428\,$\pm$\,0.032 & 0.460\,$\pm$\,0.030 & 0.462\,$\pm$\,0.031 & \textbf{0.484\,$\pm$\,0.030} \\
\bottomrule
\end{tabular}
\caption{Model comparison with argument length control across topics. R$^2$ values on held-out test set (40\%) with 95\% bootstrap CI ($\pm$ half-width). M0 uses only argument length (characters). All other models additionally include length.}
\label{tab:argument_generation_results_length}
\end{table*}

\subsubsection{Predictability Results Across Synthesis Modes}
\label{app:predictability-by-synthesis}

\begin{figure}[ht]
  \centering
  \includegraphics[width=\textwidth]{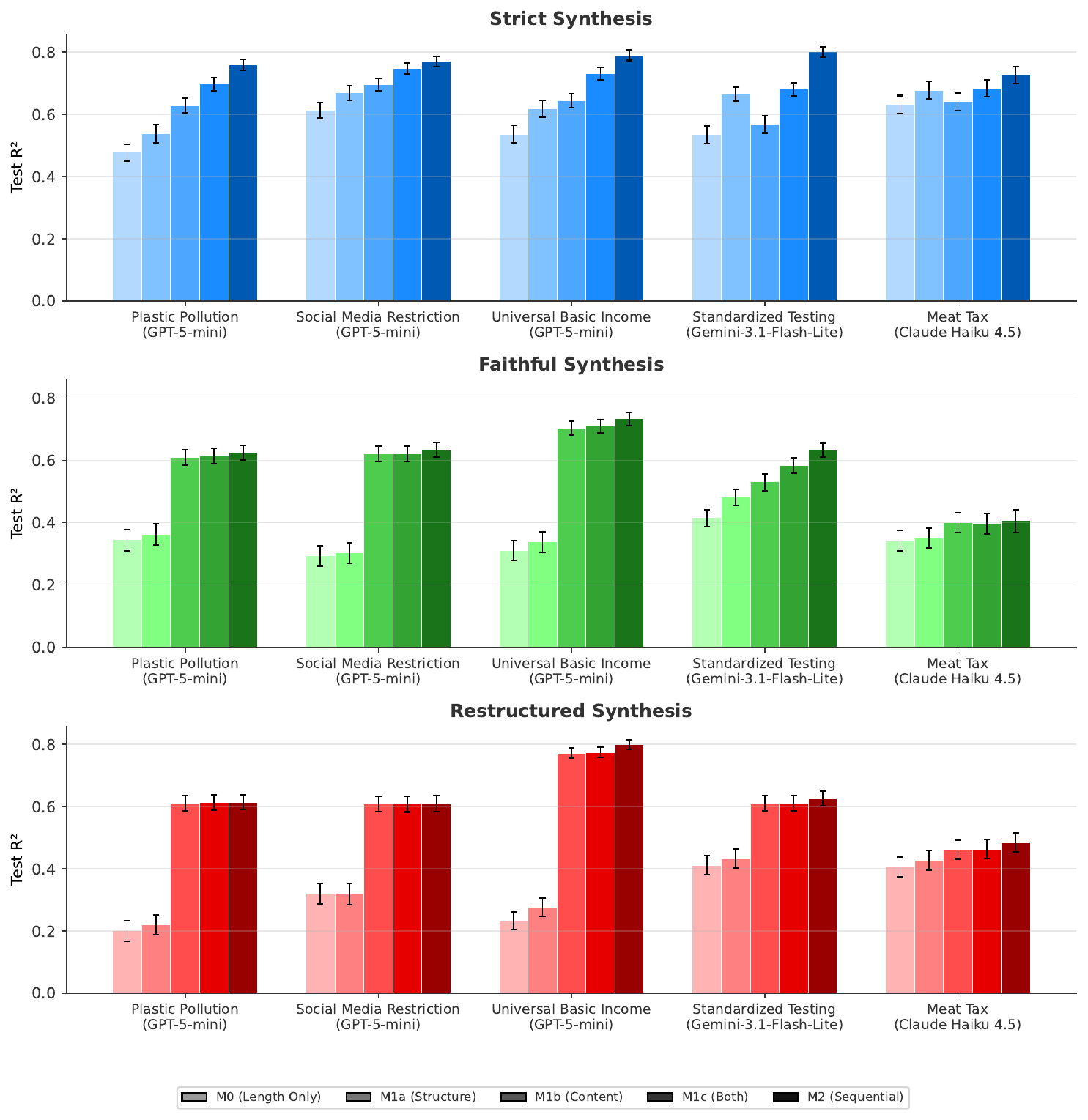}
  \caption{Predictability of argument quality from controller actions across all five debate topics, shown separately for each synthesis mode (with length control). Each row corresponds to one synthesis mode (Strict, Faithful, Restructured); within each row, bars show test-set R$^2$ (with 95\% bootstrap CIs) for M0 (length only), M1a (structure presence), M1b (content presence), M1c (both), and M2 (full sequential model).}
  \label{fig:cross-topic-by-synthesis}
\end{figure}

Figure~\ref{fig:cross-topic-by-synthesis} presents how well controller actions predict argument quality across all five debate topics, shown separately for each synthesis mode.
The consistent pattern across topics is that the sequential model (M2) outperforms presence-based baselines (M1a--c) and the length-only baseline (M0), confirming that action ordering and transitions carry predictive information beyond simple topic presence.  
Strict synthesis yields the highest predictability in most topics, while restructured synthesis generally shows the lowest, reflecting the control-fidelity gradient observed in the controllability evaluation.
Notably, the relative gain of M2 over presence-based models is largest for strict synthesis, where action sequences are most faithfully preserved.

\begin{figure}[ht]
  \centering
  \includegraphics[width=\textwidth]{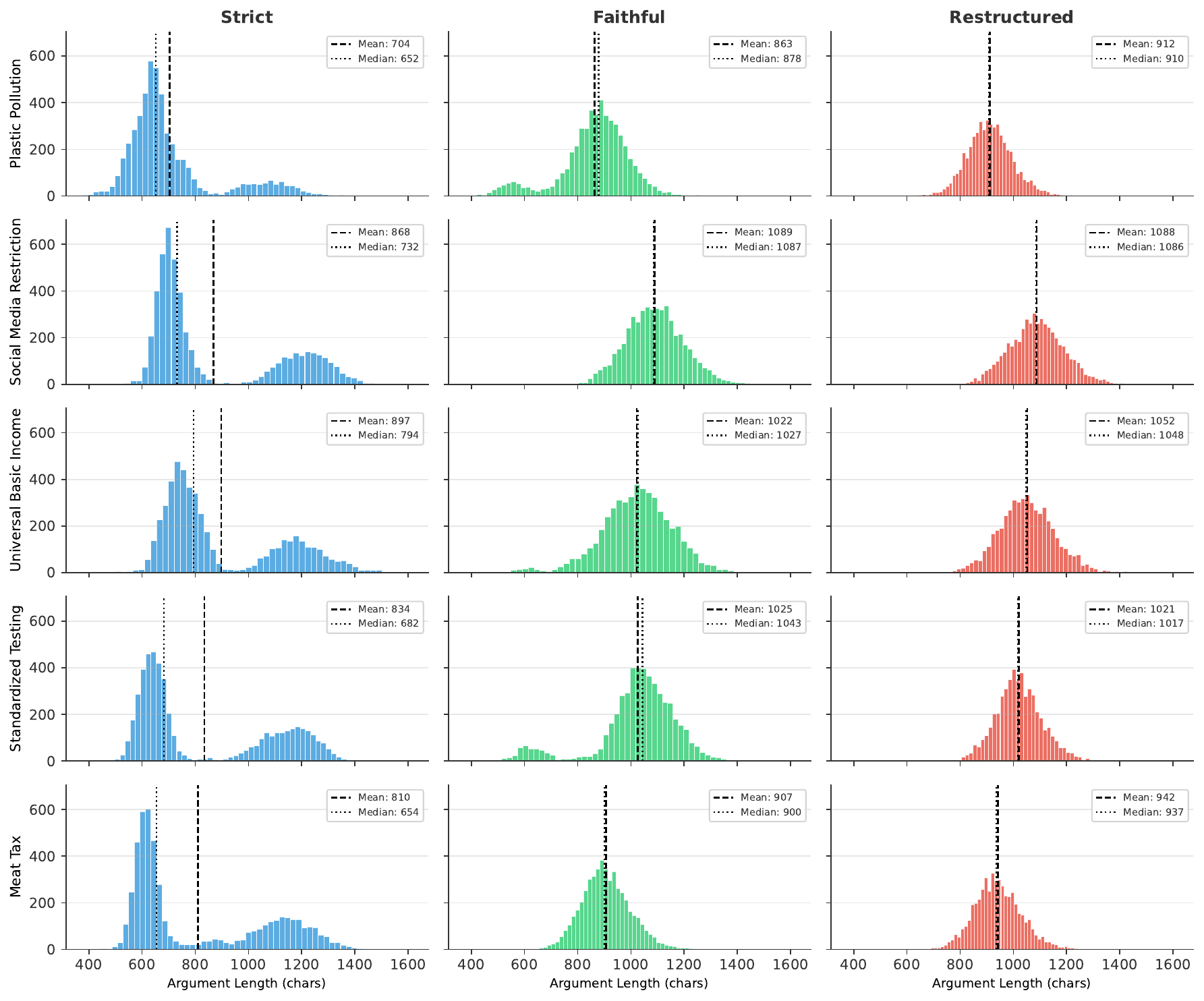}
  \caption{Distribution of argument lengths (in characters) across all five debate topics and three synthesis modes. Columns correspond to synthesis modes (strict, faithful, restructured); rows correspond to debate topics. Dashed and dotted vertical lines indicate the mean and median length, respectively.}
  \label{fig:unified-length-histograms}                     
\end{figure}

\begin{figure}[ht]
  \centering
  \includegraphics[width=\textwidth]{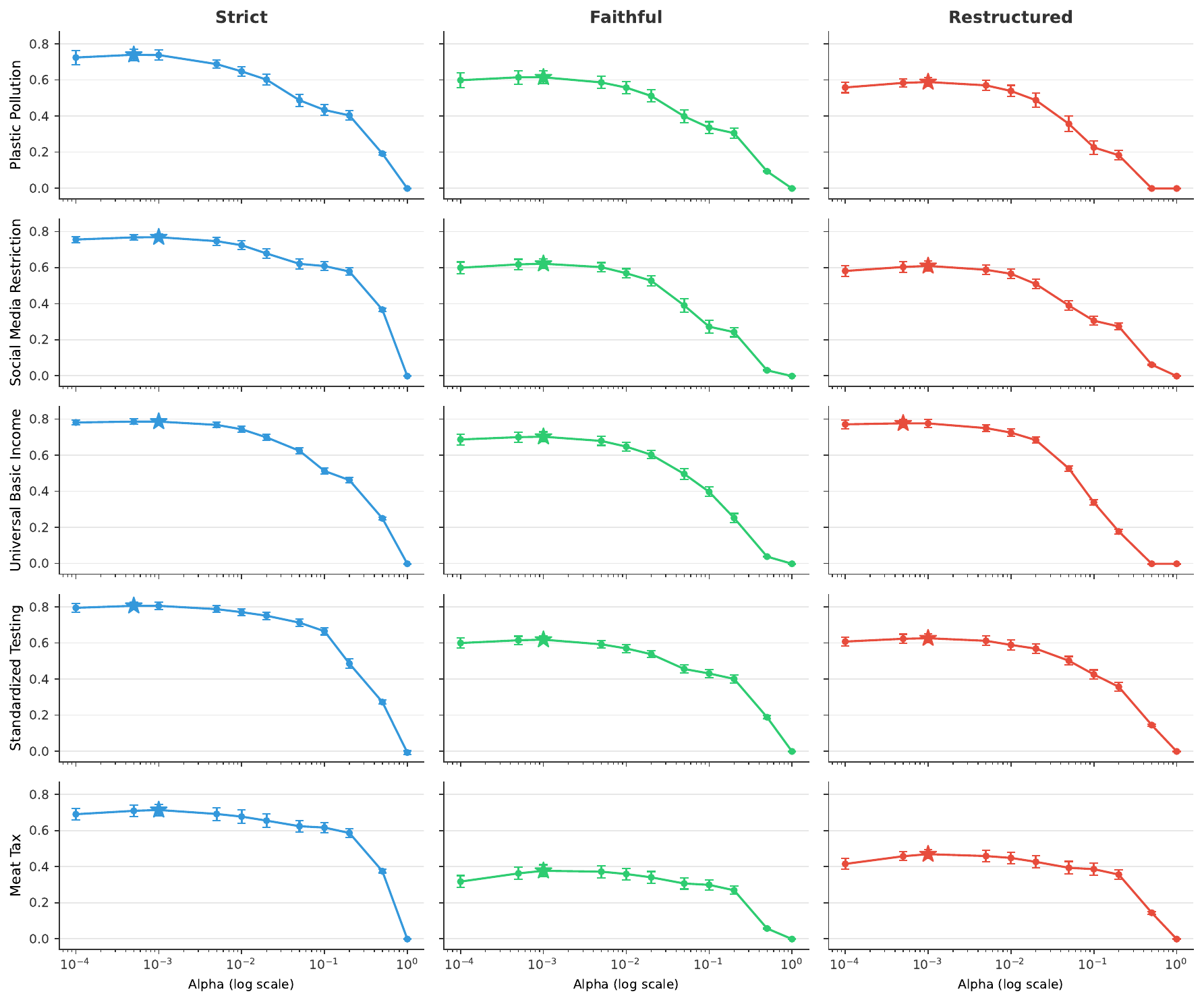}
  \caption{Cross-validation R$^2$ as a function of LASSO regularization parameter $\alpha$ across all five topics.
  Columns correspond to synthesis modes (strict, faithful, restructured); rows correspond to debate topics.
  Stars indicate the selected $\alpha$ for each configuration.}
  \label{fig:unified-m2-alpha-selection}                     
\end{figure}

\begin{figure}[ht]
  \centering
  \includegraphics[width=\textwidth]{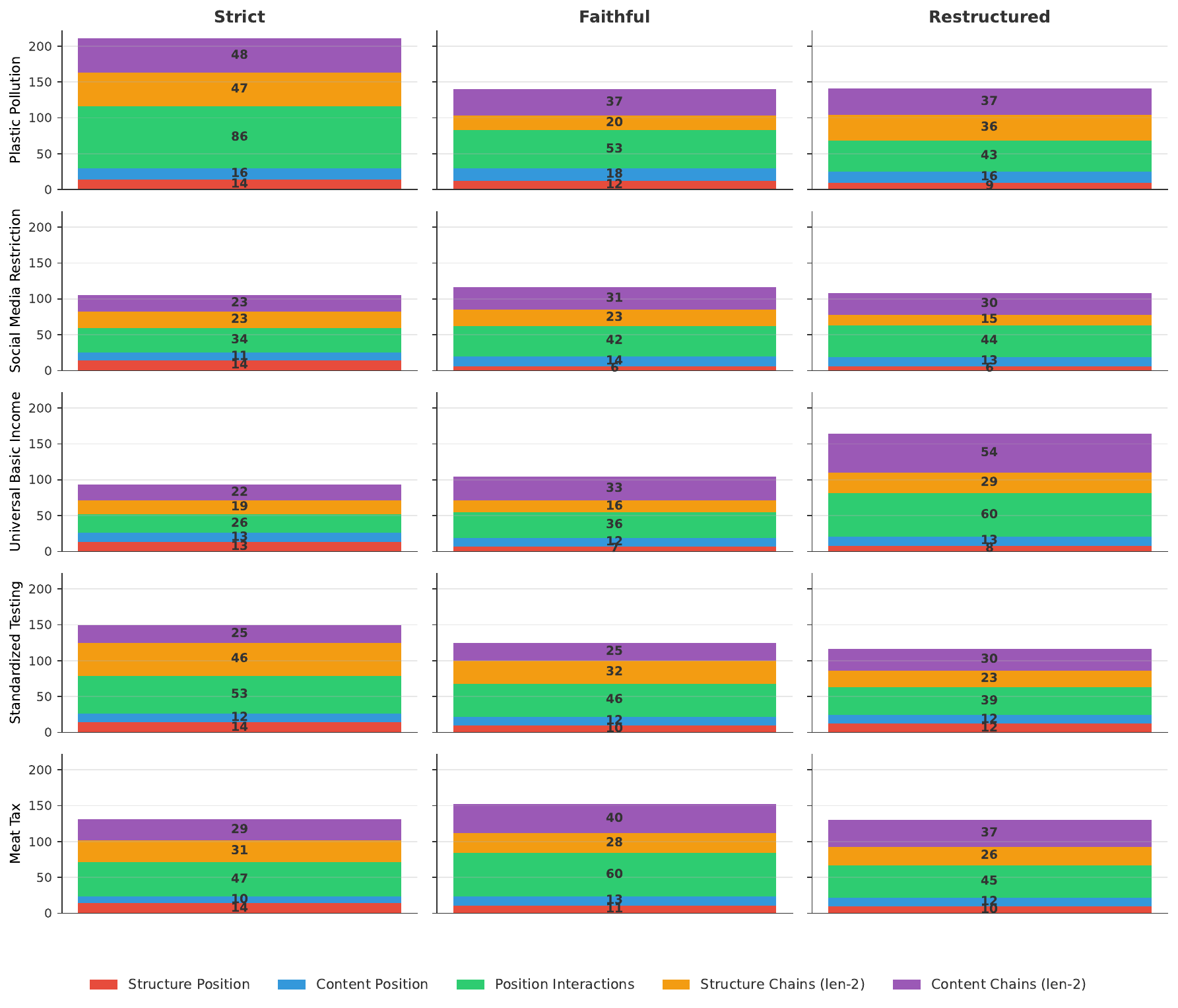}
  \caption{Distribution of LASSO-selected features by category across all five topics.
  Columns correspond to synthesis modes (strict, faithful, restructured); rows correspond to debate topics.
  Position effects and transitions within each dimension (structure, content) contribute to predictions across all configurations.}
  \label{fig:unified-m2-feature-categories}                     
\end{figure}

Cross-topic variation in absolute $R^2$ values is likely driven by differences in argument length distributions (see Figure~\ref{fig:unified-length-histograms}), judge consistency, and adherence to intended synthesis mode behavior (see controllability results discussed in Appendix~\ref{app:controllability_by_synthesis}).
Some length distributions are bimodal, which appears to increase predictability across all models under consideration.
Figures~\ref{fig:unified-m2-alpha-selection} and~\ref{fig:unified-m2-feature-categories} show the selected $\alpha$ values and the composition of retained features for M2 across topics and synthesis modes, illustrating how the LASSO regularization selects different feature subsets depending on the setting.

\FloatBarrier

\subsubsection{Targeted Trajectory Exploration}
\label{app:targeted-trajectory-details}

\begin{figure}[ht]
\centering
\includegraphics[width=0.9\textwidth]{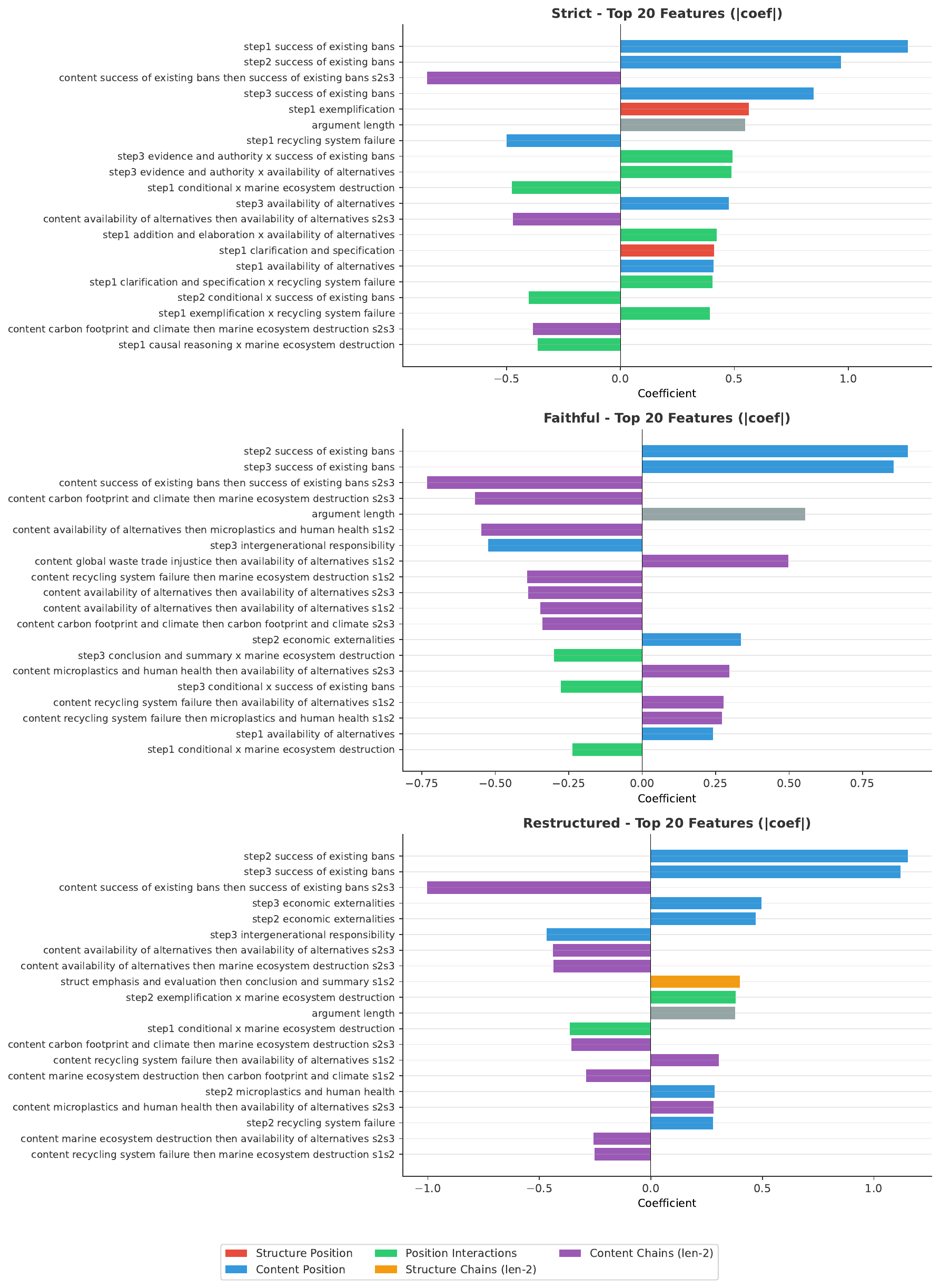}
\caption{Top 20 features by absolute LASSO coefficient for each synthesis type for M2 models for arguments on plastic pollution. Positive coefficients indicate patterns associated with higher persuasiveness scores; negative coefficients indicate patterns associated with lower persuasiveness scores.}
\label{fig:m2-top-features-arg-length}
\end{figure}

This subsection provides the full setup and results details for Section~\ref{sec:results-targeted-trajectory}.
Again, we expand the analysis with ablations covering the two additional synthesis modes (faithful and restructured) introduced in Appendix~\ref{app:synthesis-modes}.
We focus only on the plastic pollution topic in this experiment.
The attribution analysis establishes that M2's sequential features explain significant variance in argument quality.
A natural extension is testing whether these estimates generalize beyond the observed feature combinations.
M2 was trained on $\sim$5,000 arguments per synthesis type, but the full trajectory space contains $100^3 = 1{,}000{,}000$ possible 3-step sequences, the vast majority of which are unobserved.
If M2's learned coefficients generalize, we can use them to identify promising unexplored regions of the trajectory space and generate targeted arguments with those features.
This corresponds to the final step of our proposed workflow in Figure~\ref{fig:workflow}: using attribution estimates to guide targeted generation.
Using M2's fitted coefficients\footnote{
    Figure~\ref{fig:m2-top-features-arg-length} displays the top 20 features by coefficient magnitude for the M2 models across synthesis modes.
}, we score all possible trajectories and rank them by predicted quality.
We select the top 50 trajectories per synthesis type, all of which were never observed in the training data, and use \state's forced controller mechanism\footnote{
    Our implementation allows for forcing controller choices rather than using the controller module to choose the next action.
    For details, please refer to the repository.
} to generate arguments following each trajectory exactly.
For each trajectory, we generate 5 samples, yielding 250 targeted arguments per synthesis type.

We evaluate these targeted arguments against three baselines that test different aspects of M2's contribution.
First, the \textbf{Random} baseline samples trajectories uniformly from the unobserved trajectory space; if M2's selection provides no value, targeted arguments should perform at chance (50\%) against random exploration.
Second, the \textbf{M1b (Topic Presence)} baseline tests whether simply knowing which content topics correlate with quality is sufficient.
This emulates what a simpler topic modelling approach might discover: which topics matter, but without M2's sequential and structural information \citep{saenger-etal-2024-autopersuade, fong-grimmer-2016-discovery}.
Specifically, we identify the top-3 topics based on the M1b model and filter for trajectories that contain only these topics, then sample randomly from this filtered set.
Third, the \textbf{Original Top 5\%} baseline compares targeted arguments against the best 5\% of arguments from the original pairwise evaluation (by BT score), testing whether M2-guided generation can match or exceed the quality of the best observed arguments.

\begin{figure}[ht]
\centering
\includegraphics[width=\textwidth]{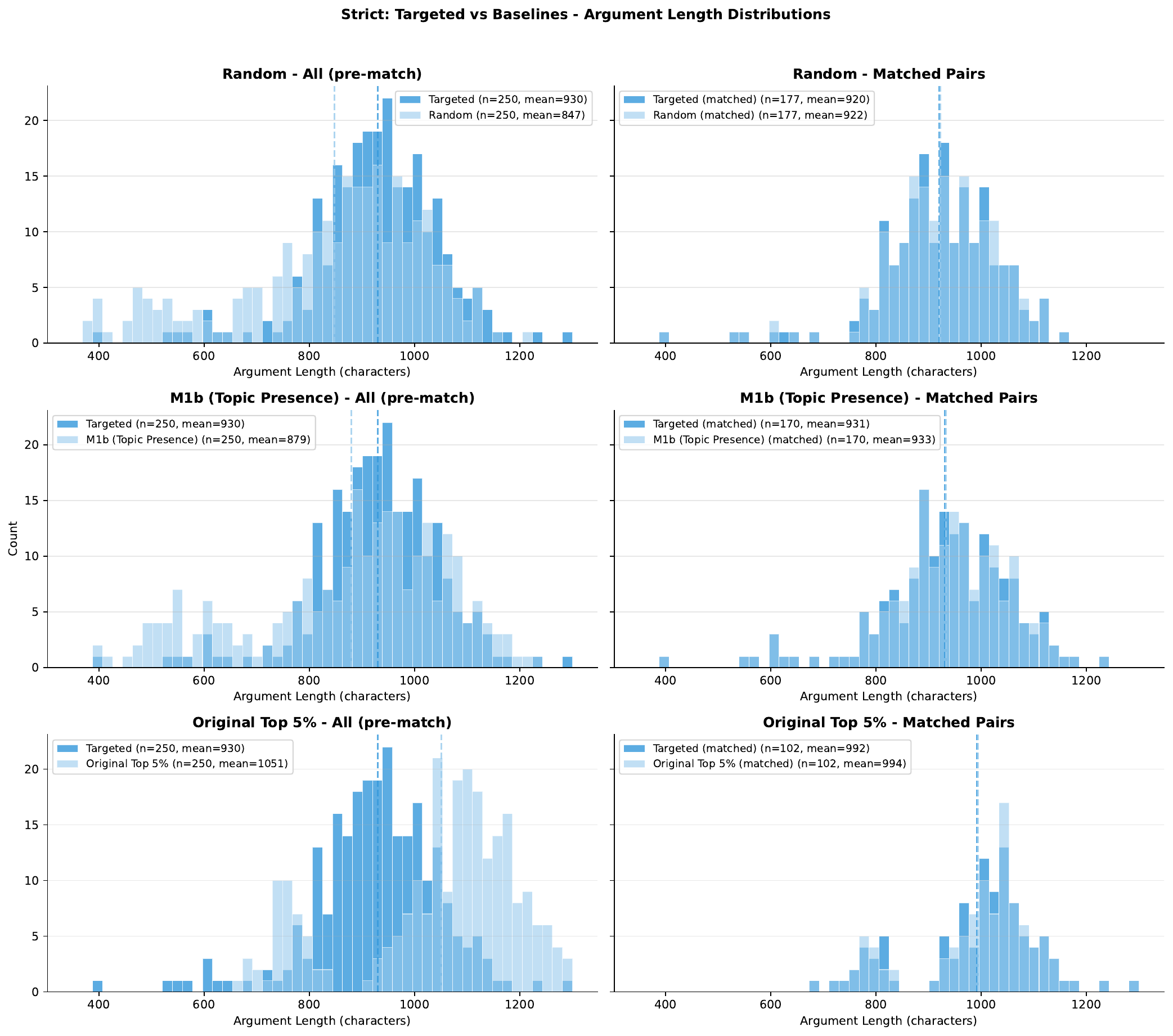}
\caption{Length distributions for targeted trajectory evaluation for \texttt{strict} synthesis. \textbf{Left:} All arguments before length matching, including 250 targeted arguments (generated from M2's top-50 predicted trajectories) and the 250 arguments of each baseline method. The original top-5\% arguments skew longer, reflecting the correlation between length and judged quality. \textbf{Right:} Length-matched subset used for evaluation. Greedy pairing within $\pm 5$ characters produces groups with comparable length distributions, enabling fair comparison. Dashed lines indicate group means.}
\label{fig:targted-arguments-length-histogram-strict}
\end{figure}

\begin{figure}[ht]
\centering
\includegraphics[width=\textwidth]{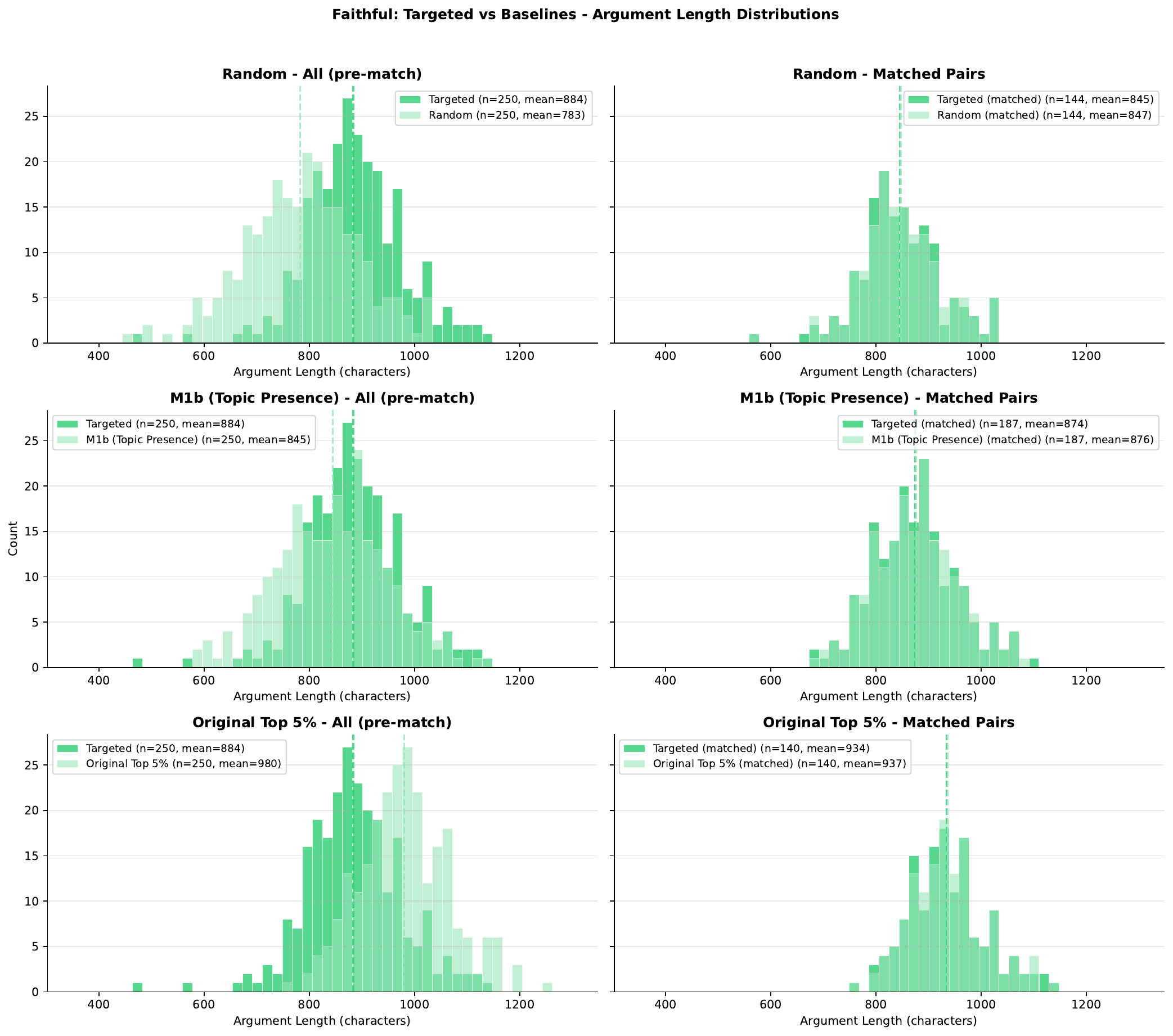}
\caption{Length distributions for targeted trajectory evaluation for \texttt{faithful} synthesis. \textbf{Left:} All arguments before length matching, including 250 targeted arguments (generated from M2's top-50 predicted trajectories) and the 250 arguments of each baseline method. The original top-5\% arguments skew longer, reflecting the correlation between length and judged quality. \textbf{Right:} Length-matched subset used for evaluation. Greedy pairing within $\pm 5$ characters produces groups with comparable length distributions, enabling fair comparison. Dashed lines indicate group means.}
\label{fig:targted-arguments-length-histogram-faithful}
\end{figure}

\begin{figure}[ht]
\centering
\includegraphics[width=\textwidth]{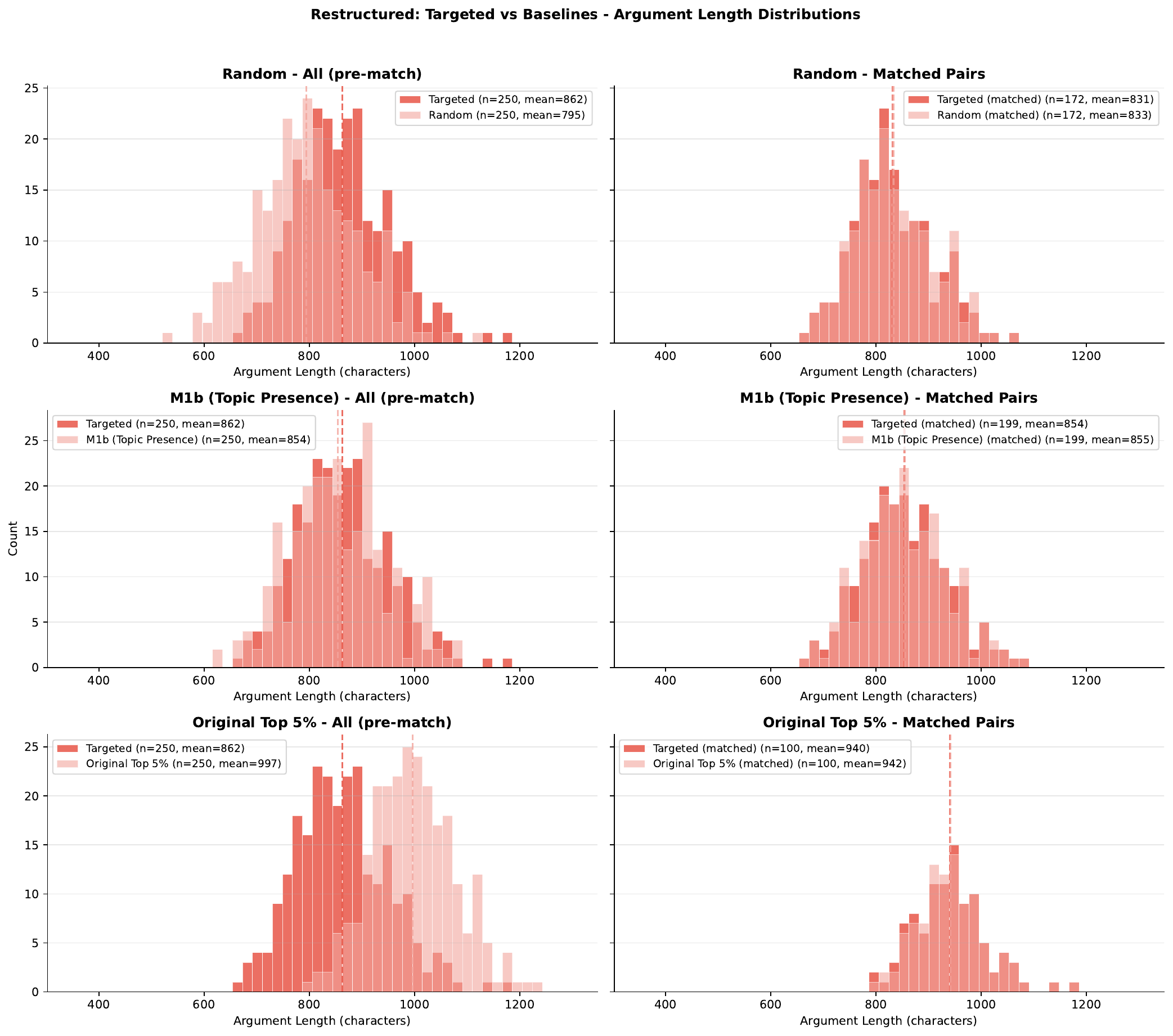}
\caption{Length distributions for targeted trajectory evaluation for \texttt{restructured} synthesis. \textbf{Left:} All arguments before length matching, including 250 targeted arguments (generated from M2's top-50 predicted trajectories) and the 250 arguments of each baseline method. The original top-5\% arguments skew longer, reflecting the correlation between length and judged quality. \textbf{Right:} Length-matched subset used for evaluation. Greedy pairing within $\pm 5$ characters produces groups with comparable length distributions, enabling fair comparison. Dashed lines indicate group means.}
\label{fig:targted-arguments-length-histogram-restructured}
\end{figure}

Argument length correlates strongly with performance in this setting, as shown in the predictability results of Appendix~\ref{app:predictability-by-synthesis}.
In particular, the original top 5\% arguments tend to be disproportionately long, creating a confounder that would affect any direct comparison \citep{dubois2024length}.
We construct length-matched evaluation sets using greedy pairing of arguments of similar length.
\footnote{
    For each targeted argument, we find the closest-length baseline argument within $\pm 5$ characters, using each baseline argument at most once.
    This effectively leaves us with the intersection of the length histograms shown in Figures~\ref{fig:targted-arguments-length-histogram-strict}, \ref{fig:targted-arguments-length-histogram-faithful}, and \ref{fig:targted-arguments-length-histogram-restructured}.
}
This yields balanced datasets ranging from 200 to 398 arguments, depending on synthesis type and baseline.
We evaluate these datasets by running 5,000 random pairwise comparisons within each and calculating new Bradley-Terry scores.
Win rates are computed over the cross-group comparisons (targeted vs.\ baseline).
Because each argument appears in roughly 15--25 comparisons, comparisons sharing an argument are correlated, and treating them as independent would understate uncertainty.
We therefore estimate 95\% CIs with a two-way cluster bootstrap.
In each of $B{=}10{,}000$ iterations, we resample the targeted and the baseline arguments with replacement and recompute the win rate, weighting each comparison by the product of its arguments' resample multiplicities.
Clustering at the argument level reflects variability across generated arguments rather than across repeated judgments of the same fixed arguments.

Table~\ref{tab:targeted-trajectory-results-all} shows that targeted arguments substantially outperform both the random baseline (73--81\% win rate) and the topic-presence baseline (61--67\% win rate) across synthesis types.
This confirms that M2's trajectory rankings identify genuinely promising regions of the action space, more so than a simpler topic-based approach might do.
Against the original top 5\%, targeted arguments remain competitive (31--68\% win rate), substantially exceeding the win rate of less than 5\% that we would expect if M2's rankings failed to generalize beyond the observed samples.
Especially in comparisons against the original top 5\%, the familiar predictability gradient emerges: strict synthesis shows the strongest performance, while restructured synthesis exhibits greater variability.
We also report the share of targeted arguments among the top-10 and top-100 of the performance-ranked, length-matched datasets.
This highlights that when the goal is to find the very best arguments, M2-guided trajectory selection offers a promising approach.

\begin{table}[htbp]
\centering
\small
\begin{tabular}{l l r r c r r}
\toprule
Baseline & Type & $N$ & Win (T) (\%) & 95\% CI & Top-10 & Top-100 \\
\midrule
Random & Strict & 354 & 78.7 & [73.9, 83.2] & 8/10 & 78/100 \\
M1b (Topic Presence) & Strict & 340 & 63.3 & [57.3, 69.2] & 6/10 & 57/100 \\
Original Top 5\% & Strict & 204 & 68.0 & [60.9, 74.8] & 9/10 & 68/100 \\
\midrule
Random & Faithful & 288 & 73.2 & [67.8, 78.6] & 7/10 & 75/100 \\
M1b (Topic Presence) & Faithful & 374 & 61.4 & [56.0, 66.8] & 5/10 & 61/100 \\
Original Top 5\% & Faithful & 280 & 35.7 & [30.2, 41.1] & 0/10 & 38/100 \\
\midrule
Random & Restructured & 344 & 81.0 & [76.6, 84.8] & 10/10 & 88/100 \\
M1b (Topic Presence) & Restructured & 398 & 67.2 & [62.2, 72.0] & 8/10 & 69/100 \\
Original Top 5\% & Restructured & 200 & 30.7 & [25.1, 36.6] & 1/10 & 34/100 \\
\bottomrule
\end{tabular}
\caption{
Targeted Trajectory Exploration: Evaluating new, targeted vs.\ new baseline explorations across synthesis modes. $N$ is the total number of length-matched argument pairs (balanced: $N/2$ targeted, $N/2$ baseline).
Comparisons count is the number of random pairwise comparisons evaluated. Win Rate is the share of comparisons between targeted and baseline arguments that the targeted argument won with 95\% bootstrap CIs ($B = 10{,}000$). Top-10 and Top-100 counts show the number of targeted arguments in the top-n arguments of the length-matched dataset when sorting by Bradley-Terry score based on the pairwise comparisons.}
\label{tab:targeted-trajectory-results-all}
\end{table}

\FloatBarrier

\subsection{Human Validation of the LLM Judge}
\label{app:human-study}

We conducted a human study to evaluate whether the LLM judge used for pairwise argument evaluation (Section~\ref{sec:experiments-argument-generation}) is a valid proxy for human preferences.
Our study is IRB-approved (approval number 2026-055) and participants in our Qualtrics survey were compensated with \$4 for completing our 12-question survey.
We used Prolific to distribute our survey to US residents who speak English fluently and are over 18 years old.
We only accepted participants with an approval rating of at least 99\% on Prolific.
On average, participants took 13.5 minutes to complete the survey.
We paid an additional \$1 performance bonus if participants passed both attention-check questions.

\subsubsection{Study Design}
\label{app:human-study-design}

The study focuses on arguments generated for a single topic (total ban on single-use plastics) and a single LLM judge (GPT-5-mini; \citealp{openai_gpt5_systemcard_2025}). Participants read two arguments supporting the motion and chose the more persuasive one in a forced binary choice (no ties).
We selected 50 argument pairs from the pool of 5,000 arguments previously scored via Bradley--Terry analysis of 50,000 LLM-judged pairwise comparisons, stratified by BT score gap: small ($< 1.0$; 17 pairs), medium ($1.0$--$2.0$; 17 pairs), and large ($\geq 2.0$; 16 pairs).
The LLM judge evaluated the same pairs via pairwise comparison with effectively the same instruction format used in the human study.

We recruited 310 Prolific workers (US-based, English fluency, high approval rating), each evaluating 10 pairs.
Participants were assigned to one of 10 templates, each containing a fixed set of 10 pairs.
Every pair appeared in exactly 2 templates (order of the arguments switched).
Six participants submitted duplicate responses (first submission kept).
Two attention-check items were embedded in each survey, pairing a top-ranked argument against a deliberately degraded counterpart for which the stronger argument is unambiguous; 94.2\% of participants (292/310) passed both checks.
After filtering to unique participants who passed both attention checks, the final sample contained $n{=}288$ participants.
% After filtering to unique participants who passed both attention checks: $n = 288$.

\subsubsection{The ALT Test}
\label{sec:alt-test}

We used the ALT test \citep{calderon2025alternativeannotatortestllmasajudge} to determine whether the LLM judge is interchangeable with a human annotator.
The test uses a leave-one-out procedure where for each human annotator $h$, it excludes $h$ from the pool, scores both $h$ and the LLM against the remaining annotators on each instance, and checks whether the LLM matches or exceeds $h$'s alignment with the group.
Formally, for each instance $i$ annotated by $h$, the alignment score is the fraction of remaining annotators who gave the same answer.

The ALT test's primary statistic, the \emph{winning rate}, applies per-annotator hypothesis tests with FDR correction \citep{benjamini2001control}.
However, our study design (288 annotators, each rating only 10 pairs) leaves these individual tests severely underpowered, making the winning rate uninformative in our setting.
We therefore focus on the \emph{advantage probability}, a non-parametric summary that does not depend on per-annotator statistical power: across all annotators and all instances, how often does the LLM match or exceed the excluded human's alignment with the group?

\subsubsection{Results}
\label{app:human-study-results}

Table~\ref{tab:human-agreement} reports agreement between human annotators and between humans and the LLM judge.
Human pairwise agreement of 56.0\% reflects the inherent subjectivity of persuasiveness judgments; even among humans, annotators disagree nearly half the time.
Krippendorff's $\alpha = 0.118$ confirms low absolute agreement.
Despite this low inter-annotator reliability, the LLM agrees with the human majority vote 72.0\% of the time, substantially above the human pairwise baseline.

\begin{table}[ht]
\centering
\begin{tabular}{lcccc}
\toprule
Metric & Overall & Small & Medium & Large \\
\midrule
LLM vs.\ human majority    & 0.720 & 0.706 & 0.706 & 0.750 \\
Human pairwise agreement    & 0.560 & 0.584 & 0.515 & 0.583 \\
Human majority proportion   & 0.652 & 0.672 & 0.593 & 0.693 \\
Krippendorff's $\alpha$     & 0.118 & 0.164 & 0.031 & 0.160 \\
\bottomrule
\end{tabular}
\caption{Agreement statistics for the human validation study ($n = 288$, passed 2/2 attention checks).}
\label{tab:human-agreement}
\end{table}

The advantage probability of the ALT test reaches 0.841. Concretely, when a single human is removed from the pool and compared against the LLM on the same instances, the LLM agrees with the remaining annotators at least as often as the excluded human does in 84.1\% of comparisons. Table~\ref{tab:advantage-prob-stratified} breaks this metric down by the Bradley-Terry score gap between the two arguments in each pair, as determined by the LLM judge's own prior pairwise evaluations (Section~\ref{app:human-study-design}).
The advantage probability remains high across all three strata (0.82--0.86), suggesting that the LLM is equally reliable on pairs it previously rated as close calls and on pairs it previously rated as clearly different.
The slight decrease from small-gap to large-gap pairs does not indicate weaker LLM performance on easier pairs; rather, humans are more internally consistent on large-gap pairs (fewer outliers), which raises the baseline the LLM is compared against.

\begin{table}[ht]
\centering
\begin{tabular}{lcc}
\toprule
Stratum & Advantage Prob. & Annotators Tested \\
\midrule
Overall & 0.841 & 288 \\
\midrule
Small  & 0.860 & 229 \\
Medium & 0.834 & 200 \\
Large  & 0.823 & 202 \\
\bottomrule
\end{tabular}
\caption{Advantage probability by BT score gap stratum ($n = 288$).}
\label{tab:advantage-prob-stratified}
\end{table}

The LLM's choices are also significantly correlated with human preference rates (Pearson $r = 0.493$, $p < .001$); when more humans prefer argument A, the LLM is more likely to also choose A.
This correlation strengthens for large-gap pairs ($r = 0.657$, $p = .006$), where the quality difference is clearest.
The LLM is also more accurate on pairs where humans agree more strongly ($r = 0.300$, $p = .035$), though this effect is driven primarily by large-gap pairs ($r = 0.609$, $p = .012$).

\FloatBarrier

\section{Limitations}
\label{app:limitations}

\state relies on \emph{prefilling} text prefixes to implement intervention-based branching.
Modern closed-source APIs (e.g., GPT, Claude, Gemini) generally do not provide robust support for the kind of assistant-prefill control required by this workflow, so the method is currently most straightforward to deploy with open-source/self-hosted models.
In addition, our analysis of action--outcome relationships is associative: while we find that action sequences are predictive of downstream judgments and that targeted, previously unseen trajectories can perform well, we do not make causal claims about how any particular action choice affects the final text or the downstream outcome.
Our current setup violates sequential ignorability, as we choose actions conditional on existing reasoning.

A second limitation is rigidity in the action and prefix design.
In our implementation, each action is realized by a fixed textual prefix, but many interventions admit multiple natural surface forms (e.g., synonymous discourse markers for ``causal reasoning'' include ``Because'', ``Therefore'', and ``As a result''). Representing these variants naively would require expanding the action space substantially, while many actions would share identical definitions.
Moreover, some prefixes are well-formed as mid-document transitions but can sound unnatural as the first step of an answer (e.g., ``Therefore'' or ``However''), which can create stylistic artifacts unless one conditions the action space on position or introduces context-aware prefix variants.
These issues point to a broader limitation: action spaces require careful, task-specific engineering, and the best granularity of actions (coarse vs. fine) may vary across domains.

Relatedly, the synthesis step that converts reasoning traces into final outputs introduces a trade-off between control and quality.
Strict synthesis preserves a tight coupling between action sequences and output text, enabling high predictability in the argument quality experiment, but potentially producing stilted prose that mechanically concatenates reasoning steps.
More flexible synthesis modes allow the model to smooth transitions and improve eloquence, but this freedom attenuates the mapping from actions to output properties.
This trade-off has practical implications: when the goal is to study how specific rhetorical choices affect perceived quality, stricter synthesis provides cleaner attribution, whereas producing high-quality arguments for deployment may favor more flexible synthesis despite reduced interpretability.
In practice, our synthesis modes do not always behave as intended.
Under strict synthesis, outputs are typically simple concatenations of reasoning steps, but occasionally the synthesis step adds concluding sentences, producing unexpectedly longer arguments.
Results in Appendix~\ref{app:controllability_by_synthesis} and~\ref{app:predictability-by-synthesis} show that the expected gradient in control and predictability across the three
evaluated synthesis modes did not materialize, warranting further investigation into synthesis prompt design and more explicit length control.

Additionally, our current approach to generating multiple realizations of the same trajectory relies on random seeds and achieves only limited diversity.
Future work could introduce variation at the tree level, such as including personas \citep{park_etal_2023_simulacra, park2024generativeagentsimulations1000} in the input or using different language models, to enable better estimation of trajectory-level effects.

Finally, our framework makes several scope assumptions.
\state currently focuses on single-turn, multi-step generation and does not explicitly model multi-turn conversational dynamics.
\footnote{
    We do not support tool calls and tool call outputs as conversation turns, instead opting to directly include them as part of the thinking process of assistant messages.
}
We also treat actions as textual interventions and do not support general tool calling beyond what is encoded in the action templates; integrating retrieval (for additional diversity) and tool-based verification (e.g., for numerical or algorithmic claims) could improve both generation and evaluation.

\section{Extended Future Work}
\label{app:future-work-extended}

\subsection{Causal inference for sequential action choices}
\label{app:future-work-causal}
Our current attribution analysis is focused on associations and predictability rather than causal claims.
% A natural extension is to formalize action sequences as sequential treatments, where at each depth $i\in {1, \ldots, d}$ the controller selects $a_i$ conditional on previous actions ($a_1, \ldots, a_{i-1}$) and current state ($s_i$).
A natural extension is to formalize action sequences as sequential treatments, where, at each depth $i \in \{1,\ldots,d\}$, the controller selects $a_i$ conditional on previous actions $(a_1,\ldots,a_{i-1})$ and the current parent state $s_{i-1}$.
This framework is directly related to marginal structural models and sequential ignorability \citep{hernan2000marginal, Hernan2024-WhatIf}.
In this framework, g-computation or inverse probability weighting could estimate the per-step causal effects of $a_i$ on the quality of the final output \citep{Hernan2024-WhatIf}.
Mediation analysis \citep{vanderweele2015explanation} could further distinguish whether an action affects the outcome directly or through its influence on subsequent action choices.
Crucially, because \state's controller can randomize action selections at each step, it enables experimental designs that eliminate the need for sequential ignorability assumptions.

\subsection{Human evaluation and behavioral outcomes}
\label{app:future-work-human}
Our argument evaluation currently relies on LLM judges (Section~\ref{sec:experiments-argument-generation}).
Although this offers repeated access to stable preferences and is broadly correlated with human judgments, it does not substitute for rigorous human experimentation.
The preferences of groups and individuals are complex, context-dependent, and shaped by heterogeneous prior beliefs that are difficult to simulate with language models.
Future work should therefore conduct controlled human subject experiments with pre- and post-intervention measurements of beliefs or behaviors.
\state's sequential and multi-dimensional action traces provide a uniquely informative design space for such studies, enabling systematic manipulation of rhetorical structure, topical framing, and ordering effects that would be difficult to isolate using other methods. 
Furthermore, such experiments would allow us to report notions of persuasiveness (the effect that an argument has on a reader) rather than argument quality (i.e., human preference towards argumentative text). 

\subsection{Search and optimization over action spaces}
\label{app:future-work-search}
We currently explore action spaces using fixed beam search and show that regression-based estimates can identify promising, previously unobserved trajectories (Section~\ref{sec:results-targeted-trajectory}).
A natural extension is to employ principled tree search algorithms such as Monte Carlo Tree Search (MCTS) \citep{kocsis_and_szepesvari_2006_mcts, coulom_2006_mcts, browne_etal_2012_mcts, silver2016mastering, silver_etal_2018_alphazero, hao-etal-2023-reasoning}.
Such approaches could iteratively generate arguments, update effect estimates, and adapt exploration toward high-performing regions of the action space under constrained evaluation budgets.
Importantly, tree search methods also extend naturally to multi-turn and adversarial settings.
This opens the possibility of integrating \state with Multi-Agent Debate (MAD) frameworks to identify optimal multi-turn conversational or adversarial strategies, rather than optimizing single-turn outputs alone.

\subsection{Weight-based optimization with Reinforcement Learning}
\label{app:future-work-rl}

Group-wise policy optimization methods for LLMs \citep{shao2024deepseekmathpushinglimitsmathematical, liu2025understanding} often suffer from mode collapse, where multiple sampled trajectories converge to near-identical completions.
Prior work highlights this as a central limitation of RLHF-style training regimes \citep{casper2023open_problems_rlhf, gai2025differentialsmoothingmitigatessharpening}.
By sampling across discrete and interpretable action sequences rather than relying solely on token-level stochasticity, \state increases semantic diversity while preserving quality (Section~\ref{sec:experiments-noveltybench}), potentially mitigating collapse in group-based rollout sets.
Weight-based optimization via PPO \citep{schulman2017ppo}, GRPO \citep{shao2024deepseekmathpushinglimitsmathematical}, or related methods could train the controller, generator, or evaluator to improve downstream performance.

\subsection{Prompt optimization}
\label{app:future-work-prompt-opt}

Given a reliable ORM signal from a real downstream task (e.g., human preferences, click-through rates, or task success), \state's components can be optimized in a cascading fashion.
The PRM can be calibrated to assign high scores to intermediate states that yielded strong final outputs and low scores to states that yielded weak outputs.
The controller can be optimized to select actions that maximize the expected ORM score, and the generator can be optimized to maximize PRM or ORM scores subject to the chosen action.
Prompt optimization methods \citep{khattab2023dspy, opsahl-ong-etal-2024-optimizing, agrawal2025gepareflectivepromptevolution, yuksekgonul2025optimizing} offer one path: they can yield better or more precise task instructions, in-context examples that exemplify desired behaviors, and broadly improve the reliability of each component.

Separately, \state's structured action spaces also offer a mechanism to diversify prompt-search strategies within reflective prompt evolution frameworks \citep{agrawal2025gepareflectivepromptevolution}.
That is, with an action space that reflects sequential prompt edits (e.g., including a new in-context example, or adding an edge case to the instructions), \state can be used to search for the best prompt configuration to maximize a provided metric.

\section{Action Spaces}
\label{app:action-spaces}

This appendix lists the structured action templates used by \state in each experiment suite.

\subsection{NoveltyBench action spaces}
\label{app:action-spaces-noveltybench}

\begingroup
\footnotesize
\sloppy
\renewcommand{\arraystretch}{1.15}
\setlength{\tabcolsep}{2pt}
\begin{longtable}{@{}L{0.20\textwidth} L{0.31\textwidth} L{0.13\textwidth} L{0.26\textwidth}@{}}
\toprule
\textbf{Name} & \textbf{Definition} & \textbf{Prefix} & \textbf{Internal reasoning} \\
\midrule
\endfirsthead
\toprule
\textbf{Name} & \textbf{Definition} & \textbf{Prefix} & \textbf{Internal reasoning} \\
\midrule
\endhead
\path{openness} &
Emphasizes creativity, intellectual curiosity, and preference for novelty over tradition. &
\textit{(none)} &
Approach with curiosity and creativity; seek novel ideas; think abstractly and imaginatively. \\
\midrule
\path{conscientiousness} &
Emphasizes organization, self-discipline, reliability, and goal-oriented achievement. &
\textit{(none)} &
Be methodical and detail-oriented; work systematically toward clear goals; be thorough. \\
\midrule
\path{extraversion} &
Emphasizes enthusiasm, assertiveness, sociability, and high energy. &
\textit{(none)} &
Be energetic and confident; engage boldly; express thoughts with passion and optimism. \\
\midrule
\path{agreeableness} &
Emphasizes empathy, cooperation, warmth, and concern for others. &
\textit{(none)} &
Be empathetic and collaborative; consider stakeholders; seek harmony; show sincere concern. \\
\midrule
\path{neuroticism} &
Emphasizes caution, risk awareness, and sensitivity to potential problems. &
\textit{(none)} &
Be cautious and attentive to risks; examine uncertainties; consider worst-case scenarios. \\
\bottomrule
\caption{NoveltyBench action space: Personality (Big-5 traits).} \\
\end{longtable}
\endgroup
\begingroup
\footnotesize
\sloppy
\renewcommand{\arraystretch}{1.15}
\setlength{\tabcolsep}{2pt}
\begin{longtable}{@{}L{0.20\textwidth} L{0.31\textwidth} L{0.13\textwidth} L{0.26\textwidth}@{}}
\toprule
\textbf{Name} & \textbf{Definition} & \textbf{Prefix} & \textbf{Internal reasoning} \\
\midrule
\endfirsthead
\toprule
\textbf{Name} & \textbf{Definition} & \textbf{Prefix} & \textbf{Internal reasoning} \\
\midrule
\endhead
\path{children} &
Writes for children ages 5--12 using simple language, examples, and enthusiasm. &
\textit{(none)} &
Use very simple words and short sentences; cheerful tone; fun, concrete examples. \\
\midrule
\path{teenagers} &
Writes for teenagers ages 13--19 using relatable language, current trends, and engaging tone. &
\textit{(none)} &
Use casual, relatable language; energetic tone; socially current examples. \\
\midrule
\path{young_adults} &
Writes for young adults ages 20--35 using modern, direct language with practical examples. &
\textit{(none)} &
Use clear, modern language; practical examples; confident, approachable tone. \\
\midrule
\path{middle_aged} &
Writes for adults ages 36--55 using professional, balanced tone with real-world applications. &
\textit{(none)} &
Use professional, balanced tone; grounded examples; pragmatic framing. \\
\midrule
\path{seniors} &
Writes for seniors (ages 56+) using clear, respectful, and warm language. &
\textit{(none)} &
Use clear, respectful language; gentle pacing; thoughtfully explained examples. \\
\bottomrule
\caption{NoveltyBench action space: Target Audience (age demographics).} \\
\end{longtable}
\endgroup

\newpage

\subsection{Argument generation action spaces}
\label{app:action-spaces-argument-generation}

\begingroup
\footnotesize
\sloppy
\renewcommand{\arraystretch}{1.15}
\setlength{\tabcolsep}{2pt}
\begin{longtable}{@{}L{0.18\textwidth} L{0.29\textwidth} L{0.08\textwidth} L{0.45\textwidth}@{}}
\label{tab:action-space-content-plastic} \\
\toprule
\textbf{Name} & \textbf{Definition} & \textbf{Prefix} & \textbf{Internal reasoning} \\
\midrule
\endfirsthead
\toprule
\textbf{Name} & \textbf{Definition} & \textbf{Prefix} & \textbf{Internal reasoning} \\
\midrule
\endhead
\path{marine_ecosystem_destruction} &
Examines plastic pollution's impact on ocean life, food chains, and marine habitats. &
\textit{(none)} &
I should consider how single-use plastics kill marine animals through ingestion and entanglement, degrade coral reefs, and disrupt ocean ecosystems at massive scale.\\
\midrule
\path{microplastics_and_human_health} &
Analyzes the infiltration of plastic particles into human bodies and potential health consequences. &
\textit{(none)} &
I should examine emerging evidence that microplastics accumulate in human tissue, blood, and organs, posing uncertain but potentially serious health risks.\\
\midrule
\path{recycling_system_failure} &
Evaluates why recycling has proven inadequate as a solution to plastic waste. &
\textit{(none)} &
I should consider that global plastic recycling rates remain below 10\%, much plastic is downcycled or exported, and source reduction through bans is therefore necessary.\\
\midrule
\path{carbon_footprint_and_climate} &
Connects plastic production to fossil fuel extraction and greenhouse gas emissions. &
\textit{(none)} &
I should analyze how single-use plastics are petrochemical products that contribute to emissions throughout their lifecycle, from extraction to incineration.\\
\midrule
\path{availability_of_alternatives} &
Assesses whether practical substitutes exist for common single-use plastic applications. &
\textit{(none)} &
I should demonstrate that reusable, compostable, and biodegradable alternatives are available for bags, straws, containers, and packaging, making a ban practically feasible.\\
\midrule
\path{economic_externalities} &
Quantifies societal costs of plastic pollution not reflected in product prices. &
\textit{(none)} &
I should calculate how cleanup costs, healthcare burdens, tourism losses, and ecosystem damage are borne by the public rather than producers, a market failure requiring correction.\\
\midrule
\path{success_of_existing_bans} &
Draws lessons from jurisdictions that have already implemented plastic restrictions. &
\textit{(none)} &
I should examine case studies from Rwanda, the EU, Kenya, and various US states showing that bans are enforceable and produce measurable reductions in pollution.\\
\midrule
\path{global_waste_trade_injustice} &
Addresses how plastic waste burdens fall disproportionately on developing nations. &
\textit{(none)} &
I should consider how wealthy countries export plastic waste to poorer nations, making this a global justice issue that requires upstream intervention at the production stage.\\
\midrule
\path{corporate_behavior_and_voluntary_failure} &
Examines why industry self-regulation has proven insufficient to reduce plastic use. &
\textit{(none)} &
I should analyze how manufacturers default to cheap plastics despite pledges, and why only binding regulation, not voluntary commitments, can shift industry norms.\\
\midrule
\path{inter} \path{generational_responsibility} &
Considers obligations to future generations given plastic's persistence. &
\textit{(none)} &
I should evaluate how plastics persist for centuries in the environment, meaning today's convenience imposes long-term burdens on future generations who had no say in their creation.\\
\bottomrule
\caption{Subtopics for \textbf{Plastic Pollution} (domain-specific topical lenses for the single-use plastic ban proposition).}
\label{tab:action-space-content-plastic-waste}
\end{longtable}
\endgroup

\begingroup
\footnotesize
\sloppy
\renewcommand{\arraystretch}{1.15}
\setlength{\tabcolsep}{2pt}
\begin{longtable}{@{}L{0.18\textwidth} L{0.29\textwidth} L{0.08\textwidth} L{0.45\textwidth}@{}}
\label{tab:action-space-content-social-media} \\
\toprule
\textbf{Name} & \textbf{Definition} & \textbf{Prefix} & \textbf{Internal reasoning} \\
\midrule
\endfirsthead
\toprule
\textbf{Name} & \textbf{Definition} & \textbf{Prefix} & \textbf{Internal reasoning} \\
\midrule
\endhead
\path{adolescent_brain_development} &
Examines how prefrontal cortex immaturity makes under-16s vulnerable to addictive design and impulsive online behavior. &
\textit{(none)} &
I should consider how the prefrontal cortex, responsible for impulse control and risk assessment, does not fully mature until the mid-20s, making adolescents particularly susceptible to dopamine-driven engagement loops and poor decision-making online.\\
\midrule
\path{mental_health_impact} &
Analyzes the relationship between social media use and anxiety, depression, and body image issues in young people. &
\textit{(none)} &
I should examine how longitudinal studies link heavy social media use in under-16s to increased rates of anxiety, depression, loneliness, and body dissatisfaction driven by social comparison and curated self-presentation.\\
\midrule
\path{online_predation_and_safety} &
Evaluates the risks of grooming, exploitation, and data privacy violations targeting minors on social platforms. &
\textit{(none)} &
I should analyze how social media platforms expose minors to predatory adults through direct messaging, how children's data is harvested for targeted advertising, and how age restrictions reduce the attack surface for exploitation.\\
\midrule
\path{cyberbullying_and_harassment} &
Examines the prevalence and psychological harm of online bullying among younger users and its spillover into schools. &
\textit{(none)} &
I should consider how cyberbullying disproportionately affects younger adolescents, extends school-based conflicts into 24/7 online environments, and correlates with self-harm and suicidal ideation in vulnerable youth.\\
\midrule
\path{attention_and_academic_performance} &
Analyzes the effects of social media use on concentration, learning outcomes, and sleep quality in young people. &
\textit{(none)} &
I should examine how constant notifications and context-switching fragment attention spans, how screen time before bed disrupts sleep architecture critical for learning, and how academic performance declines correlate with social media usage hours.\\
\midrule
\path{addictive_design_exploitation} &
Evaluates how platform features like infinite scroll, notifications, and variable rewards deliberately exploit developing minds. &
\textit{(none)} &
I should analyze how social media platforms employ behavioral psychology techniques, including variable ratio reinforcement schedules and social validation feedback loops, that are specifically designed to maximize engagement and are disproportionately effective on developing brains.\\
\midrule
\path{corporate_accountability_failure} &
Examines platforms' consistent failure to self-regulate and their prioritization of profit over child safety. &
\textit{(none)} &
I should consider how internal documents from major platforms reveal awareness of harm to young users alongside decisions to prioritize engagement metrics, demonstrating that voluntary self-regulation has consistently failed to protect children.\\
\midrule
\path{age_verification_feasibility} &
Evaluates the technical viability and proportionate privacy tradeoffs of implementing effective age restrictions. &
\textit{(none)} &
I should examine how existing age verification technologies, including document verification and age estimation, can achieve high accuracy while maintaining proportionate privacy protections, as demonstrated by successful implementations in Australia and the EU.\\
\midrule
\path{social_development_and_identity} &
Analyzes how premature social media exposure distorts self-concept and undermines healthy offline socialization. &
\textit{(none)} &
I should consider how early exposure to curated online personas, adult content, and performative social dynamics interferes with the natural development of identity, empathy, and face-to-face social skills during critical developmental periods.\\
\midrule
\path{international_regulatory_precedents} &
Draws lessons from countries that have implemented or proposed social media age restrictions and their outcomes. &
\textit{(none)} &
I should analyze how Australia's social media ban for under-16s, the EU Digital Services Act's enhanced protections for minors, and China's youth usage time limits provide evidence on implementation feasibility, enforcement mechanisms, and positive outcomes.\\
\bottomrule
\caption{Subtopics for \textbf{Social Media Age Restriction} (domain-specific topical lenses for enforcing a minimum age restriction of 16 on social media platforms).}
\end{longtable}
\endgroup

\begingroup
\footnotesize
\sloppy
\renewcommand{\arraystretch}{1.15}
\setlength{\tabcolsep}{2pt}
\begin{longtable}{@{}L{0.18\textwidth} L{0.29\textwidth} L{0.08\textwidth} L{0.45\textwidth}@{}}
\label{tab:action-space-content-ubi} \\
\toprule
\textbf{Name} & \textbf{Definition} & \textbf{Prefix} & \textbf{Internal reasoning} \\
\midrule
\endfirsthead
\toprule
\textbf{Name} & \textbf{Definition} & \textbf{Prefix} & \textbf{Internal reasoning} \\
\midrule
\endhead
\path{poverty_elimination_effectiveness} &
Examines direct cash transfers as the most efficient mechanism for poverty reduction across developed and developing nations. &
\textit{(none)} &
I should consider how unconditional cash transfers have consistently outperformed in-kind assistance programs, with evidence from multiple countries showing that recipients invest in education, health, and productive assets rather than wasteful consumption.\\
\midrule
\path{bureaucratic_waste_elimination} &
Analyzes how replacing complex means-tested welfare with universal payments reduces administrative overhead and errors. &
\textit{(none)} &
I should examine how current welfare systems spend significant portions of their budgets on eligibility determination, compliance monitoring, and fraud prevention, while means-testing creates poverty traps and excludes many eligible recipients through administrative burden.\\
\midrule
\path{fiscal_sustainability} &
Evaluates viable funding mechanisms including VAT, wealth taxes, carbon dividends, and sovereign wealth funds. &
\textit{(none)} &
I should analyze how UBI can be funded through combinations of value-added taxes, progressive wealth taxes, carbon pricing revenue, and reallocation of existing welfare spending, while long-term savings from reduced poverty-related costs improve fiscal sustainability.\\
\midrule
\path{entrepreneur} \path{ship_and_innovation} &
Examines how guaranteed income security enables risk-taking, startup creation, and creative pursuits. &
\textit{(none)} &
I should consider how economic insecurity prevents talented individuals from pursuing entrepreneurship, and how a guaranteed income floor would enable more people to start businesses, invest in skills development, and engage in socially valuable creative work.\\
\midrule
\path{mental_health_and_wellbeing} &
Analyzes how guaranteed income reduces financial stress and eliminates the stigma of means-tested assistance. &
\textit{(none)} &
I should examine how financial insecurity is a primary driver of chronic stress, anxiety, and depression, and how UBI's universality eliminates the shame and administrative burden associated with applying for targeted welfare programs.\\
\midrule
\path{automation_and_technological_unemployment} &
Evaluates UBI as a necessary safety net for workforce displacement driven by AI and robotics. &
\textit{(none)} &
I should analyze how accelerating automation in manufacturing, transportation, retail, and professional services threatens to displace millions of workers, making a universal income floor essential for maintaining social stability during the economic transition.\\
\midrule
\path{labor_market_empowerment} &
Examines how UBI strengthens workers' bargaining power and ability to refuse exploitative employment. &
\textit{(none)} &
I should consider how a guaranteed income floor gives workers genuine freedom to negotiate better conditions, leave abusive employers, pursue education or retraining, and refuse work that is unsafe, underpaid, or degrading.\\
\midrule
\path{gender_equity_effects} &
Analyzes how UBI values unpaid care work and reduces gendered economic dependency. &
\textit{(none)} &
I should examine how women disproportionately perform unpaid caregiving and domestic labor that GDP fails to capture, and how UBI provides economic recognition and independence for caregivers while reducing the gendered poverty gap.\\
\midrule
\path{pilot_program_evidence} &
Draws on positive results from UBI experiments in Finland, Kenya, Stockton, and Alaska to demonstrate real-world effectiveness. &
\textit{(none)} &
I should analyze how Finland's basic income experiment improved wellbeing and employment confidence, GiveDirectly's Kenya program boosted economic activity, Stockton's SEED program increased full-time employment, and Alaska's Permanent Fund has distributed dividends for decades without adverse effects.\\
\midrule
\path{economic_stimulus_effects} &
Evaluates how universal cash transfers boost consumer spending, local economies, and small business growth. &
\textit{(none)} &
I should consider how direct cash transfers to all citizens increase aggregate demand, particularly among lower-income households with high marginal propensity to consume, creating local economic multiplier effects that benefit small businesses and community economies.\\
\bottomrule
\caption{Subtopics for \textbf{Universal Basic Income} (domain-specific topical lenses for implementing a government UBI program).}
\end{longtable}
\endgroup

\begingroup
\footnotesize
\sloppy
\renewcommand{\arraystretch}{1.15}
\setlength{\tabcolsep}{2pt}
\begin{longtable}{@{}L{0.18\textwidth} L{0.29\textwidth} L{0.08\textwidth} L{0.45\textwidth}@{}}
\label{tab:action-space-content-testing} \\
\toprule
\textbf{Name} & \textbf{Definition} & \textbf{Prefix} & \textbf{Internal reasoning} \\
\midrule
\endfirsthead
\toprule
\textbf{Name} & \textbf{Definition} & \textbf{Prefix} & \textbf{Internal reasoning} \\
\midrule
\endhead
\path{measurement_validity} &
Examines whether standardized tests actually measure the knowledge and skills they claim to assess. &
\textit{(none)} &
I should consider how standardized tests often measure test-taking ability rather than deep understanding, failing to capture critical thinking, creativity, and practical knowledge.\\
\midrule
\path{equity_and_access_gaps} &
Analyzes how testing disparities reflect and reinforce socioeconomic, racial, and linguistic inequalities. &
\textit{(none)} &
I should examine how score gaps correlate with family income, access to test prep, and neighborhood resources rather than innate ability, perpetuating systemic inequality.\\
\midrule
\path{curriculum_narrowing} &
Evaluates how test-centric accountability causes schools to abandon rich curriculum for test prep. &
\textit{(none)} &
I should consider how high-stakes testing incentivizes schools to cut arts, science labs, and social studies in favor of drilling tested subjects, impoverishing the educational experience.\\
\midrule
\path{student_mental_health} &
Addresses the psychological toll of high-stakes testing on student wellbeing and motivation. &
\textit{(none)} &
I should analyze how test anxiety, performance pressure, and fear of failure harm student mental health and can undermine intrinsic motivation to learn.\\
\midrule
\path{teacher_professional_impact} &
Examines how test-based evaluation constrains teacher autonomy and degrades the profession. &
\textit{(none)} &
I should consider how tying teacher evaluations to test scores forces teaching to the test, drives talented educators from the profession, and reduces pedagogical innovation.\\
\midrule
\path{alternative_assessment_methods} &
Assesses portfolio, project-based, and formative assessment approaches as viable replacements. &
\textit{(none)} &
I should evaluate how alternative assessments like portfolios, capstone projects, and performance-based evaluations can provide richer, more authentic measures of student learning.\\
\midrule
\path{cultural_and_linguistic_bias} &
Analyzes how test construction embeds cultural assumptions that disadvantage certain populations. &
\textit{(none)} &
I should examine how question framing, vocabulary choices, and cultural references in standardized tests systematically disadvantage English language learners and minority students.\\
\midrule
\path{predictive_validity_failure} &
Evaluates evidence that standardized test scores poorly predict real-world academic and career success. &
\textit{(none)} &
I should consider research showing that SAT/ACT scores are weak predictors of college GPA, graduation rates, and career achievement compared to high school grades and non-cognitive factors.\\
\midrule
\path{economic_costs_and_industry} &
Examines the multi-billion dollar testing industry and whether resources could be better allocated. &
\textit{(none)} &
I should analyze how billions spent on test development, administration, and preparation could instead fund teachers, counselors, and direct educational improvements.\\
\midrule
\path{international_comparison} &
Draws lessons from countries that have reduced or eliminated standardized testing with positive outcomes. &
\textit{(none)} &
I should examine how top-performing education systems like Finland have minimized standardized testing while achieving superior outcomes through trust in teachers and holistic assessment.\\
\bottomrule
\caption{Subtopics for \textbf{Standardized Testing} (domain-specific topical lenses for the abolition of standardized testing as a primary performance measure).}
\end{longtable}
\endgroup

\begingroup
\footnotesize
\sloppy
\renewcommand{\arraystretch}{1.15}
\setlength{\tabcolsep}{2pt}
\begin{longtable}{@{}L{0.18\textwidth} L{0.29\textwidth} L{0.08\textwidth} L{0.45\textwidth}@{}}
\label{tab:action-space-content-meat} \\
\toprule
\textbf{Name} & \textbf{Definition} & \textbf{Prefix} & \textbf{Internal reasoning} \\
\midrule
\endfirsthead
\toprule
\textbf{Name} & \textbf{Definition} & \textbf{Prefix} & \textbf{Internal reasoning} \\
\midrule
\endhead
\path{greenhouse_gas_emissions} &
Examines livestock's contribution to methane and CO2 emissions and how taxation can drive reduction. &
\textit{(none)} &
I should consider how animal agriculture accounts for approximately 14.5\% of global greenhouse gas emissions, with cattle producing significant methane, making meat taxation a climate mitigation tool.\\
\midrule
\path{land_and_water_resource_use} &
Analyzes the disproportionate land, water, and feed resources consumed by meat production. &
\textit{(none)} &
I should examine how livestock production uses 77\% of agricultural land while providing only 18\% of calories, and requires vastly more water per calorie than plant-based alternatives.\\
\midrule
\path{public_health_outcomes} &
Evaluates links between high meat consumption and chronic diseases including heart disease and cancer. &
\textit{(none)} &
I should analyze how excessive red and processed meat consumption is linked to colorectal cancer, cardiovascular disease, and diabetes, creating preventable health burdens.\\
\midrule
\path{animal_welfare_ethics} &
Examines the moral case for reducing industrial animal farming through economic disincentives. &
\textit{(none)} &
I should consider how factory farming subjects billions of sentient animals to confinement, suffering, and premature death, and how price signals can reduce demand for these practices.\\
\midrule
\path{healthcare_cost_externalities} &
Quantifies the public healthcare burden of diet-related diseases not reflected in meat prices. &
\textit{(none)} &
I should calculate how diet-related chronic diseases cost healthcare systems hundreds of billions annually, representing externalities that current meat prices fail to incorporate.\\
\midrule
\path{effectiveness_of_pigouvian_taxes} &
Draws on evidence from tobacco, sugar, and carbon taxes to assess behavioral taxation efficacy. &
\textit{(none)} &
I should evaluate how sin taxes on tobacco reduced smoking rates by 30--50\%, sugar taxes decreased consumption in Mexico and the UK, demonstrating that price signals effectively shift behavior.\\
\midrule
\path{food_equity_and_access} &
Analyzes how a meat tax could disproportionately affect low-income households and mitigation strategies. &
\textit{(none)} &
I should consider that while a regressive tax concern is valid, revenue can be redistributed through subsidies for plant-based foods and direct transfers to low-income families.\\
\midrule
\path{alternative_protein_innovation} &
Evaluates how price signals can accelerate the development and adoption of plant-based and cultured meat. &
\textit{(none)} &
I should examine how making conventional meat more expensive creates market incentives for investment in plant-based proteins, cellular agriculture, and precision fermentation.\\
\midrule
\path{international_policy_precedents} &
Draws lessons from countries and jurisdictions that have proposed or implemented meat taxation. &
\textit{(none)} &
I should analyze proposals and implementations in Denmark, Germany, the Netherlands, and New Zealand, examining political feasibility, design choices, and projected impacts.\\
\midrule
\path{agricultural_subsidy_reform} &
Examines how current subsidies artificially lower meat prices and how taxation can correct market distortions. &
\textit{(none)} &
I should consider how governments spend billions subsidizing animal feed, grazing land, and livestock operations, creating artificially cheap meat that masks its true environmental and health costs.\\
\bottomrule
\caption{Subtopics for \textbf{Meat Tax} (domain-specific topical lenses for imposing a tax on meat products).}
\end{longtable}
\endgroup

% Generic (topic-independent) subtopics used across all debate topics
\begingroup
\footnotesize
\sloppy
\renewcommand{\arraystretch}{1.15}
\setlength{\tabcolsep}{2pt}
\begin{longtable}{@{}L{0.18\textwidth} L{0.29\textwidth} L{0.08\textwidth} L{0.45\textwidth}@{}}
\label{tab:action-space-generic-subtopics} \\
\toprule
\textbf{Name} & \textbf{Definition} & \textbf{Prefix} & \textbf{Internal reasoning} \\
\midrule
\endfirsthead
\toprule
\textbf{Name} & \textbf{Definition} & \textbf{Prefix} & \textbf{Internal reasoning} \\
\midrule
\endhead
\path{cost_benefit_and_impact_analysis} &
Weighs economic, social, and practical consequences systematically. &
\textit{(none)} &
I should quantify and compare costs, benefits, and real-world impacts across economic, social, and environmental dimensions.\\
\midrule
\path{rights_and_liberties} &
Protects fundamental rights, freedoms, privacy, and individual autonomy. &
\textit{(none)} &
I should consider inalienable human rights, civil liberties, privacy protections, and the freedom to make one's own choices.\\
\midrule
\path{justice_and_fairness} &
Ensures equitable treatment, fair distribution, and equal opportunity. &
\textit{(none)} &
I should analyze whether outcomes, processes, and distributions are fair to all parties involved.\\
\midrule
\path{ethical_principles} &
Applies moral frameworks including duties, virtues, and care for others. &
\textit{(none)} &
I should evaluate actions based on moral rules, character virtues, relationships, and ethical obligations.\\
\midrule
\path{governance_and_accountability} &
Examines rule of law, transparency, democratic legitimacy, and institutional responsibility. &
\textit{(none)} &
I should consider legal frameworks, accountability mechanisms, democratic principles, and proper authority.\\
\midrule
\path{risk_and_unintended_consequences} &
Anticipates potential harms, unforeseen effects, and slippery slopes. &
\textit{(none)} &
I should identify risks, unintended outcomes, cascading effects, and potential for escalation.\\
\midrule
\path{feasibility_and_implementation} &
Assesses practical workability, technical constraints, and enforcement challenges. &
\textit{(none)} &
I should evaluate whether the proposal can actually be implemented and enforced effectively.\\
\midrule
\path{incentives_and_power_dynamics} &
Analyzes how rewards, penalties, and power structures shape behavior. &
\textit{(none)} &
I should examine what behaviors are encouraged, who holds power, and how interests align or conflict.\\
\midrule
\path{precedent_and_long_term_effects} &
Considers established patterns and future implications across generations. &
\textit{(none)} &
I should evaluate historical precedents, long-term vs short-term tradeoffs, and obligations to future generations.\\
\midrule
\path{stakeholder_responsibility} &
Clarifies duties of government, individuals, and institutions. &
\textit{(none)} &
I should analyze who bears responsibility, whether government, individuals, corporations, or other institutions.\\
\bottomrule
\caption{Generic subtopics used for the controllability experiment.}
\end{longtable}
\endgroup

\begingroup
\footnotesize
\sloppy
\renewcommand{\arraystretch}{1.15}
\setlength{\tabcolsep}{2pt}
\begin{longtable}{@{}L{0.18\textwidth} L{0.33\textwidth} L{0.14\textwidth} L{0.25\textwidth}@{}}
\toprule
\textbf{Name} & \textbf{Definition} & \textbf{Prefix} & \textbf{Internal reasoning} \\
\midrule
\endfirsthead
\toprule
\textbf{Name} & \textbf{Definition} & \textbf{Prefix} & \textbf{Internal reasoning} \\
\midrule
\endhead
\path{causal_reasoning} &
States causes, effects, consequences, or logical implications. &
\texttt{Therefore} &
\textit{(none)} \\
\midrule
\path{conditional} &
Introduces conditional, hypothetical, or counterfactual scenarios. &
\texttt{If} &
\textit{(none)} \\
\midrule
\path{concession_and_contrast} &
Acknowledges counterpoints or highlights opposing perspectives. &
\texttt{However} &
\textit{(none)} \\
\midrule
\path{addition_and_elaboration} &
Adds supporting information, elaborates, or strengthens a point. &
\texttt{Moreover} &
\textit{(none)} \\
\midrule
\path{evidence_and_authority} &
Cites evidence, data, or authoritative sources. &
\texttt{Evidence shows that} &
\textit{(none)} \\
\midrule
\path{exemplification} &
Provides concrete examples, illustrations, or case studies. &
\texttt{For example} &
\textit{(none)} \\
\midrule
\path{clarification_and_specification} &
Restates, clarifies, defines, or narrows down to specifics. &
\texttt{In other words} &
\textit{(none)} \\
\midrule
\path{emphasis_and_evaluation} &
Stresses importance or offers evaluative judgment. &
\texttt{Importantly} &
\textit{(none)} \\
\midrule
\path{sequence_and_transition} &
Signals progression through steps or shifts to a new topic. &
\texttt{Next} &
\textit{(none)} \\
\midrule
\path{conclusion_and_summary} &
Summarizes, concludes, or states the practical takeaway. &
\texttt{In conclusion} &
\textit{(none)} \\
\bottomrule
\caption{Argument generation action space: Structures (discourse moves).} \\
\end{longtable}
\endgroup

\subsection{Practitioner Guidance for Action Space Design}
\label{app:action-space-guidance}

Designing an effective action space is one of the most consequential choices when applying \state to a new domain. Below, we outline key decision points, trade-offs, and recommendations based on our experience across the experiments in this paper.

\paragraph{1. Identify controllable dimensions.}
Begin by enumerating the aspects of generation that can be meaningfully controlled at each step. These typically fall into categories such as content (what to say), structure (how to organize it), style (tone, register, formality), and strategy (rhetorical or reasoning approach). Where possible, ground dimensions in existing domain taxonomies---for example, \citet{wachsmuth-etal-2018-argumentation} for argumentation structure, \citet{dong2024formal} and \citet{didolkar_etal_2024_metacognitive_capabilities} for math, or other established reasoning taxonomies for reasoning-intensive tasks.

\paragraph{2. Sequential vs.\ single-step: Should dimensions vary per step?}
If the task involves multi-step generation (e.g., constructing an argument claim-by-claim), the action space should allow different choices at each step. For example, in argument generation, content strategy and structural choices naturally vary per claim. In contrast, some features of interest, such as target audience, argument topic, and stance, could be varied at the tree level rather than at the per-step level for trees with reasoning depth greater than one.

% \paragraph{3. Goal: explainability vs.\ performance.}
% \begin{itemize}[nosep]
%     \item \textbf{Explainability focus}: Design dimensions with clear semantic meaning. Use strict, faithful, or restructured synthesis modes depending on how closely generated text should match the action template.
%     \item \textbf{Performance focus}: The recommended design choices depend on the evaluator. With a \emph{programmatic evaluator} (e.g., checking code correctness or constraint satisfaction), prefer stepwise content additions, as programmatic evaluators cannot assess meta-reasoning. With a \emph{generator or reranker controller}, meta-cognitive actions (reflect \citep{muennighoff2025s1}, revise, verify, backtrack) can improve output quality; use conclusion synthesis to give the model flexibility in how it incorporates these actions.
% \end{itemize}

\paragraph{3. Prefix vs.\ internal reasoning.}
\state supports two mechanisms for injecting action guidance into the generator: \emph{prefix} (prefilled text that begins the generation) and \emph{internal reasoning} ((contextual guidance included in the assistant-message prefill before that prefix)).
\emph{Only one dimension can use a prefix}, since the prefix occupies a fixed position in the generated text and sets the beginning of the next reasoning step. All dimensions can use internal reasoning. If more than one dimension includes internal reasoning, we concatenate them one after the other (e.g., ``I should ...'' and ``I will ...''). In practice, the structural or discourse dimension benefits most from prefix control, since it directly shapes the opening of each generation step (e.g., ``\texttt{First, I will present a counterexample...}'').

\paragraph{4. Early stopping (\texttt{FINISH} action).}
Including a \texttt{FINISH} action allows the controller to terminate generation early, preventing overthinking or redundant steps. However, variable-length trajectories complicate attribution: trajectories of different lengths have different numbers of positional features, and shorter trajectories may systematically differ from longer ones for reasons unrelated to the actions chosen.

\paragraph{Application-specific tuning.}
Both the choice of synthesis mode and the action space itself may require domain-specific exploration. The recommendations above provide starting points, but practitioners should expect to iterate on action definitions and synthesis settings based on early experimental results.

\end{document}